%% file: main.tex
\documentclass{article} 
\PassOptionsToPackage{table}{xcolor}
\usepackage{preprint,times}

\input{math_commands.tex}

\usepackage{xcolor}
\usepackage{hyperref}
\usepackage{url}
\usepackage{xspace}
\usepackage{graphicx}
\usepackage{booktabs}
\usepackage{makecell}
\usepackage{enumitem}
\usepackage{subcaption}
\newcommand{\crank}[1]{\textsuperscript{\##1}} 
\newcommand{\system}{\textsc{CHORD}\xspace}
\definecolor{DetectedBlue}{RGB}{218,235,247}
\definecolor{BenignGray}{RGB}{242,242,242}
\newcommand{\detected}[1]{\cellcolor{DetectedBlue}#1}
\newcommand{\benign}[1]{\cellcolor{BenignGray}#1}

\usepackage{caption}
\usepackage{placeins}
\title{Coherence-Aware Distributional Evaluation of Open-Ended Text Generation}

\author{%
\textbf{Jinnuo Liu}$^{1,3,*}$\quad
\textbf{Junhao Zhu}$^{2,3,*}$\quad
\textbf{Weifeng Jiang}$^{3}$\quad
\textbf{Haoming Liu}$^{1, 3}$\quad
\textbf{Hongyi Wen}$^{1,3,\dagger}$\\[6pt]
$^{1}$New York University\\
$^{2}$Georgia Institute of Technology\\
$^{3}$Center for Data Science, NYU Shanghai\\[6pt]
{\small $^{*}$Equal contribution\quad $^{\dagger}$Corresponding author}\\[4pt]
{\small Code: \url{https://github.com/MAPS-research/CHORD}}\\
{\small Experiments: \url{https://github.com/MAPS-research/CHORD-Experiment}}%
}

\iclrfinalcopy 
\begin{document}

\maketitle

\input{sec/0_abstract}
\label{abstract}

\section{Introduction}
\begin{figure}[h]
    \centering
    \includegraphics[width=0.98\linewidth]{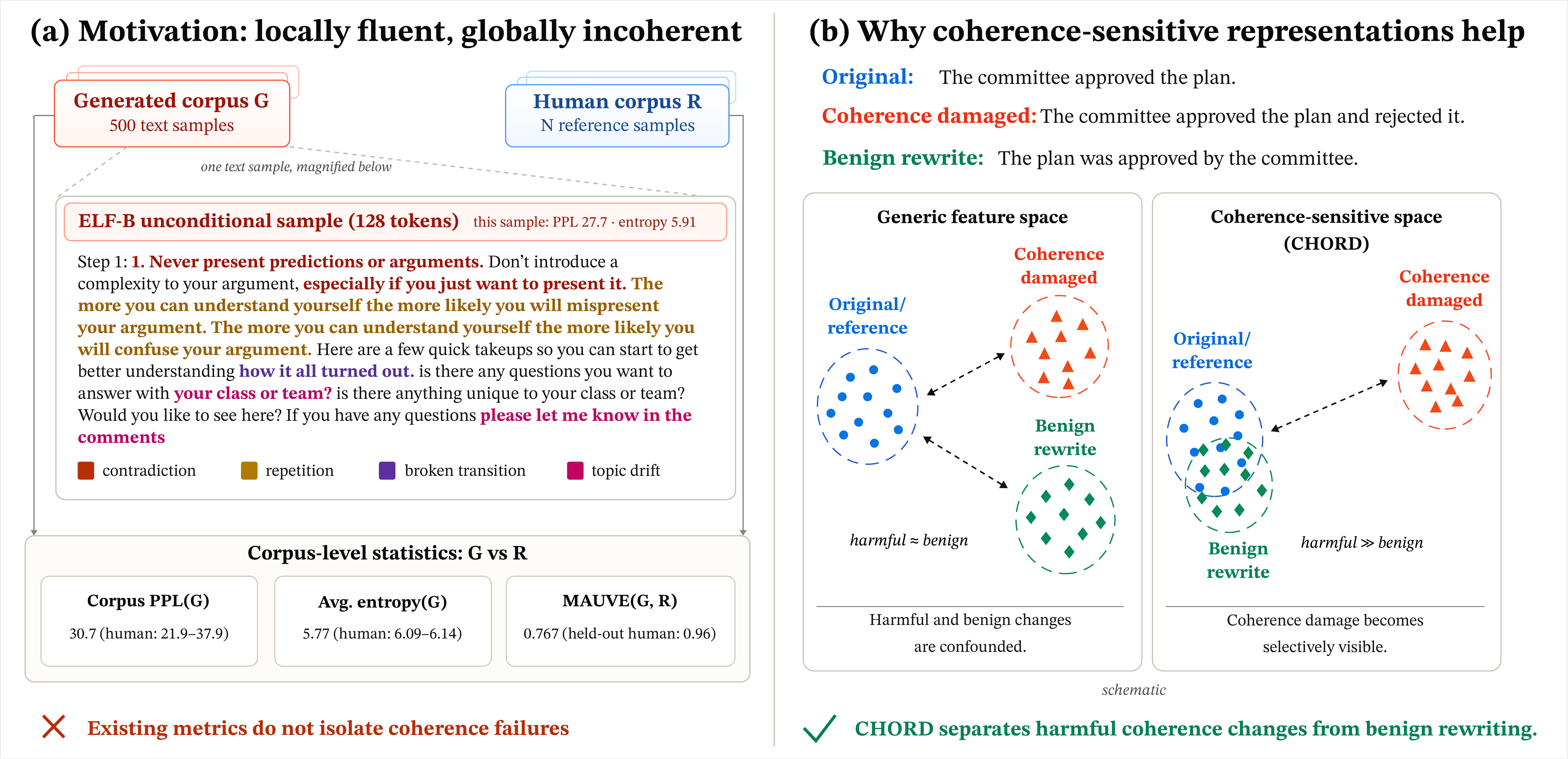}
    \caption{\textbf{Motivation and intuition for \system.} Existing corpus-level metrics may overlook text that is locally fluent yet globally incoherent; \system selectively distinguishes coherence degradation from meaning-preserving rewriting.}
    \label{fig:motivation}
\end{figure}
\input{sec/1_intro}

\section{Related Work}
\input{sec/2_related}

\section{Methodology}
\label{sec:method}
\input{sec/3_methodology}

\section{Experiments}
\label{sec:experiments}
\input{sec/4_experiments}

\section{Discussion}
\label{sec:discussions}
\input{sec/5_discussions}

\section{Conclusion}
\label{sec:conclusion}
\input{sec/6_conclusion}

\section*{Code Availability}
The \system metric, including featurization with the 27B encoder or a distilled student, RBF-MMD scoring with null calibration, and the complete distillation pipeline, is available at \url{https://github.com/MAPS-research/CHORD}.
The code, configurations, and instructions needed to reproduce every table and figure in the main paper and appendices, including the construction of the counterfactual evaluation set, are available at \url{https://github.com/MAPS-research/CHORD-Experiment}.

\input{sec/statements/ethics}

\clearpage
\bibliography{references}
\bibliographystyle{preprint}

\clearpage
\appendix
\input{sec/appendix}

\end{document}

%% file: math_commands.tex
\usepackage{amsmath,amsfonts,bm}

\def\eqref#1{equation~\ref{#1}}

\def\1{\bm{1}}

\DeclareMathAlphabet{\mathsfit}{\encodingdefault}{\sfdefault}{m}{sl}
\SetMathAlphabet{\mathsfit}{bold}{\encodingdefault}{\sfdefault}{bx}{n}



%% file: sec/0_abstract.tex
\begin{abstract}
Existing metrics for open-ended text generation measure likelihood, lexical diversity, or distributional similarity in generic representation space, yet they can miss fundamental dimensions of quality. A prominent blind spot is global coherence: a generated passage may be locally fluent while remaining globally contradictory, causally inconsistent, or topically disconnected. Such failures can still preserve the token-level and lexical statistics that existing metrics rely on. We identify representation as a central bottleneck in detecting these failures and introduce \system (\textbf{C}oherence-aware \textbf{H}idden-state \textbf{O}pen-generation \textbf{R}eference \textbf{D}istance), a coherence-sensitive distributional metric. \system encodes generated and human-written corpora in the hidden-state space of a frozen LLM using a coherence-eliciting prompt, and compares the resulting distributions using MMD with an RBF kernel. To validate that the metric responds to coherence degradation but not generic textual change, we construct a counterfactual evaluation suite that pairs graded coherence-degrading perturbations with meaning-preserving controls. \system selectively detects relation, discourse, structural, and mixture failures that perplexity, entropy, MAUVE, FBD, and MMD-based baselines either miss or cannot separate from benign rewriting. Factorial ablations show that representation is the primary source of coherence sensitivity, while RBF-MMD improves sample efficiency once the relevant distinctions become visible. Larger backbones capture finer-grained distinctions, but coherence prompting improves selectivity only when the backbone can follow the prompt.
On unconditional generation and prefix continuation, \system yields model rankings that strongly align with human judgments of whether outputs make sense and appear human-written. Together, these results establish representation design as central to reliable distributional evaluation.
\end{abstract}

%% file: sec/1_intro.tex
Evaluating open-ended text generation remains fundamentally challenging because generation quality is multidimensional, whereas existing automatic metrics capture only particular aspects of it: generative perplexity measures the predictability of generated text under an external language model \citep{holtzman2020curiouscaseneuraltext}, unigram entropy quantifies lexical diversity \citep{shannon1948mathematical}, and Self-BLEU estimates repetition across generated samples \citep{zhu2018texygenbenchmarkingplatformtext}. Yet recent studies show that optimizing these metrics need not yield improvements in overall generation quality \citep{perplexity-cannot-right-wrong-2026,hacking-gen-ppl-2026,generative-frontiers-2026}.

Open-ended generation often admits multiple valid outputs, making one-to-one  matching against reference outputs ill-suited for evaluation. Distributional metrics avoid the need for aligned references by comparing corpora of generated and human-written text in a shared feature space, thereby evaluating corpus-level similarity rather than sample-level correspondence. MAUVE \citep{pillutla2021mauvemeasuringgapneural,mauve-theory-practice-2023}, Fr\'echet-BERT Distance \citep{xiang2021assessing}, and MMD-based metrics \citep{chan2024distribution} instantiate this paradigm with different representations and divergences. However, a distributional metric can detect only those differences preserved by its feature space. Prior stress tests have shown that commonly used representations can be insensitive to discourse-level perturbations, including sentence reordering \citep{blind-spots-model-metrics-2023}.

\textit{Coherence} poses a particularly demanding test of this representation bottleneck. Current generators often produce multi-sentence passages that are locally fluent yet globally incoherent. A generated text sample may contradict earlier claims, reverse causal relations, disrupt discourse transitions, or drift across unrelated topics. Such failures can leave local fluency and much of the lexical content intact, providing little signal for perplexity and entropy. Generic representations can likewise conflate coherence violations with benign textual variation: for example, a contradiction-inducing edit and a meaning-preserving paraphrase may produce shifts of comparable magnitude in the representation space. Figure~\ref{fig:motivation} illustrates this failure mode. 

Dedicated coherence metrics \citep{entity-grid-2008,sentence-ordering-rnn-2018,union-2020,zhu2024coudacoherenceevaluationunified,zhao2023discoscoreevaluatingtextgeneration,ke2022ctrlevalunsupervisedreferencefreemetric} offers a complementary, sample-level approach. These methods score individual texts and are often evaluated on sentence ordering and topical consistency, with relation-level violations such as contradiction and causal reversal receiving less systematic coverage. Because they score samples individually, however, they are not designed to compare generated and human corpora as distributions. Our goal is to bridge this gap by making corpus-level distributional evaluation sensitive to a broader range of coherence failures. Accordingly, we define \textit{selectivity} as a central evaluation criterion: \textbf{a coherence-degrading perturbation should elicit a stronger metric response than a comparable meaning-preserving perturbation}.

We introduce \textbf{\system} (\textbf{C}oherence-aware \textbf{H}idden-state \textbf{O}pen-generation \textbf{R}eference \textbf{D}istance), a distributional metric designed to capture coherence differences between generated and human text. \system represents each text using hidden states extracted from a frozen LLM under a coherence-eliciting prompt, and compares the resulting distributions using MMD with an RBF kernel \citep{JMLR:v13:gretton12a}. To evaluate selectivity, we construct a counterfactual evaluation suite that pairs graded coherence-degrading perturbations with meaning-preserving controls. Across this suite, \system detects all tested failure types, whereas perplexity, entropy, and MAUVE with generic features either miss relation- and discourse-level errors or fail to distinguish them from benign rewriting.

A factorial ablation over feature representations and distance functions confirms that representation is the bottleneck: no distance function can recover coherence distinctions that are absent from the feature space. Further experiments reveal an interaction between backbone capability and coherence elicitation. Larger backbones capture finer-grained coherence distinctions, whereas targeted prompting improves selectivity only when the backbone can follow the instruction; with GPT-2 \citep{radford2019language}, it can even reduce selectivity. Finally, \system's model rankings on unconditional generation and prefix continuation strongly agree with human judgments.


%% file: sec/2_related.tex
\paragraph{Automatic evaluation of open-ended generation.}
Reference-free diagnostics characterize generated text alone: generative perplexity measures predictability \citep{holtzman2020curiouscaseneuraltext}, unigram entropy lexical diversity \citep{shannon1948mathematical}, and Self-BLEU cross-sample repetition \citep{zhu2018texygenbenchmarkingplatformtext}. They remain common for diffusion language models \citep{hu2026elfembeddedlanguageflows,guo2026continuouslatentdiffusionlanguage}, although recent analyses show they give an incomplete account of generation quality \citep{perplexity-cannot-right-wrong-2026,hacking-gen-ppl-2026,generative-frontiers-2026}. Distributional metrics instead compare generated and human corpora in a shared feature space, via information frontiers in MAUVE \citep{pillutla2021mauvemeasuringgapneural,mauve-theory-practice-2023}, the Fr\'echet distance in FBD \citep{xiang2021assessing}, or MMD \citep{chan2024distribution}. Their sensitivity depends on the representation: commonly used features can miss discourse-level perturbations such as sentence reordering \citep{blind-spots-model-metrics-2023}.

\paragraph{Coherence modeling and evaluation.}
Coherence evaluation has largely developed around two synthetic tasks. Permutation detection, distinguishing sentence-shuffled text from the original, motivated entity-grid models \citep{entity-grid-2008}, neural coherence models \citep{li-jurafsky-2017-neural}, and sentence-ordering objectives \citep{sentence-ordering-rnn-2018}. Sentence intrusion detection, spotting a sentence from an unrelated document, drove discriminators such as UNION \citep{union-2020} and CoUDA \citep{zhu2024coudacoherenceevaluationunified}, trained on programmatically corrupted negatives. Other metrics derive coherence scores from pretrained encoders without task-specific training, including DiscoScore \citep{zhao2023discoscoreevaluatingtextgeneration} and CTRLEval \citep{ke2022ctrlevalunsupervisedreferencefreemetric}. These approaches score individual samples and are mainly tested on ordering and topic consistency. \system instead compares corpora as distributions and extends detection beyond sentence permutation and document mix to relation-level failures such as contradiction and causal reversal.

\paragraph{Prompt-based sentence embeddings.}
\label{sec:prompt-embed}
Prompt-based embeddings place text in a task-specific template and read a hidden state from a frozen language model. PromptEOL \citep{prompteol-2024} asks the model to summarize the text in one word and uses the final-token state; MetaEOL \citep{metaeol-2024} varies the prompt task to capture different aspects of the text. \system adopts a fixed coherence-oriented template and uses the resulting last-token states for corpus-level distributional comparison.

%% file: sec/3_methodology.tex
\begin{figure}[t]
    \centering
    \includegraphics[width=0.88\linewidth]{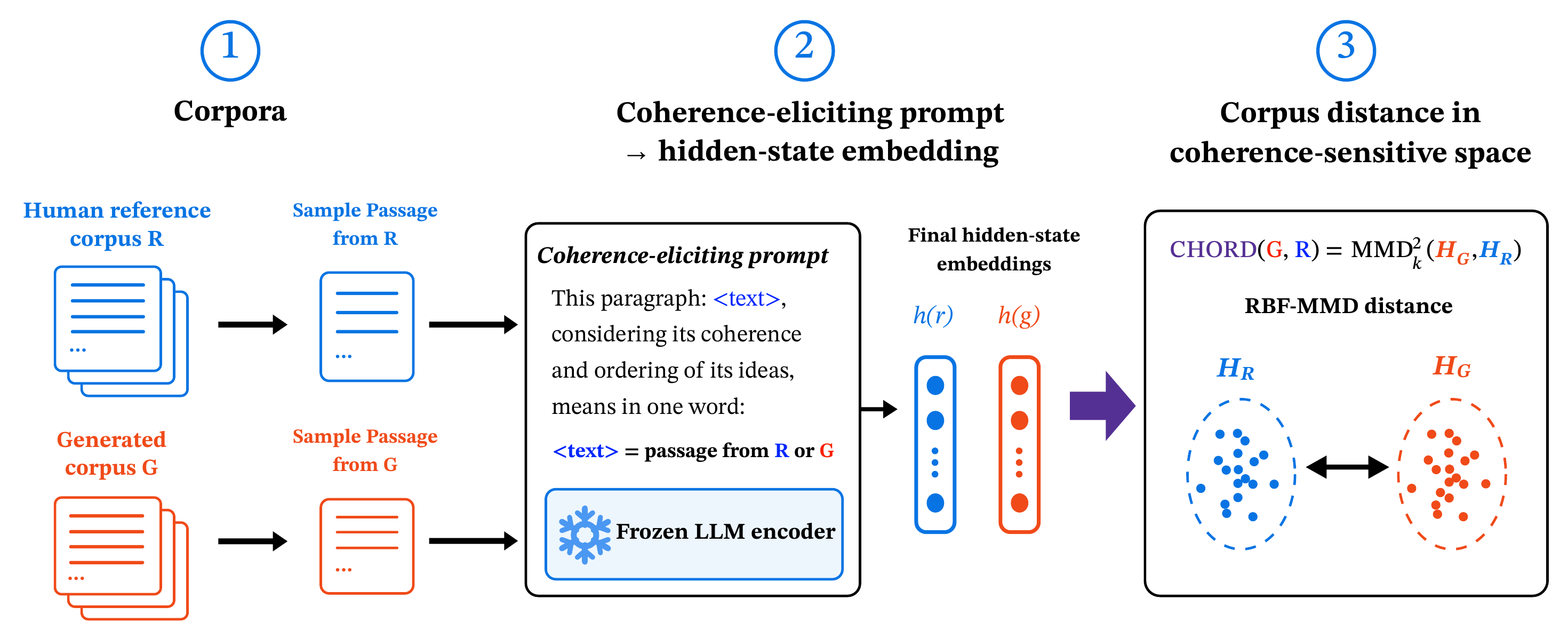}
    \caption{\textbf{\system overview.} A frozen LLM encodes prompted human and generated passages into coherence-oriented hidden states; RBF-MMD compares the resulting corpus distributions.}
    \label{fig:main_fig}
\end{figure}

\subsection{CHORD: Hidden-State Reference Distance}
\label{sec:chord}

Figure~\ref{fig:main_fig} summarizes the three-stage evaluation pipeline.
Let $R = \{r_i\}_{i=1}^{N_R}$ be a human reference corpus and $G = \{g_j\}_{j=1}^{N_G}$ a generated corpus. \system measures their distributional discrepancy using coherence-oriented representations and a kernel two-sample distance.

\paragraph{Coherence-oriented hidden-state representation.}
To represent a text sample $x$ with a frozen language model, we place it in a fixed coherence-eliciting template adapted from PromptEOL \citep{prompteol-2024}:
\begin{quote}
\texttt{This paragraph: "\{x\}", considering its coherence and ordering of its ideas, means in one word:}
\end{quote}
The compact-completion format requires the model to compress the input to a single next-token prediction, concentrating coherence information at the last token position. Let $p(x)$ denote the prompted input. The sample representation is the hidden state at layer $\ell$ of the last token position $t$:
\begin{equation}
h(x) = f_\theta\bigl(p(x)\bigr)_{\ell,\, t} \in \mathbb{R}^d,
\end{equation}
where $f_\theta$ is a frozen language model. \system uses these hidden states to compare corpora; it neither decodes continuations nor scores output tokens. Appendix~\ref{app:llm-judge} compares this signal with bag-of-words features, next-token predictions, and LLM-as-a-Judge scoring \citep{zheng2023judging}.

Unless otherwise specified, the \system encoder in this paper is a frozen Qwen3.5-27B model \citep{qwen3.5}. We extract its features from the \textbf{third-to-last} transformer layer. Appendix~\ref{app:layer} reports the layer-depth ablation motivating this choice, and Appendix~\ref{app:implementation} shows the full detail of the 27B encoder configuration.

The same design transfers across model families. Table~\ref{tab:main} shows that Gemma-2 \citep{gemmateam2024gemma2improvingopen}, Mistral \citep{mistral2025small3,ministral8b2410}, and Llama-3.1 \citep{grattafiori2024llama3} backbones detect a broad range of coherence failures using the same prompt, extraction rule, and distance, without family-specific tuning.

\paragraph{Gaussian RBF-MMD distance.}
Let $H_R=\{h(r_i)\}_{i=1}^{N_R}$ and $H_G=\{h(g_j)\}_{j=1}^{N_G}$. We compare these embedding distributions using the biased squared Maximum Mean Discrepancy (MMD) with a Gaussian RBF kernel \citep{JMLR:v13:gretton12a}:
\begin{equation}
\mathrm{CHORD}(G,R)=\widehat{\mathrm{MMD}}^2_k(H_G,H_R),
\qquad k(\mathbf{a},\mathbf{b})=\exp\!\left(-\frac{\|\mathbf{a}-\mathbf{b}\|^2}{2\sigma^2}\right).
\end{equation}
For each encoder and evaluation set, we set the bandwidth $\sigma$ to the median pairwise embedding distance on a held-out human-reference split, then keep it fixed across corpus comparisons. Appendix~\ref{app:implementation} gives the biased MMD estimator, calibration split sizes, and further details of the bandwidth-selection procedure.

\subsection{Diagnostic Meta-evaluation Protocol for Coherence Sensitivity}
\label{sec:meta_eval}

We evaluate coherence sensitivity using controlled perturbations paired with benign rewrites, null-standardized metric responses, and a statistical criterion for selective detection.

\paragraph{Perturbations and benign controls.}
We construct \textbf{nine} perturbation types from a shared pool of seed samples, organized into four categories that each target a different aspect of coherence: \textit{relation} perturbations (\textit{contradiction}, \textit{causal reversal}) introduce contradictions and causal reversals between sentences; \textit{discourse} perturbations (\textit{broken transition}, \textit{topic drift}) introduce logically disconnected transitions and topic drift; \textit{structural} perturbations (\textit{sentence permutation}, \textit{word shuffle}, \textit{repetition}) disrupt sentence order, shuffle word order, or duplicate sentences; and \textit{mixture} perturbations (\textit{DLM mix}, \textit{document mix}) replace sentences with material from unrelated sources. As benign controls, we paraphrase the same seed samples while preserving their claims and logical structure, establishing the metric response attributable to surface rewriting alone. Each perturbation type is applied at multiple severity levels. Appendix~\ref{app:counterfactuals} provides the full construction procedure and examples.

\paragraph{Null standardization.}
For each metric, let $s_M$ be the calculated raw distributional distance between the candidate set and the human reference set.
We repeatedly compute the same discrepancy between two
disjoint human-reference subsets, matched to the evaluation
sample sizes. The resulting null mean $\mu_0$ and standard
deviation $\sigma_0$ define
\begin{equation}
z_M=\frac{s_M-\mu_0}{\sigma_0}.
\end{equation}
This expresses each response relative to its own reference
sampling variation. Appendix~\ref{app:implementation}
specifies the statistic used for each metric and the
resampling procedure.

\paragraph{Detection criterion.}
Since benign rewriting also shifts scores, we compare each harmful perturbation against the most heavily rewritten benign control built from the same seed samples, a conservative baseline:
\begin{equation}
\Delta z = z_{\mathrm{harm}} - z_{\mathrm{benign}},
\end{equation}
where $z$ denotes the null-standardized score, so $\Delta z > 0$ means the harmful perturbation is penalized beyond benign rewriting. A condition is \emph{detected} if the lower bound of the 95\% CI of $\Delta z$, estimated from 200 paired bootstrap resamples over seeds, is positive (Appendix~\ref{app:implementation}).

%% file: sec/4_experiments.tex

We organize our experiments around four research questions:
\begin{itemize}[leftmargin=*, itemsep=1pt, topsep=2pt]
  \item \textbf{RQ1 (Selectivity).} Does \system respond to coherence-damaging perturbations while remaining stable under benign rewriting? (\S\ref{sec:exp-main})
  \item \textbf{RQ2 (Attribution).} Which representation and distance choices account for this selectivity? (\S\ref{sec:exp-attribution})
  \item \textbf{RQ3 (Model ranking).} Does \system yield informative rankings of existing generation models? (\S\ref{sec:exp-case})
  \item \textbf{RQ4 (Human alignment).} How well do \system scores agree with human judgments? (\S\ref{sec:exp-human})
\end{itemize}

\subsection{Setup}
\label{sec:exp-setup}

\paragraph{Counterfactual evaluation set.}
We sample passages from OpenWebText \citep{Gokaslan2019OpenWeb}, WikiText-103 \citep{merity2017pointer}, and Reddit TL;DR posts \citep{volske2017tldr,stiennon2020learning}, retaining those with at least four sentences and 100--400 words. The filtered passages form three disjoint pools: 500 seeds for counterfactual editing, 3{,}600 passages for the human reference corpus, and a replacement pool for mixture perturbations.

From the 500 seeds, we construct the nine perturbation types and benign control defined in Section~\ref{sec:meta_eval}. Qwen3-30B-A3B \citep{qwen3} generates the relation and discourse perturbations and benign controls, while rule-based operations produce the structural and mixture perturbations. The main findings persist with a Mistral editor (Appendix~\ref{app:editor-robustness}). We apply each perturbation type at multiple severity levels. Table~\ref{tab:main} reports the highest-severity condition for each type; Appendix~\ref{app:severity} gives the full severity results, and Appendix~\ref{app:counterfactuals} details their construction. All scores are null-standardized, and detection follows the paired bootstrap criterion in Section~\ref{sec:meta_eval}.

\paragraph{Baselines.}
We compare \system with GPT-2 generative perplexity \citep{holtzman2020curiouscaseneuraltext}, unigram entropy \citep{shannon1948mathematical}, MAUVE \citep{pillutla2021mauvemeasuringgapneural} using GPT-2 or ELECTRA \citep{clark2020electrapretrainingtextencoders} features, Fr\'echet-BERT Distance \citep{xiang2021assessing}, and MMD with MiniLM sentence embeddings \citep{chan2024distribution,wang2020minilmdeepselfattentiondistillation} (MMD-MiniLM). Appendix~\ref{app:baselines} provides implementation details and backbone references. For MAUVE, we null-standardize the frontier integral returned by the official implementation, a divergence that is zero when the two cluster histograms match, rather than the MAUVE score itself, using the same reference null as for the other metrics (Appendix~\ref{app:baselines}). Appendix~\ref{app:modern-baselines} reports comparisons with modern instruction-tuned embeddings, UniEval \citep{zhong2022unifiedmultidimensionalevaluatortext}, and BARTScore \citep{yuan2021bartscoreevaluatinggeneratedtext}; Appendix~\ref{app:llm-judge} evaluates same-backbone LLM judges, including G-Eval \citep{liu2023gevalnlgevaluationusing}.

\paragraph{\system encoders.}
Alongside the default Qwen3.5-27B encoder, we provide two distilled student encoders trained to match the teacher's PCA-projected embeddings, following OPRD's representation-alignment approach \citep{yang2026oprd}: Qwen3.5-2B, a high-fidelity variant that most closely tracks the teacher, and Qwen3.5-0.8B, a lightweight variant for faster, lower-memory inference. On 500 packed OpenWebText samples processed on a single H200, featurization takes 46.5\,s and 54.2\,GB peak GPU memory with the 27B encoder, compared with 5.3\,s and 7.1\,GB for 2B and 4.4\,s and 4.5\,GB for 0.8B. Appendices~\ref{app:distill} and~\ref{app:cost} give training, validation, and cost details.

\begin{table}[h]
\caption{\textbf{Null-standardized responses $z_M$ at the highest severity for each perturbation type.}
Blue shading marks detected conditions: the lower bound of the paired-bootstrap confidence interval for selectivity $\Delta z$ exceeds zero (Section~\ref{sec:meta_eval}). All \system backbones use the same coherence prompt and feature-extraction rule. Appendix~\ref{app:severity} reports results at every severity level.}
\label{tab:main}
\centering
\setlength{\tabcolsep}{3pt}
\renewcommand{\arraystretch}{0.92}
\resizebox{\textwidth}{!}{%
\small
\begin{tabular}{lcccccccccc}
\toprule
& \multicolumn{2}{c}{Relation} & \multicolumn{2}{c}{Discourse} & \multicolumn{3}{c}{Structural} & \multicolumn{2}{c}{Mixture} & \cellcolor{BenignGray}Control \\
\cmidrule(lr){2-3}\cmidrule(lr){4-5}\cmidrule(lr){6-8}\cmidrule(lr){9-10}\cmidrule(lr){11-11}
& \multicolumn{9}{c}{\scriptsize$\xleftarrow{\hspace{3.5cm}}$ harder to detect \hfill easier to detect $\xrightarrow{\hspace{3.5cm}}$} & \\
Method & \makecell{Causal\\reversal} & \makecell{Contra-\\diction} & \makecell{Broken\\transition} & \makecell{Topic\\drift} & \makecell{Sentence\\permutation} & \makecell{Word\\shuffle} & Repetition & DLM mix & Document mix & \cellcolor{BenignGray}\makecell{Benign\\paraphrase} \\
\midrule
\multicolumn{11}{l}{\emph{Likelihood / diversity statistics}} \\
gen-PPL (GPT-2) & 0.8 & 1.1 & 0.5 & 1.8 & \detected{7.4} & \detected{46.9} & \detected{62.4} & \detected{62.4} & \detected{35.9} & \benign{1.4} \\
Unigram entropy & $-$0.1 & 0.6 & 6.0 & 0.0 & $-$0.0 & 2.1 & 0.9 & \detected{22.1} & \detected{17.4} & \benign{2.6} \\
\midrule
\multicolumn{11}{l}{\emph{Distributional metrics}} \\
MAUVE (GPT-2) & 23.1 & 27.6 & 21.0 & 18.9 & 24.5 & 1.1 & \detected{56.5} & \detected{35.8} & \detected{44.9} & \benign{27.3} \\
MAUVE (ELECTRA) & 3.0 & 2.6 & 3.7 & 2.4 & \detected{13.7} & \detected{72.5} & \detected{89.7} & \detected{81.2} & \detected{76.2} & \benign{2.5} \\
FBD (BERT) & 0.2 & 0.2 & 0.8 & 0.5 & 1.0 & \detected{9.7} & \detected{60.7} & \detected{24.5} & \detected{12.5} & \benign{0.2} \\
MMD (MiniLM) & $-$0.0 & $-$0.0 & \detected{33.3} & \detected{4.9} & $-$0.3 & 0.4 & \detected{18.9} & \detected{69.5} & \detected{62.5} & \benign{$-$0.3} \\
\midrule
\multicolumn{11}{l}{\emph{Ours}} \\
\system (Qwen3.5-27B) & \detected{25.6} & \detected{84.3} & \detected{331} & \detected{79.7} & \detected{358} & \detected{740} & \detected{857} & \detected{1150} & \detected{1082} & \benign{6.0} \\
\system (Qwen3.5-9B) & 17.4 & \detected{26.9} & \detected{109} & \detected{27.5} & \detected{96.3} & \detected{344} & \detected{671} & \detected{882} & \detected{726} & \benign{10.9} \\
\system (Qwen3.5-2B distilled) & \detected{20.1} & \detected{73.1} & \detected{258} & \detected{45.9} & \detected{190} & \detected{579} & \detected{1001} & \detected{1241} & \detected{1211} & \benign{4.3} \\
\system (Qwen3.5-0.8B distilled) & \detected{16.3} & \detected{43.3} & \detected{257} & \detected{34.7} & \detected{173} & \detected{770} & \detected{1191} & \detected{1222} & \detected{1166} & \benign{2.5} \\
\addlinespace[2pt]
\system (Gemma-2-27B-it) & \detected{24.7} & \detected{80.5} & \detected{270} & \detected{66.1} & \detected{182} & \detected{514} & \detected{948} & \detected{845} & \detected{627} & \benign{6.6} \\
\system (Gemma-2-9B-it) & \detected{11.0} & \detected{27.0} & \detected{135} & \detected{27.0} & \detected{120} & \detected{328} & \detected{449} & \detected{804} & \detected{691} & \benign{3.0} \\
\system (Mistral-Small-24B) & \detected{17.1} & \detected{40.7} & \detected{161} & \detected{41.6} & \detected{175} & \detected{315} & \detected{808} & \detected{653} & \detected{550} & \benign{3.5} \\
\system (Ministral-8B) & 10.8 & \detected{16.5} & \detected{74.8} & 12.1 & \detected{62.9} & \detected{235} & \detected{467} & \detected{535} & \detected{368} & \benign{5.7} \\
\system (Llama-3.1-8B-Instruct) & 5.6 & \detected{14.0} & \detected{109} & \detected{11.7} & \detected{53.6} & \detected{152} & \detected{143} & \detected{543} & \detected{306} & \benign{3.3} \\
\bottomrule
\end{tabular}}
\end{table}

\subsection{Counterfactual Robustness}
\label{sec:exp-main}

Table~\ref{tab:main} compares all metrics on our counterfactual evaluation set at the highest perturbation severity. The metrics differ substantially in the failures they detect.

Generative perplexity detects word shuffle, repetition, and mixture but responds to contradiction and causal reversal much like the benign control. Unigram entropy detects mainly mixture perturbations. FBD also fails to detect relation-level errors, while MMD-MiniLM detects discourse but not relation-level failures.

MAUVE depends strongly on its features. With GPT-2, its contradiction shift (27.6) nearly matches benign paraphrasing (27.3); ELECTRA reduces the benign response but is still insensitive to relation-level failures.

\textbf{\system distinguishes all nine perturbation types from benign paraphrasing.} Responses are strongest for structural and mixture failures but remain selective for contradiction, causal reversal, broken transitions, and topic drift. Both distilled Qwen3.5-2B and 0.8B encoders detect all nine types, though with weaker relation-level responses than the 27B encoder.

\textbf{This pattern transfers across backbone families.} With the prompt, extraction layer, and distance fixed, all Gemma, Mistral, and Llama backbones in Table~\ref{tab:main} detect structural and mixture failures. The 24--27B Gemma and Mistral models detect all nine types, whereas smaller Ministral and Llama models miss some relation or discourse failures.

\subsection{What Makes CHORD Coherence-Sensitive?}
\label{sec:exp-attribution}

MAUVE and \system both compare distributions of text representations, but differ in three components:\textbf{ the backbone, the representation extraction method, and the distributional distance}. We isolate their contributions below.

\subsubsection{Representation and distance}

\begin{figure}[t]
\centering
\begin{minipage}[t]{0.48\textwidth}
\centering
\includegraphics[width=\linewidth]{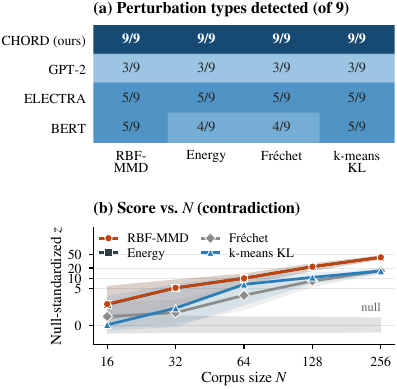}
\caption{\textbf{Representation and distance ablations.}
\textbf{(a)}~Number of perturbation types detected for each representation and distributional distance. Off-diagonal settings include GPT-2 features with RBF-MMD and coherence-prompted Qwen features with MAUVE's k-means KL.
\textbf{(b)}~Null-standardized score versus corpus size on contradiction.}
\label{fig:factorial}
\end{minipage}\hfill
\begin{minipage}[t]{0.48\textwidth}
\centering
\includegraphics[width=\linewidth]{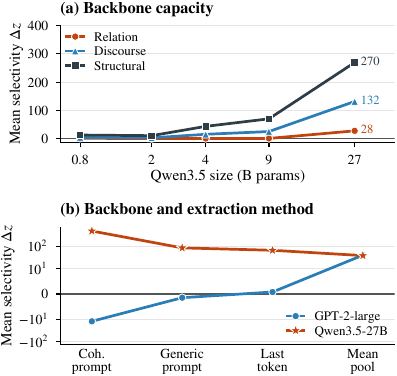}
\caption{\textbf{Backbone capacity and coherence prompting.}
Both panels report mean selectivity $\Delta z=z_{\mathrm{harm}}-z_{\mathrm{benign}}$.
\textbf{(a)} Qwen3.5 scaling with the coherence prompt, representation layer, and RBF-MMD fixed.
\textbf{(b)} Four extraction methods on Qwen3.5-27B and GPT-2-large, averaged over all harmful conditions (symmetric-log scale). }
\label{fig:encoder}
\end{minipage}
\end{figure}

We pair four representations (coherence-prompted Qwen, GPT-2, ELECTRA, and BERT \citep{devlin2019bertpretrainingdeepbidirectional}) with four distributional distances: RBF-MMD, energy distance \citep{szekely2013energy}, Fr\'echet distance \citep{fid-ttur-2017}, and MAUVE's k-means KL (its frontier integral; Appendix~\ref{app:baselines}). This yields 16 combinations evaluated on the nine perturbation types in Table~\ref{tab:main}.

Figure~\ref{fig:factorial}a shows that detection depends primarily on the representation. Swapping distances between MAUVE and \system makes this clear: GPT-2 with RBF-MMD detects only three perturbation types, whereas coherence-prompted Qwen with k-means KL detects all nine. Across the full grid, changing representations affects detection much more than changing distances. \textbf{The representation determines which coherence failures are exposed to the distributional comparison.}

\textbf{Distance choice mainly affects sample efficiency.} Figure~\ref{fig:factorial}b shows that RBF-MMD detects contradiction at smaller corpus sizes and with lower variance than Fr\'echet distance or k-means KL. We therefore use RBF-MMD as the default distance.

\subsubsection{Backbone capacity and coherence prompting}
\label{sec:backbone-elicitation}

Figure~\ref{fig:encoder}a isolates the effect of Qwen3.5 backbone size, holding the prompt, extraction layer, and distance fixed. Selectivity for structural and discourse failures improves at intermediate scales, whereas responses to relation-level failures remain weak until the largest model. This pattern agrees with the cross-family results in Table~\ref{tab:main}: \textbf{larger backbones capture finer coherence distinctions in their hidden representations.}

\textbf{The benefit of coherence prompting also depends on the backbone.} In Figure~\ref{fig:encoder}b, the coherence-prompted final-token representation is substantially more selective than unprompted alternatives on Qwen3.5-27B. On GPT-2-large, however, mean pooling performs best, and coherence prompting can yield negative selectivity. This pattern persists across all GPT-2 sizes (Appendix~\ref{app:prompt-ablation}, Table~\ref{tab:gpt2-extraction-grid}), suggesting that its effectiveness depends on whether the backbone can encode the target property.

Appendices~\ref{app:prompt-ablation} and~\ref{app:positional} ablate prompt design, extraction, and error position; Appendix~\ref{app:llm-judge} compares the hidden-state signal with token-space measures and direct LLM judgments.

\subsection{Evaluating Unconditional Generation and Conditional Text}
\label{sec:exp-case}

To assess how well \system and existing metrics distinguish among text generation models, we evaluate them on unconditional generation and prefix continuation, two common settings for open-ended text generation.

\begin{table}[t]
\caption{\textbf{Unconditional generation on OpenWebText.}
We report mean$\pm$std over 10 non-overlapping sample sets, each containing 500 samples of 512 tokens. \system reports RBF-MMD$^2$ ($\times 10^{-2}$); scores are comparable only within the same encoder column. Details are provided in Appendix~\ref{app:mauve-length}.}
\label{tab:case-uncond}
\centering
\setlength{\tabcolsep}{3pt}
\renewcommand{\arraystretch}{1.1}
\small
\resizebox{\textwidth}{!}{%
\begin{tabular}{lcccccc}
\toprule
Generator & \makecell{\system~$\downarrow$\\(Qwen3.5-27B)} & \makecell{\system~$\downarrow$\\(Qwen3.5-2B distilled)} & \makecell{\system~$\downarrow$\\(Qwen3.5-0.8B distilled)} & gen-PPL $\downarrow$ & MAUVE $\uparrow$ & entropy $\uparrow$ \\
\midrule
Held-out human (packed) & 0.17\crank{1}\,{\scriptsize($\pm$0.04)} & 0.16\crank{1}\,{\scriptsize($\pm$0.03)} & 0.15\crank{1}\,{\scriptsize($\pm$0.05)} & 18.94\crank{3}\,{\scriptsize($\pm$0.31)} & 0.95\crank{1}\,{\scriptsize($\pm$0.01)} & 7.51\crank{2}\,{\scriptsize($\pm$0.02)} \\
\midrule
\makecell[l]{GPT-2-large\\(774M, nucleus)} & 19.84\crank{2}\,{\scriptsize($\pm$1.02)} & 9.98\crank{2}\,{\scriptsize($\pm$0.74)} & 10.40\crank{3}\,{\scriptsize($\pm$0.77)} & 6.70\crank{1}\,{\scriptsize($\pm$0.10)} & 0.77\crank{5}\,{\scriptsize($\pm$0.04)} & 7.03\crank{7}\,{\scriptsize($\pm$0.02)} \\
\makecell[l]{GPT-2-medium\\(355M, nucleus)} & 28.37\crank{3}\,{\scriptsize($\pm$0.90)} & 10.78\crank{3}\,{\scriptsize($\pm$0.59)} & 10.36\crank{2}\,{\scriptsize($\pm$0.66)} & 10.10\crank{2}\,{\scriptsize($\pm$0.11)} & 0.81\crank{4}\,{\scriptsize($\pm$0.03)} & 7.06\crank{6}\,{\scriptsize($\pm$0.01)} \\
\makecell[l]{ELF-L (652M)\\\citep{hu2026elfembeddedlanguageflows}} & 51.19\crank{4}\,{\scriptsize($\pm$1.09)} & 68.20\crank{4}\,{\scriptsize($\pm$1.33)} & 69.72\crank{4}\,{\scriptsize($\pm$1.45)} & 23.15\crank{5}\,{\scriptsize($\pm$0.53)} & 0.09\crank{7}\,{\scriptsize($\pm$0.01)} & 7.07\crank{5}\,{\scriptsize($\pm$0.02)} \\
\makecell[l]{LangFlow (171M)\\\citep{chen2026langflowcontinuousdiffusionrivals}} & 52.58\crank{5}\,{\scriptsize($\pm$1.12)} & 70.84\crank{5}\,{\scriptsize($\pm$1.54)} & 78.79\crank{7}\,{\scriptsize($\pm$1.73)} & 18.98\crank{4}\,{\scriptsize($\pm$0.49)} & 0.65\crank{6}\,{\scriptsize($\pm$0.05)} & 7.60\crank{1}\,{\scriptsize($\pm$0.03)} \\
\makecell[l]{SEDD-small (170M)\\\citep{sedd-2024}} & 57.05\crank{6}\,{\scriptsize($\pm$1.02)} & 74.95\crank{6}\,{\scriptsize($\pm$1.27)} & 77.64\crank{6}\,{\scriptsize($\pm$1.14)} & 69.21\crank{6}\,{\scriptsize($\pm$0.96)} & 0.89\crank{2}\,{\scriptsize($\pm$0.02)} & 7.21\crank{4}\,{\scriptsize($\pm$0.02)} \\
\makecell[l]{MDLM (170M)\\\citep{mdlm-2024}} & 57.97\crank{7}\,{\scriptsize($\pm$1.20)} & 75.06\crank{7}\,{\scriptsize($\pm$1.64)} & 77.21\crank{5}\,{\scriptsize($\pm$1.49)} & 73.92\crank{7}\,{\scriptsize($\pm$2.01)} & 0.87\crank{3}\,{\scriptsize($\pm$0.02)} & 7.35\crank{3}\,{\scriptsize($\pm$0.03)} \\
\bottomrule
\end{tabular}}
\end{table}

\paragraph{Unconditional generation.}
Table~\ref{tab:case-uncond} shows a consistent three-tier pattern across the 27B encoder and both distilled variants: held-out human text is closest to the reference distribution, followed by GPT-2 outputs, then diffusion and flow outputs from SEDD, MDLM, ELF-L, and LangFlow. The 2B encoder exactly reproduces the 27B ranking (Spearman $\rho=1.00$). The 0.8B encoder preserves the three tiers but differs within them ($\rho=0.82$), assigning nearly identical scores to the two GPT-2 models and ranking LangFlow last. Qualitative inspection reveals more frequent topic shifts, repetition, and disrupted sentence transitions in diffusion and flow outputs (Appendix~\ref{app:qualitative-uncond}); a blind human evaluation corroborates the ordering (Appendix~\ref{app:quantitative-uncond}).

The baseline metrics do not recover this grouping. MAUVE ranks SEDD and MDLM closest to human text, above both GPT-2 models, and assigns ELF-L a near-zero score even though \system places it next to LangFlow. Generative perplexity favors predictable text rather than closeness to the human reference: both GPT-2 models score well below human text, LangFlow and ELF-L fall close to it, and only SEDD and MDLM are clearly separated. Unigram entropy varies little across generators and tracks lexical diversity, ranking LangFlow above human text and both GPT-2 models last, the opposite of their placement under \system.

\begin{figure}[h]
\centering
\includegraphics[width=0.92\linewidth]{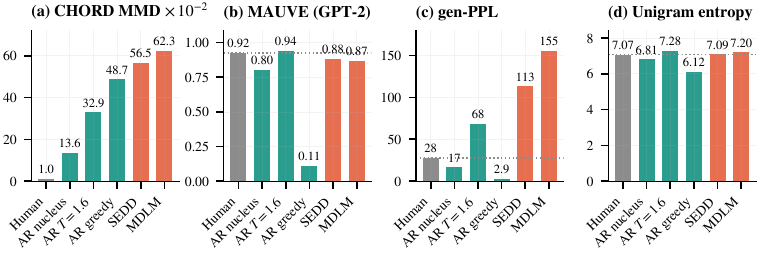}
\caption{\textbf{Prefix continuation on OpenWebText.}
Each model generates a 128-token continuation from a 128-token human prefix.
Dotted lines mark the held-out human value.}
\label{fig:continuation}
\end{figure}

\paragraph{Prefix continuation.}
In this task, each model receives the same 128-token human prefix from OpenWebText and generates a 128-token continuation. Figure~\ref{fig:continuation} reports \system scores from the Qwen3.5-27B encoder alongside the comparison metrics. \system recovers the same three-tier pattern: human continuations are closest to the reference, autoregressive continuations occupy the middle tier, and diffusion-LM continuations receive the largest shifts. Within the autoregressive tier, less natural decoding strategies (greedy and high temperature) move progressively farther from the human baseline.

The comparison metrics again produce different orderings. MAUVE places high-temperature autoregressive continuations above the human baseline and keeps diffusion continuations close to human text. Generative perplexity favors greedy decoding, while unigram entropy mainly captures the loss of diversity under greedy generation.

\subsection{Agreement with Human Judgments}
\label{sec:exp-human}

\begin{table}[!htbp]
\caption{\textbf{Agreement with human system rankings.}
Spearman correlation ($\rho$) between each metric and Bradley--Terry rankings derived from the human study of \citet{pillutla2021mauvemeasuringgapneural}. The study contains 3{,}240 pairwise judgments over eight GPT-2 generation settings.}
\label{tab:human-eval-main}
\centering
\setlength{\tabcolsep}{5pt}
\small
\begin{tabular}{lccccccc}
\toprule
& \multicolumn{2}{c}{\system (ours)} & \multicolumn{5}{c}{Existing metrics} \\
\cmidrule(lr){2-3}\cmidrule(lr){4-8}
Human dimension & \makecell{Qwen3.5\\27B} & \makecell{Qwen3.5-2B\\distilled} & \makecell{MAUVE\\(GPT-2)} & \makecell{MAUVE\\(ELECTRA)} & \makecell{FBD\\(BERT)} & \makecell{MMD\\(MiniLM)} & \makecell{gen-PPL\\(GPT-2)} \\
\midrule
Interesting & 0.86 & 0.76 & 0.00 & 0.79 & 0.76 & \textbf{0.90} & 0.64 \\
Makes sense & \textbf{0.98} & 0.93 & $-$0.19 & 0.93 & 0.93 & 0.90 & 0.88 \\
Human-like & \textbf{0.98} & 0.95 & $-$0.21 & 0.90 & 0.95 & 0.93 & 0.88 \\
\bottomrule
\end{tabular}
\end{table}

To independently validate the metric rankings, we use the human judgments released by \citet{pillutla2021mauvemeasuringgapneural}. Following the original protocol, we fit Bradley--Terry scores \citep{bradley1952rank} for \emph{interesting}, \emph{makes sense}, and \emph{human-like}, then compute their Spearman correlations with metric rankings (Table~\ref{tab:human-eval-main}). Settings vary GPT-2 size and decoding (Appendix~\ref{app:human-eval}).

\system (Qwen3.5-27B) achieves the highest correlation on \emph{makes sense} and \emph{human-like} ($\rho=0.98$ on both), the two dimensions most closely tied to coherence. The distilled Qwen3.5-2B encoder reaches $\rho=0.93$ and $0.95$, matching the best existing metric on each dimension. On \emph{interesting}, which mixes coherence with novelty and subjective preference, MMD-MiniLM is slightly stronger ($0.90$ vs.\ $0.86$). MAUVE with GPT-2 features correlates poorly under this protocol, consistent with its window-length sensitivity (Appendix~\ref{app:mauve-length}).

%% file: sec/5_discussions.tex
\paragraph{Coherence failures differ in detection difficulty.}
Our scaling experiments suggest a hierarchy: smaller backbones detect structural errors, discourse sensitivity grows with scale, and relation-level failures remain the hardest to detect.
Contradictions and causal reversals preserve much of the vocabulary and local fluency while changing logical relations.
\textbf{They therefore provide demanding tests of whether generation evaluators capture global coherence.}
A further direction is to explore \system as a corpus-level training objective, building on work that optimizes generators with representation-space metrics \citep{yang2026representation}.

\paragraph{Targeted representations are central to evaluation.}
Our ablations show that the representation largely determines which failures become detectable. Coherence prompting improves selectivity on Qwen3.5-27B but reduces it on GPT-2, suggesting that its effectiveness depends on backbone capability.
This motivates designing distributional evaluators around representations that emphasize the intended quality dimension.
A preliminary experiment beyond coherence supports this direction (Appendix~\ref{app:safety}). When a small fraction of the safe assistant responses in a corpus is replaced with unsafe ones, \system detects the shift at 5\% prevalence, and swapping the coherence prompt for a safety-oriented prompt widens the margin over matched safe replacements, without safety labels.

\paragraph{Selectivity is a useful criterion for metric validation.}
Our counterfactual experiments show that a metric can respond strongly to both coherence damage and benign rewriting.
Comparing these responses on the same source texts helps distinguish sensitivity to coherence from sensitivity to textual change in general. This validation principle may extend to other quality dimensions by pairing targeted
failures with controls that preserve the property being evaluated.

%% file: sec/6_conclusion.tex
Evaluating open-ended generation remains difficult when generic feature spaces fail to encode important quality distinctions. This work identifies representation design as a central bottleneck in distributional text evaluation and develops a counterfactual protocol for testing whether a metric responds selectively to quality degradation rather than benign variation. We introduce \system, a coherence-sensitive distributional metric that distinguishes relation, discourse, structural, and mixture failures from matched meaning-preserving rewrites when existing metrics miss or conflate them. \system produces informative rankings on real generation systems that align with human evaluations. More broadly, our results show that distributional metrics can capture additional dimensions of generation quality when their representations are designed to make those dimensions visible.

%% file: sec/statements/ethics.tex
\section*{Broader Impact}

This work studies corpus-level evaluation of textual coherence.
Our experiments include synthetic coherence errors and safety-related examples that may contain harmful or offensive content; these are used solely to evaluate metric behavior. 
Coherence does not imply factual accuracy, fairness, or safety, and a low distributional discrepancy does not establish that individual outputs are reliable or suitable for deployment.
The metric may also reflect biases in the underlying language model and reference corpus. Its scores should therefore be interpreted alongside task-specific evaluation and human judgment.

%% file: sec/appendix.tex
\input{sec/appendix/A_limitations}

\input{sec/appendix/B_implementation}
\input{sec/appendix/C_evalset}
\input{sec/appendix/D_baselines}
\input{sec/appendix/E_modern_baselines}
\input{sec/appendix/F_llm_judge}
\input{sec/appendix/G_attribution}
\input{sec/appendix/H_severity}
\input{sec/appendix/I_distillation}
\input{sec/appendix/J_mauve_length}
\input{sec/appendix/K_continuation}
\input{sec/appendix/L_qualitative}
\input{sec/appendix/M_quantitative_uncond}
\input{sec/appendix/N_positional}
\input{sec/appendix/O_layer_ablation}
\input{sec/appendix/P_compute_cost}
\input{sec/appendix/Q_qa}
\input{sec/appendix/R_human_eval}
\input{sec/appendix/S_safety}

%% file: sec/appendix/A_limitations.tex
\section{Limitations}
\label{sec:limitation}
\input{sec/limitation}

%% file: sec/limitation.tex
\system is a coherence-sensitive reference distance, not a universal measure of text quality. A high score indicates a distributional shift in the selected hidden-state space; it is not an absolute judgment of writing quality, factuality, usefulness, or human preference. Scores are also relative to the human reference corpus and scoring protocol, so meaningful comparisons should keep the reference domain, passage format, encoder, prompt, and representation layer fixed.

Our primary evidence concerns global coherence. More fine-grained relation errors remain more difficult. Extending the representation-centered approach to faithfulness, controllability, style, or other properties requires property-specific prompts, controls, and validation. The source-conditioned QA study in Appendix~\ref{app:qa-extension} demonstrates feasibility for one such setting but does not establish a general-purpose faithfulness metric.

The counterfactual evaluation is intentionally aligned with the target property and therefore should be viewed as a diagnostic test of selective coherence sensitivity instead of a complete proxy for real-world generation quality. Moreover, relation-level errors such as contradiction and causal reversal necessarily alter propositional content, making it impossible to construct benign controls that exactly match every dimension of semantic change.

The default Qwen3.5-27B encoder is computationally demanding. The distilled Qwen3.5-2B variant substantially reduces cost and preserves the broad detection pattern, but loses sensitivity on some difficult relation-level errors and may generalize less reliably to unseen domains. More broadly, the representation depends on the language, instruction-following behavior, and pretraining biases of the backbone. Evaluating additional model families and multilingual settings remains future work.

Finally, our counterfactual evaluation set is not exhaustive. Real generations may exhibit long-range planning failures, pragmatic inconsistencies, domain-specific discourse conventions, or multi-document conflicts that are not covered by the current perturbations. Although our real-system experiments are supported by both targeted blind human evaluation and correlations with released human judgments, broader validation across stronger model families, domains, and generation settings would further establish the generality of the metric.

%% file: sec/appendix/B_implementation.tex
\section{Implementation Details}
\label{app:implementation}

We list here the configuration details needed to reproduce the experiments; the metric and evaluation protocol are defined in Section~\ref{sec:method}, and compute and memory costs are reported in Appendix~\ref{app:cost}.

\paragraph{Encoder configuration.}
The default \system encoder is a frozen Qwen3.5-27B model \citep{qwen3.5}. Each text sample is placed in the coherence-eliciting template from Section~\ref{sec:chord} and processed with a single forward pass. The representation is the hidden state at the last token position of the prompt, extracted from the third-to-last transformer layer, which for Qwen3.5-27B is layer 62 of 64 and yields a 5120-dimensional vector. Section~\ref{sec:backbone-elicitation} examines how detection of coherence failures varies across Qwen3.5 encoders from 0.8B to 27B, and Appendix~\ref{app:layer} reports a layer-depth sweep on the 9B and 27B backbones showing that the third-to-last layer is among the strongest read depths for detecting coherence damage.

For the cross-family rows of Table~\ref{tab:main}, we change only the frozen backbone. Gemma-2, Mistral, and Llama-3.1 use the same coherence prompt, 512-token budget, third-to-last-layer rule, RBF-MMD, null calibration, and bootstrap criterion as Qwen3.5; the prompt and layer are not tuned separately by different backbone families. Inputs are encoded as raw completion strings. Gemma-2 uses eager attention to preserve its native attention soft-capping during hidden-state extraction.

\paragraph{RBF-MMD estimator.}
For the embedding sets $H_G=\{h(g_j)\}_{j=1}^{N_G}$ and $H_R=\{h(r_i)\}_{i=1}^{N_R}$, we use the biased squared MMD estimator
\begin{equation}
\begin{aligned}
\widehat{\mathrm{MMD}}^2_k(H_G,H_R)
&=\frac{1}{N_G^2}\sum_{j,j'=1}^{N_G}k\bigl(h(g_j),h(g_{j'})\bigr)
+\frac{1}{N_R^2}\sum_{i,i'=1}^{N_R}k\bigl(h(r_i),h(r_{i'})\bigr)\\
&\quad-\frac{2}{N_GN_R}\sum_{j=1}^{N_G}\sum_{i=1}^{N_R}k\bigl(h(g_j),h(r_i)\bigr),
\end{aligned}
\end{equation}
where $k(\mathbf{a},\mathbf{b})=\exp(-\|\mathbf{a}-\mathbf{b}\|^2/(2\sigma^2))$. The within-corpus sums include self-comparisons; the unbiased estimator excludes these terms and adjusts the denominators. We use the biased estimator for its non-negativity and lower variance at the corpus sizes considered here. Its null offset is accounted for by the reference standardization below.

\paragraph{Bandwidth calibration.}
For each representation and each evaluation set, we fit the RBF bandwidth once on a calibration split $R_{\mathrm{cal}}$ of that set's human reference, disjoint from every evaluated corpus, and keep it fixed for all comparisons on that set. We set $\sigma$ using the median heuristic \citep{JMLR:v13:gretton12a}:
\begin{equation}
\sigma=\operatorname{median}\bigl\{\lVert h(r)-h(r')\rVert \;:\; r \neq r',\; r,r' \in R_{\mathrm{cal}}\bigr\}.
\end{equation}
The calibration split has 1{,}500 windows for the counterfactual evaluation set (Table~\ref{tab:main}) and 150 windows for the unconditional-generation reference (Table~\ref{tab:case-uncond}).
Since embedding scales differ across encoders, each backbone is calibrated separately.
On the counterfactual set and the generation reference respectively, this gives $\sigma=105.01$ and $105.38$ for Qwen3.5-27B, $77.49$ and $72.51$ for the distilled Qwen3.5-2B, and $77.16$ and $72.61$ for the distilled Qwen3.5-0.8B; Qwen3.5-9B gives $58.93$ on the counterfactual set.

\paragraph{Null standardization.}
We apply the same reference-based standardization to \system and every baseline. For distributional metrics, $s_M$ is the distance between the evaluated and reference corpora. For corpus-level scalar statistics such as perplexity and entropy, we use the absolute difference in the corpus statistic, $s_M=|\bar v_G-\bar v_R|$, with each statistic computed as specified in Appendix~\ref{app:baselines}. For MAUVE, we standardize the frontier integral of its divergence frontier rather than the MAUVE score, as detailed in Appendix~\ref{app:baselines}.

For each metric, we estimate a null distribution from $B=200$ draws. In each draw, we sample two disjoint subsets from the human reference pool, matching their sizes to those used in the actual evaluation, and compute the same discrepancy $s_M$ between them. The mean $\mu_0$ and standard deviation $\sigma_0$ of these null scores define
\begin{equation}
z_M=\frac{s_M-\mu_0}{\sigma_0}.
\end{equation}
This score expresses the observed discrepancy relative to the metric's reference sampling variation; values near zero are close to the null mean. It is not itself the criterion for detecting coherence damage, since benign rewriting can also produce a nonzero shift.

\paragraph{Paired-bootstrap detection.}
For each harmful perturbation condition, we compare its null-standardized score with that of the most extensively rewritten benign paraphrase condition constructed from the same seed samples. Their difference is $\Delta z=z_{\mathrm{harm}}-z_{\mathrm{benign}}$. To quantify uncertainty, we draw 200 bootstrap resamples of the seed samples with replacement. Within each draw $b$, we use the same resampled seed indices to form both the harmful and benign corpora and compute
\begin{equation}
\Delta z^{(b)}=z_{\mathrm{harm}}^{(b)}-z_{\mathrm{benign}}^{(b)}.
\end{equation}
The empirical 2.5th and 97.5th percentiles of these differences form the 95\% confidence interval. A condition is detected only when its lower bound is strictly positive, indicating a stronger response to coherence damage than to benign rewriting.

%% file: sec/appendix/C_evalset.tex
\section{Construction of the Counterfactual Evaluation Set}
\label{app:counterfactuals}
\label{app:evalset}

The perturbation categories and their diagnostic roles are defined in Section~\ref{sec:meta_eval}; here we specify how the counterfactual evaluation set is constructed for each perturbation type. Representative examples are shown in Table~\ref{tab:evalset-examples}.

\begin{table}[t]
\caption{\textbf{Representative examples from the counterfactual evaluation set.}
Only the targeted sentence or operation is modified; the surrounding text is preserved. Examples are shortened for readability.}
\label{tab:evalset-examples}
\begin{center}
\setlength{\tabcolsep}{5pt}
\small
\begin{tabular}{p{2.1cm} p{4.9cm} p{4.9cm}}
\toprule
Perturbation & Before & After \\
\midrule
Contradiction 
& ``\ldots became the worst kind of media free-for-all.'' 
& ``\ldots became the most respectful kind of media coverage.'' \\
\addlinespace
Causal reversal 
& ``You may opt out at any time.'' 
& ``You may opt in at any time.'' \\
\addlinespace
Broken transition 
& ``His suicide suddenly made more sense.'' 
& ``His suicide suddenly made more sense, but the moon is made of green cheese.'' \\
\addlinespace
Topic drift 
& ``\ldots spoke strongly on behalf of the virtues of physical education.'' 
& ``\ldots spoke strongly on behalf of the virtues of chess strategy.'' \\
\addlinespace
Sentence permutation
& clean text sample
& selected sentences are reordered within the sample \\
\addlinespace
Word shuffle
& clean text sample
& words are locally shuffled within selected sentences \\
\addlinespace
Repetition 
& clean text sample 
& selected sentences are duplicated in place \\
\addlinespace
DLM mix
& clean text sample 
& selected sentences are replaced by diffusion-LM sentences \\
\addlinespace
Document mix
& clean text sample
& selected sentences are replaced by sentences from unrelated documents \\
\midrule
Benign rewriting 
& ``I know what I did is wrong and horrible.'' 
& ``I realize my actions were incorrect and deeply wrong.'' \\
\bottomrule
\end{tabular}
\end{center}
\end{table}

\paragraph{Data pools.}
The sources, length filter, and three-way split into seed, reference, and replacement pools are described in Section~\ref{sec:exp-setup}. The replacement pool contains 2{,}453 text samples. The three sources (OpenWebText, WikiText-103, and Reddit TL;DR) contribute roughly equal numbers of seed samples, and all pools are deduplicated against each other using normalized-text hashes to prevent overlap.

\paragraph{LLM-edited perturbations.}
Relation perturbations (contradiction, causal reversal), discourse perturbations (broken transition, topic drift), and the benign control are produced by the Qwen3-30B-A3B \citep{qwen3} editor, which is distinct from the \system encoder. For each seed sample, a fixed seeded sampler selects the target sentences, and the editor rewrites each independently; all non-target sentences are copied back unchanged.

\textit{Contradiction} rewrites a target sentence so that it conflicts with a preserved anchor sentence in the same sample.
\textit{Causal reversal} reverses an expressed cause--effect relation while preserving the entities involved.
\textit{Broken transition} replaces a target sentence with a locally fluent sentence that is incompatible with the preceding discourse.
\textit{Topic drift} replaces a target sentence with an off-topic sentence of comparable length.
\textit{Benign paraphrase} rewrites target sentences while preserving claims, entities, numbers, stance, and discourse order.

\paragraph{Structural perturbations.}
These are rule-based transformations applied without LLM involvement. \textit{Sentence permutation} reorders a fraction of the sentences. \textit{Word shuffle} perturbs word order within selected sentences. \textit{Repetition} duplicates selected sentences in place. All three preserve most of the original lexical material while disrupting organization.

\paragraph{Mixture perturbations.}
\textit{DLM mix} replaces a fraction of the sentences with unconditional OpenWebText generations from a combined pool: three seeded runs from SEDD-small at 256 sampling steps \citep{sedd-2024} and three from MDLM-OWT at 512 sampling steps \citep{mdlm-2024}. \textit{Document mix} replaces sentences with material from one or more unrelated human documents in the replacement pool. A human-splice control applies the same replacement pattern using other human-written text, serving as an additional reference for mixture-induced shifts.

\paragraph{Perturbation severity levels.}
LLM-edited perturbations rewrite one, two, or three target sentences. Each edit introduces exactly one relation- or discourse-level violation, so the number of rewritten sentences directly controls the number of violations per sample. The benign control follows the same one-to-three schedule, so harmful and benign conditions are matched in the amount of edited text. Because every seed sample contains at least four sentences, rewriting at most three leaves most of the original text intact. Each perturbation is therefore a targeted edit, not a complete rewrite.

Structural and mixture perturbations use type-specific fraction schedules. Sentence permutation, word shuffle, and repetition affect $10\%$, $25\%$, $50\%$, $75\%$, or $100\%$ of eligible sentences. DLM mix and its matched human-splice control replace $10\%$, $25\%$, or $50\%$ of sentences. Document mix replaces $10\%$, $25\%$, $50\%$, or $75\%$ of sentences; an additional source-diversity sweep holds the replacement rate at $50\%$ and draws from one, two, or four source documents. For a sample with $m$ eligible sentences, a target fraction $r$ selects $\min(m,\max(1,\lceil mr\rceil))$ sentences. Sentence permutation additionally requires at least two selected sentences so that the operation changes their order. Thus, for example, a four-sentence sample changes one sentence at $25\%$ for shuffle, repetition, and mixture, but permutes two sentences. Score trajectories across severity levels are reported in Appendix~\ref{app:severity}.

\paragraph{Sample filtering and validation.}
Because LLM-generated edits can fail to produce the intended perturbation, we apply a post-hoc filtering step to ensure the validity of every edited sample. We discard samples with \emph{no effective edit}, where the editor returns the target sentence unchanged or with only trivial variation; \emph{severe truncation}, where the output is cut off or substantially shorter than the input; and \emph{missing required context}, where the seed text lacks the structure a perturbation presupposes, such as an explicit cause--effect relation for causal reversal. For contradiction, the anchor sentence must be retained in the final sample, since a contradiction is observable only when the original claim and the conflicting rewrite coexist; samples in which the anchor is edited are discarded.

We then validate the set at the condition level. Sample counts are checked to be balanced across conditions, so that detections are compared at comparable statistical power. The realized edit rate, measured as the fraction of sentences that differ from the seed text, is checked to increase monotonically with severity within each perturbation type, confirming that the nominal sentence schedule produces graded textual change. The same check applies to the benign paraphrase conditions, whose edit rates match those of the harmful conditions at each level.

The final evaluation set contains 43 conditions: 12 LLM-edited conditions (four types at three severities), 28 structural and mixture conditions, and three benign paraphrase severities; the most extensively rewritten one serves as the benign control in the detection criterion (Section~\ref{sec:meta_eval}).

\paragraph{Robustness to the editor model.}
\label{app:editor-robustness}
The default editor (Qwen3-30B-A3B; \citealp{qwen3}) and the default encoder of \system (Qwen3.5-27B) belong to the same model family, raising the concern that \system's sensitivity may partly reflect recognition of family-specific editing artifacts rather than coherence degradation itself. To rule this out, we regenerate all LLM-edited conditions with Mistral-Small-24B-Instruct-2501 \citep{mistral2025small3}, a model from an entirely different family, keeping the seed samples, target sentences, editing instructions, and severity levels fixed. Structural and mixture perturbations are unchanged.

Table~\ref{tab:editor-robustness} reports the null-standardized scores of the default \system encoder on the relation and discourse conditions, whose test texts differ between the two editors; the structural and mixture conditions are identical across the two sets and are therefore omitted. The main finding is preserved. \system detects 11 of the 12 LLM-edited conditions on the Mistral-edited set, compared with 10 of 12 under the default editor. Contradiction is detected at all severity levels under both editors. Causal reversal is the hardest perturbation type in the main experiments and remains the least consistently detected under both editors. All four perturbation types show increasing responses with severity, and \system does not confuse benign paraphrases with coherence degradation at any severity level. \system's selective sensitivity to coherence degradation is therefore not an artifact of the editor--encoder family overlap.

\begin{table}[t]
\caption{\textbf{Robustness to the editor model.}
Null-standardized \system scores $z_M$ at the highest severity of each LLM-edited perturbation type. Shaded entries are detected relative to the benign control under the paired bootstrap criterion (Section~\ref{sec:meta_eval}). The final column reports detections across the 12 LLM-edited conditions. Scores are calibrated separately for each set and are not comparable across rows.}
\label{tab:editor-robustness}
\begin{center}
\small
\begin{tabular}{lcccccc}
\toprule
& \multicolumn{2}{c}{Relation}
& \multicolumn{2}{c}{Discourse}
& & \\
\cmidrule(lr){2-3}\cmidrule(lr){4-5}
Editor model
& \makecell{Contra-\\diction}
& \makecell{Causal\\reversal}
& \makecell{Broken\\transition}
& \makecell{Topic\\drift}
& \cellcolor{BenignGray}\makecell{Benign\\paraphrase}
& \makecell{Detected\\(of 12)} \\
\midrule
Qwen3-30B-A3B (default)
& \detected{84.3}
& \detected{25.6}
& \detected{331}
& \detected{79.7}
& \benign{6.0}
& 10 \\
Mistral-Small-24B
& \detected{151}
& \detected{90.5}
& \detected{559}
& \detected{406}
& \benign{9.5}
& 11 \\
\bottomrule
\end{tabular}
\end{center}
\end{table}

%% file: sec/appendix/D_baselines.tex
\section{Baseline Metric Implementations}
\label{app:baselines}

The following baselines are compared with \system on the counterfactual evaluation set (Table~\ref{tab:main}) and on the outputs of existing generation models (Section~\ref{sec:exp-case}). All baselines are computed on the same corpora as \system and placed on the same scale via the null standardization described in Appendix~\ref{app:implementation}.

\paragraph{Evaluation text lengths.}
All metrics operate on text truncated or packed to the same fixed length within each experiment: 512 tokens for the counterfactual evaluation set and unconditional generation, 256 tokens for prefix continuation (128-token human prefix + 128-token model continuation), and 256 tokens for the human-judgment study of Appendix~\ref{app:human-eval}, matching the texts shown to annotators. Truncation is applied before feature extraction, so every metric sees identical text for a given sample. Fewer than 1.3\% of counterfactual samples exceed 512 tokens. For unconditional generation, all corpora are packed to exactly 512 tokens, except ELF-L, the large variant of ELF \citep{hu2026elfembeddedlanguageflows}, whose native outputs are about 930 tokens; it is evaluated on its first 512 tokens to match the other systems.

\paragraph{Generative perplexity.}
Following \citet{holtzman2020curiouscaseneuraltext}, we use GPT-2-large \citep{radford2019language} as an external autoregressive scorer. Each text sample $x$ is tokenized into $(x_1, \ldots, x_T)$ and truncated to the evaluation length above. The passage-level negative log-likelihood is
\begin{equation}
\mathrm{NLL}(x) = -\sum_{t=2}^{T} \log p_\theta\bigl(x_t \mid x_{<t}\bigr),
\end{equation}
where $T(x) = T - 1$ is the number of predicted tokens, since the first token has no conditioning context. For the generator evaluations of Section~\ref{sec:exp-case}, we report the corpus-level perplexity
\begin{equation}
\mathrm{PPL}(C)
=
\exp\left(
\frac{\sum_{x \in C} \mathrm{NLL}(x)}{\sum_{x \in C} T(x)}
\right),
\end{equation}
which pools tokens across all samples and weights each token equally regardless of sample length. On the counterfactual evaluation set, where all metrics pass through the same null standardization, we instead use the mean per-sample log-perplexity $\frac{1}{|C|}\sum_{x \in C} \mathrm{NLL}(x)/T(x)$, which weights each sample equally and matches the per-corpus form of the other scalar baselines.

\paragraph{Unigram entropy.}
We lowercase the entire corpus $C$ and split it into words by whitespace, count all occurrences $n_C(w)$ of each word type $w$, and estimate the empirical unigram distribution $p_C(w) = n_C(w) / N_C$, where $N_C = \sum_w n_C(w)$ is the total token count. The corpus-level unigram entropy \citep{shannon1948mathematical} is
\begin{equation}
H(C) = -\sum_{w} p_C(w) \log p_C(w).
\end{equation}
This statistic pools all tokens in the corpus into a single distribution and discards word order entirely, so it reflects only corpus-level lexical diversity.

\paragraph{MAUVE.}
MAUVE \citep{pillutla2021mauvemeasuringgapneural,mauve-theory-practice-2023} measures the gap between the generated and reference distributions in a discretized feature space. Each text sample is mapped to a feature vector, and the vectors from both corpora are jointly clustered into $K$ groups via $k$-means. Each corpus is then represented by its cluster-assignment histogram: $P$ for the generated corpus and $Q$ for the reference. For a mixture $M_\lambda = \lambda P + (1-\lambda) Q$, the divergence curve is formed by exponentiating the two KL divergences separately:
\begin{equation}
\mathcal{C}(P,Q) = \Bigl\{\bigl(e^{-c\,\mathrm{KL}(P\|M_\lambda)},\, e^{-c\,\mathrm{KL}(Q\|M_\lambda)}\bigr) : \lambda \in (0,1)\Bigr\}, \qquad
\mathrm{MAUVE}(G, R) = \mathrm{AUC}\bigl(\mathcal{C}(P,Q)\bigr).
\end{equation}
Here $c>0$ is a fixed scaling constant, and AUC denotes the area under the curve, completed with endpoints $(0,1)$ and $(1,0)$. The score lies in $[0,1]$; values near 1 indicate a close match. The exponential is applied to the curve coordinates before integration, not to a scalar area afterwards.

\emph{What we report.} The MAUVE score saturates near zero for large gaps, so null-standardizing it would obscure meaningful variation. The MAUVE rows of Table~\ref{tab:main}, and MAUVE's $k$-means KL in the representation--distance analysis of Section~\ref{sec:exp-attribution}, therefore report the frontier integral \citep{mauve-theory-practice-2023} returned by the official implementation: a KL-based summary of the same divergence frontier that is zero when $P=Q$, grows with the gap, and lies in $[0,1]$, reaching 1 only when the two histograms have disjoint support. It is computed directly from the two histograms rather than from the exponentiated curve, and we apply the reference-null standardization of Appendix~\ref{app:implementation} to it. The generator evaluations of Section~\ref{sec:exp-case}, the human-agreement study of Section~\ref{sec:exp-human}, and Appendix~\ref{app:mauve-length} report the standard MAUVE score. In all cases we use the official implementation with $K = n/10$ clusters, $c = 5$, and 25 values of $\lambda$; the MAUVE scores are averaged over five $k$-means seeds, whereas the frontier integral uses one $k$-means fit per bootstrap draw. In the representation--distance analysis, the Fr\'echet distance and $k$-means KL are computed in a 128-dimensional PCA subspace fitted on the human-reference features, because a full-dimensional Gaussian fit is rank-deficient at these corpus sizes; RBF-MMD and energy distance use the raw features.

We evaluate two feature spaces: GPT-2-large last-token hidden states, which is the default configuration of \citet{pillutla2021mauvemeasuringgapneural}, and ELECTRA \citep{clark2020electrapretrainingtextencoders} masked-mean pooled features. Appendix~\ref{app:mauve-length} examines sensitivity to truncation and packed-window length.

\paragraph{FBD.}
Fr\'echet-BERT Distance \citep{xiang2021assessing} represents each text sample as a feature vector using BERT \citep{devlin2019bertpretrainingdeepbidirectional}, then fits a multivariate Gaussian $\mathcal{N}(\boldsymbol{\mu}_C, \boldsymbol{\Sigma}_C)$ to each corpus $C$ by computing the feature mean $\boldsymbol{\mu}_C$ and unbiased covariance $\boldsymbol{\Sigma}_C$. FBD is the squared 2-Wasserstein distance between the two fitted Gaussians:
\begin{equation}
\mathrm{FBD}(G, R)
=
\underbrace{\lVert \boldsymbol{\mu}_G - \boldsymbol{\mu}_R \rVert^2}_{\text{mean shift}}
+
\underbrace{\mathrm{tr}\!\left(
\boldsymbol{\Sigma}_G + \boldsymbol{\Sigma}_R
- 2\bigl(\boldsymbol{\Sigma}_G^{1/2}\, \boldsymbol{\Sigma}_R\, \boldsymbol{\Sigma}_G^{1/2}\bigr)^{1/2}
\right)}_{\text{covariance mismatch}}.
\end{equation}
Because the Gaussian assumption captures only the mean and covariance of each corpus, any distributional difference that leaves these two statistics approximately unchanged is invisible to FBD.

\paragraph{MMD-MiniLM.}
Maximum Mean Discrepancy (MMD) \citep{JMLR:v13:gretton12a} is a kernel-based two-sample distance that compares distributions without assuming a parametric form for either. Following \citet{chan2024distribution}, we instantiate MMD with \texttt{all-MiniLM-L6-v2} sentence embeddings \citep{wang2020minilmdeepselfattentiondistillation,reimers2019sentencebertsentenceembeddingsusing} $\phi(\cdot)$ and the same biased RBF-MMD estimator used by \system (Section~\ref{sec:chord}). MiniLM is a general-purpose sentence encoder not designed for coherence, so comparing MMD-MiniLM with \system isolates the effect of the representation while holding the distance fixed.

The two metrics also differ in bandwidth selection. MMD-MiniLM recalibrates $\sigma$ for each comparison by taking the median pairwise distance over the joint embedding set $\Phi_G \cup \Phi_R$, where $\Phi_G = \{\phi(g_j)\}_{j=1}^{N_G}$ and $\Phi_R = \{\phi(r_i)\}_{i=1}^{N_R}$:
\begin{equation}
\sigma
=
\operatorname{median}\bigl\{
\lVert \phi(g) - \phi(r) \rVert
:\ \phi(g), \phi(r) \in \Phi_G \cup \Phi_R,\ \phi(g) \neq \phi(r)
\bigr\}.
\end{equation}
Because $\sigma$ changes with every pair of corpora, scores from different comparisons are not on the same kernel scale. \system avoids this issue by calibrating $\sigma$ once from a held-out reference split and fixing it across all evaluations (Appendix~\ref{app:implementation}). Section~\ref{sec:exp-attribution} controls representation and distance independently in the factorial ablation.

%% file: sec/appendix/E_modern_baselines.tex
\section{Modern Embedding and Evaluator Baselines}
\label{app:modern-baselines}

We compare \system with three instruction-tuned embedding models and two established text evaluators on the counterfactual evaluation set. All methods use the same reference splits, null calibration, and detection criterion as Table~\ref{tab:main}. 

\paragraph{Instruction-tuned embeddings.}
We evaluate Qwen3-Embedding-8B \citep{zhang2025qwen3embeddingadvancingtext}, e5-mistral-7b-instruct \citep{wang2024improvingtextembeddingslarge}, and gte-Qwen2-7B-instruct \citep{li2023generaltextembeddingsmultistage}. For each model, we extract the final-layer, last-token embedding, apply L2 normalization, and use the same RBF-MMD protocol as \system, with bandwidth fixed on the reference development split. We compare three input settings: no instruction, a similarity instruction (``Retrieve semantically similar text.''), and a coherence instruction (``Represent the passage in terms of its logical coherence and the order of its ideas.''). We use each model's native instruction template and preserve the final representation-eliciting token when truncating inputs to 512 tokens.

\paragraph{UniEval and BARTScore.}
Because our evaluation set contains individual passages without paired sources or references, we adapt both evaluators to score text alone. For UniEval \citep{zhong2022unifiedmultidimensionalevaluatortext}, we compute fluency using the released sentence-level scoring procedure with \texttt{unieval-sum}. We evaluate coherence with \texttt{unieval-intermediate}, using the zero-shot question ``Is this a coherent and logically consistent passage?'' Both dimensions use $P(\mathrm{Yes})/[P(\mathrm{Yes})+P(\mathrm{No})]$, and we also report their average. For BARTScore \citep{yuan2021bartscoreevaluatinggeneratedtext}, we compute length-normalized log-likelihood under \texttt{bart-large-cnn} \citep{lewis2020bart} with an empty source. For each scalar evaluator, the corpus-level statistic is the absolute difference between the candidate and reference mean scores, standardized against its clean-reference null. These source-free adaptations differ from the evaluators' original source-conditioned settings.

\begin{table}[t]
\caption{\textbf{Additional baselines on the counterfactual evaluation set.}
Entries are null-standardized responses at the highest perturbation severity; blue shading indicates detection relative to benign rewriting, as in Table~\ref{tab:main}. Embedding methods use RBF-MMD, with the input instruction shown in parentheses. Scalar evaluators use the absolute candidate--reference difference in mean score.}
\label{tab:modern-baselines}
\centering
\setlength{\tabcolsep}{3pt}
\resizebox{\textwidth}{!}{%
\begin{tabular}{lcccccccccc}
\toprule
Method & \makecell{Causal\\reversal} & \makecell{Contra-\\diction}
& \makecell{Broken\\transition} & \makecell{Topic\\drift}
& \makecell{Sentence\\permutation} & \makecell{Word\\shuffle}
& Repetition & DLM mix & Document mix & \benign{\makecell{Benign\\rewriting}} \\
\midrule
\multicolumn{11}{l}{\emph{Instruction-tuned embeddings + RBF-MMD}} \\
Qwen3-Embedding-8B (none) & 12.1 & 11.5 & \detected{19.3} & \detected{17.3} & \detected{14.5} & 12.0 & \detected{15.1} & \detected{129} & \detected{142} & \benign{7.6} \\
Qwen3-Embedding-8B (similarity) & 9.3 & \detected{13.0} & \detected{22.5} & \detected{14.8} & \detected{10.7} & 8.4 & \detected{19.0} & \detected{76.2} & \detected{73.5} & \benign{5.0} \\
Qwen3-Embedding-8B (coherence) & 10.6 & \detected{11.5} & \detected{51.4} & \detected{19.7} & \detected{13.2} & \detected{12.0} & \detected{19.9} & \detected{157} & \detected{182} & \benign{7.4} \\
e5-mistral-7b-instruct (none) & 8.3 & 7.7 & \detected{30.3} & \detected{12.8} & \detected{13.2} & \detected{22.0} & \detected{102} & \detected{87.6} & \detected{92.1} & \benign{5.6} \\
e5-mistral-7b-instruct (similarity) & 8.3 & \detected{10.9} & \detected{44.4} & \detected{13.5} & \detected{12.8} & \detected{20.7} & \detected{69.6} & \detected{92.1} & \detected{92.6} & \benign{5.9} \\
e5-mistral-7b-instruct (coherence) & 6.3 & 7.7 & \detected{46.5} & \detected{10.3} & \detected{10.5} & \detected{12.6} & \detected{75.1} & \detected{105} & \detected{98.5} & \benign{4.8} \\
gte-Qwen2-7B-instruct (none) & 7.4 & 7.1 & \detected{21.6} & \detected{11.2} & \detected{11.1} & \detected{12.8} & \detected{13.7} & \detected{101} & \detected{99.2} & \benign{6.2} \\
gte-Qwen2-7B-instruct (similarity) & 9.7 & \detected{10.9} & \detected{36.0} & \detected{15.6} & \detected{11.3} & \detected{11.0} & \detected{21.7} & \detected{109} & \detected{106} & \benign{5.7} \\
gte-Qwen2-7B-instruct (coherence) & 10.4 & \detected{11.3} & \detected{65.1} & \detected{20.8} & \detected{23.8} & \detected{21.1} & \detected{74.0} & \detected{185} & \detected{228} & \benign{7.2} \\
\midrule
\multicolumn{11}{l}{\emph{Scalar evaluators (source-free)}} \\
UniEval (fluency) & -1.1 & -0.8 & -0.3 & 0.3 & 2.4 & \detected{99.4} & 1.4 & \detected{22.1} & -1.0 & \benign{3.5} \\
UniEval (coherence, zero-shot) & -0.2 & 1.8 & 4.3 & \detected{11.2} & \detected{6.5} & \detected{41.1} & \detected{23.2} & \detected{47.9} & \detected{21.1} & \benign{1.1} \\
UniEval (fluency + coherence) & 0.2 & 0.6 & 2.6 & 8.1 & \detected{9.3} & \detected{105} & \detected{16.2} & \detected{53.3} & \detected{16.1} & \benign{3.9} \\
BARTScore (hypothesis-only) & -0.4 & -0.4 & -0.3 & -0.4 & \detected{3.0} & \detected{34.1} & \detected{30.3} & \detected{29.5} & \detected{20.3} & \benign{-0.6} \\
\midrule
\textbf{\system} (Qwen3.5-27B) & \detected{25.6} & \detected{84.3} & \detected{331} & \detected{79.7} & \detected{358} & \detected{740} & \detected{857} & \detected{1150} & \detected{1082} & \benign{6.0} \\
\bottomrule
\end{tabular}}
\end{table}

\paragraph{Results.}
Table~\ref{tab:modern-baselines} shows that modern embedding models detect both discourse perturbations and most structural and mixture perturbations at the highest severity. Relation-level errors remain harder: none detects causal reversal, and contradiction is detected only under some instruction settings. The coherence instruction increases the response to broken transitions for all three models, but does not consistently improve relation-level detection over the similarity instruction. UniEval and hypothesis-only BARTScore also miss both relation-level perturbations, responding mainly to structural and mixture errors. In contrast, \system detects all nine perturbation types. These results extend the comparison beyond older encoders, but do not isolate the effect of coherence prompting: the embedding models also differ from \system in size and training. Section~\ref{sec:backbone-elicitation} examines capacity and prompting through controlled ablations.

%% file: sec/appendix/F_llm_judge.tex
\section{Hidden-State Signals versus Token-Space Judgments}
\label{app:llm-judge}
\label{app:hidden-vs-token}

\system compares distributions of hidden representations.
We examine how much of this coherence signal is captured by token
counts and next-token predictions, then compare with direct LLM
judging and G-Eval \citep{liu2023gevalnlgevaluationusing} using the same frozen Qwen3.5-27B backbone.

\subsection{Token Counts and Next-Token Predictions}

\begin{figure}[h]
\centering
\includegraphics[width=0.52\textwidth]{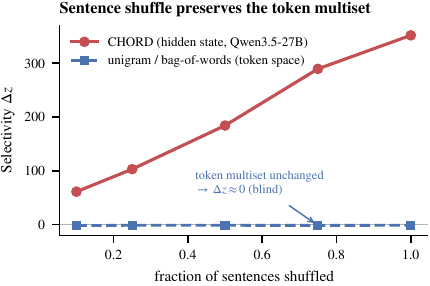}
\caption{\textbf{Hidden representations versus token counts.}
Under sentence shuffling, unigram selectivity stays near zero,
while hidden-state selectivity increases with the shuffle rate.
Both methods use the same RBF-MMD, human-reference null,
and benign-paraphrase control.}
\label{fig:hidden-vs-token}
\end{figure}

Sentence shuffling changes discourse order without changing each
document's token counts, leaving an exact unigram representation
unchanged. Figure~\ref{fig:hidden-vs-token} compares this representation
with the coherence-prompted hidden representation under the same
RBF-MMD protocol. Unigram selectivity remains near zero, while
hidden-state selectivity increases with the shuffle rate.
Token counts thus miss order-dependent coherence failures that
the hidden representation captures.

\begin{figure}[h]
\centering
\includegraphics[width=0.8\textwidth]{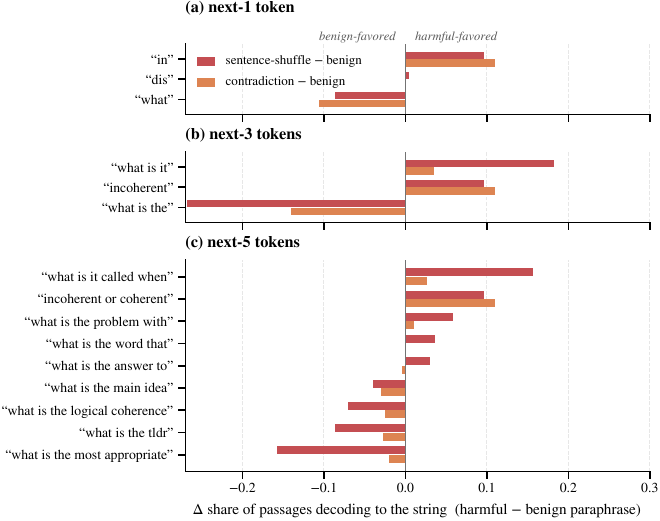}
\caption{\textbf{Next-token predictions at the \system representation-eliciting position.}
Bars show changes in the fraction of passages producing each
decoded string, relative to benign paraphrases. Coherence errors
increase negative descriptions, but the outputs do not consistently
express a clear coherence judgment.}
\label{fig:token-decode}
\end{figure}

We next examine the model's next-token predictions at the prompt
position where \system reads the hidden state, using the same
forward pass. Figure~\ref{fig:token-decode} shows how the decoded
outputs change relative to benign paraphrases. Coherence errors
shift the outputs toward the negative description ``incoherent,''
indicating that the predictions contain some coherence information.
However, the outputs do not consistently express a clear judgment:
the added ``incoherent'' continues as ``incoherent or coherent,''
and most of the remaining shift is toward continuations of the
prompt, such as ``what is it called when.'' The hidden representation at this
position provides stronger corpus-level separation.

These probes examine token counts and predictions under the
representation prompt. We next evaluate explicit coherence ratings
using prompts designed for judging.

\subsection{Same-Backbone Comparison with LLM Judges}

The judges evaluated here assign a coherence score to each
passage, whereas \system measures the discrepancy between
two corpora through distributions of hidden representations.
\system is therefore \emph{coherence-aware} instead of
\emph{coherence-only}: its representations emphasize coherence
but can retain other textual properties. Therefore, the aim of 
\system is very different from LLM-as-judge.

Nevertheless, we conduct an experiment comparing \system with a direct coherence judge
\citep{zheng2023judging} and G-Eval, aggregating their
per-passage ratings for corpus-level evaluation.
All three use the same frozen Qwen3.5-27B backbone.
This comparison tests how well each approach captures
corpus-level differences, with particular attention to
coherence errors.

\paragraph{Direct coherence judge.}
The direct judge assigns each passage an integer score from
0 to 10 using the prompt in Table~\ref{tab:judge-prompt}.
The prompt separates the passage from the instructions with
\texttt{<document>} tags and asks the model to ignore topic,
genre, and truncation at the end of the passage. It requests
the score before a brief explanation, so the explanation
does not precede and influence the rating. We run the judge
with vLLM \citep{kwon2023efficientmemorymanagementlarge}
at temperature 0 with a fixed seed.

\begin{table}[ht]
\caption{\textbf{Direct coherence judge prompt.} The same prompt is used for all passages, without a system prompt or few-shot examples.}
\label{tab:judge-prompt}
\begin{center}
\small
\begin{tabular}{p{12.5cm}}
\toprule
You are a strict, careful writing-quality rater. Read the document between the \texttt{<document>} markers and rate its logical consistency, discourse flow, and overall quality on an integer scale from 0 (incoherent, broken, or self-contradictory) to 10 (flawless, fully coherent writing). Judge only the writing itself; do not reward or penalize the topic, opinions, or genre, and ignore truncation at the very end of the document. \\[2pt]
\texttt{<document>} \\
\texttt{\{text\}} \\
\texttt{</document>} \\[2pt]
Return only JSON, the score first: \texttt{\{"score": <integer 0-10>, "reason": "<one short sentence>"\}} \\
\bottomrule
\end{tabular}
\end{center}
\end{table}

\paragraph{G-Eval.}
We adapt the released SummEval \citep{fabbri2021summeval} coherence rubric from G-Eval \citep{liu2023gevalnlgevaluationusing} to standalone passages. The evaluation steps are generated once and then fixed for all passages. Each passage receives a probability-weighted score over ratings 1--5, computed from the rating-token log
probabilities. We use Qwen3.5-27B to evaluate the G-Eval procedure under the same backbone as \system; this is an adaptation of G-Eval, whose original evaluator used GPT-4 \citep{openai2023gpt4}.

\paragraph{Corpus-level comparison.}
For each judge, we compute the absolute difference between the candidate and reference mean scores, $|\bar r_G-\bar r_R|$. We then apply the same null
standardization and benign-contrast detection criterion as in Section~\ref{sec:meta_eval}. Both judges share \system's sample sizes, reference splits, and resampling procedure.

\paragraph{Counterfactual evaluation.}
Table~\ref{tab:judge-counterfactual} shows that both judges detect all nine perturbation types at the highest severity.
Explicit ratings therefore capture coherence information when the model is prompted to evaluate it. However, \system produces much larger null-standardized responses across all nine types.
For contradiction, the direct judge and G-Eval yield $z_M=16.2$ and $9.9$, compared with $84.3$ for \system.
With the backbone held fixed, the hidden-state comparison thus provides greater separation relative to its clean-reference variation. 

\begin{table}[!htb]
\caption{\textbf{Same-backbone comparison with LLM judges.}
All methods use frozen Qwen3.5-27B. Entries are null-standardized responses at the highest perturbation severity; shaded entries indicate detection relative to benign rewriting. For each judge, the statistic is the absolute difference
between candidate and reference mean ratings.}
\label{tab:judge-counterfactual}
\begin{center}
\setlength{\tabcolsep}{3pt}
\resizebox{\textwidth}{!}{%
\small
\begin{tabular}{lcccccccccc}
\toprule
& \multicolumn{2}{c}{Relation}
& \multicolumn{2}{c}{Discourse}
& \multicolumn{3}{c}{Structural}
& \multicolumn{2}{c}{Mixture}
& \cellcolor{BenignGray}Control \\
\cmidrule(lr){2-3}\cmidrule(lr){4-5}\cmidrule(lr){6-8}
\cmidrule(lr){9-10}\cmidrule(lr){11-11}
Method
& \makecell{Contra-\\diction}
& \makecell{Causal\\reversal}
& \makecell{Broken\\transition}
& \makecell{Topic\\drift}
& \makecell{Sentence\\permutation}
& \makecell{Word\\shuffle}
& Repetition
& DLM mix
& Document mix
& \cellcolor{BenignGray}\makecell{Benign\\rewriting} \\
\midrule
LLM judge (same Qwen3.5-27B)
& \detected{16.2} & \detected{9.5} & \detected{26.0}
& \detected{16.9} & \detected{28.1} & \detected{40.0}
& \detected{42.6} & \detected{49.0} & \detected{53.0}
& \benign{0.3} \\
G-Eval (same Qwen3.5-27B)
& \detected{9.9} & \detected{5.2} & \detected{33.9}
& \detected{18.8} & \detected{41.4} & \detected{49.2}
& \detected{62.2} & \detected{75.4} & \detected{77.6}
& \benign{0.6} \\
\textbf{\system} (Qwen3.5-27B)
& \detected{84.3} & \detected{25.6} & \detected{331}
& \detected{79.7} & \detected{358} & \detected{740}
& \detected{857} & \detected{1150} & \detected{1082}
& \benign{6.0} \\
\bottomrule
\end{tabular}}
\end{center}
\end{table}

\begin{table}[h]
\caption{\textbf{Direct coherence judging on unconditional generation.}
Both methods use Qwen3.5-27B on the ten folds of Table~\ref{tab:case-uncond}; the \system column is the Qwen3.5-27B column of that table.
The judge mean is the mean rating over all 5{,}000 texts of a generator. Judge $z_M$ standardizes the absolute difference between the generator and reference fold means against a null of two disjoint 500-text reference subsets, reported as mean$\pm$std over folds.
Rows are ordered by \system rank. Rank markers (\#$k$) order the corpora by higher mean rating or lower \system score; judge $z_M$ is not ranked.
(n.s.) denotes a deviation that does not exceed the 95th percentile of the null.}
\label{tab:judge-casestudy}
\begin{center}
\small
\begin{tabular}{lcccc}
\toprule
Generator & Params & \makecell{Judge mean\\(0--10) $\uparrow$} & Judge $z_M$ & \makecell{\system $\downarrow$\\(MMD$^2$ $\times 10^{-2}$)} \\
\midrule
Human (packed, held-out) & -- & 3.96\crank{1} & $-$0.1\,{\scriptsize($\pm$0.6)} (n.s.) & 0.17\crank{1}\,{\scriptsize($\pm$0.04)} \\
GPT2-large (AR) & 774M & 2.02\crank{2} & 19.9\,{\scriptsize($\pm$1.5)} & 19.84\crank{2}\,{\scriptsize($\pm$1.02)} \\
GPT2-medium (AR) & 355M & 1.67\crank{3} & 23.7\,{\scriptsize($\pm$1.3)} & 28.37\crank{3}\,{\scriptsize($\pm$0.90)} \\
ELF-L & 652M & 1.08\crank{4} & 30.1\,{\scriptsize($\pm$1.3)} & 51.19\crank{4}\,{\scriptsize($\pm$1.09)} \\
LangFlow & 171M & 0.52\crank{7} & 36.3\,{\scriptsize($\pm$1.3)} & 52.58\crank{5}\,{\scriptsize($\pm$1.12)} \\
SEDD-small & 170M & 0.85\crank{5} & 32.7\,{\scriptsize($\pm$1.3)} & 57.05\crank{6}\,{\scriptsize($\pm$1.02)} \\
MDLM & 170M & 0.83\crank{6} & 32.9\,{\scriptsize($\pm$1.3)} & 57.97\crank{7}\,{\scriptsize($\pm$1.20)} \\
\bottomrule
\end{tabular}
\end{center}
\end{table}

\paragraph{Evaluating generation systems.}
Table~\ref{tab:judge-casestudy} compares the direct judge
with \system on the unconditional-generation folds of
Table~\ref{tab:case-uncond}. The judge rates every text of
those folds, cut to its first 512 tokens like the other
metrics. We report both the judge's mean rating and its
null-standardized deviation from the human reference, which
consists of packed 512-token sequences.

LLM-as-a-judge agrees with \system's broad ordering: human text,
followed by GPT-2 outputs, then diffusion and flow outputs.
Within the diffusion and flow group, however, the rankings diverge
more markedly: \system ranks LangFlow ahead of SEDD and MDLM,
whereas the judge ranks it last.

\paragraph{Takeaway.}
Token counts miss order-dependent errors, while next-token predictions under the representation prompt provide inconsistent coherence judgments. Dedicated judging prompts are more effective: both judges detect all nine perturbation
types at the highest severity. Even with the same backbone, however, \system yields larger null-standardized responses.
On unconditional generation, LLM-as-a-judge agrees with \system's broad grouping of human text, autoregressive outputs, and diffusion/flow outputs, but their rankings diverge more markedly within the diffusion/flow group.

%% file: sec/appendix/G_attribution.tex
\section{Representation Extraction and Prompt-Design Ablations}
\label{app:prompt-ablation}

The analyses below complement Section~\ref{sec:exp-attribution} by varying the representation extraction method, the \emph{target attribute} named in the instruction, and the \emph{prompt format} used to express that property. We also test sensitivity to prompt wording.

\paragraph{Backbone and representation extraction.}
We test whether coherence prompting still helps when Qwen3.5 is replaced by GPT-2, the backbone family used by MAUVE. For each backbone, we compare the \system coherence prompt, a generic PromptEOL prompt, the unprompted last token, and unprompted mean pooling. All settings use the third-to-last layer, a 512-token budget, RBF-MMD, the same 500-seed evaluation set, and the same bootstrap detection rule. Because each encoder has its own bandwidth and null scale, mean $\Delta z$ should be compared only among extraction methods for the same backbone; the condition counts are comparable across backbones.

\begin{table}[ht]
\centering
\caption{\textbf{How the extraction method changes coherence selectivity.}
Each row summarizes 40 harmful conditions. ``Detected'' counts conditions whose selectivity is reliably positive; ``reversed'' counts conditions whose response is reliably stronger for benign rewriting than for coherence damage. Bold marks the strongest extraction method for each backbone.}
\label{tab:gpt2-extraction-grid}
\footnotesize
\setlength{\tabcolsep}{5pt}
\begin{tabular}{llrrr}
\toprule
Backbone & Extraction method & \makecell{Mean\\$\Delta z$} & \makecell{Detected\\(of 40)} & \makecell{Reversed\\(of 40)} \\
\midrule
GPT-2 (0.124B) & Coherence prompt & 33.0 & 11 & 18 \\
 & Generic prompt & 7.8 & 10 & 20 \\
 & Raw last token & 27.2 & 14 & 11 \\
 & \textbf{Mean pooling} & \textbf{76.2} & \textbf{24} & \textbf{0} \\
\addlinespace[2pt]
GPT-2-medium (0.355B) & Coherence prompt & 26.2 & 9 & 15 \\
 & Generic prompt & 31.6 & 10 & 14 \\
 & Raw last token & 13.2 & 10 & 15 \\
 & \textbf{Mean pooling} & \textbf{87.3} & \textbf{24} & \textbf{0} \\
\addlinespace[2pt]
GPT-2-large (0.774B) & Coherence prompt & $-$12.5 & 8 & 19 \\
 & Generic prompt & $-$1.5 & 8 & 19 \\
 & Raw last token & 0.8 & 8 & 17 \\
 & \textbf{Mean pooling} & \textbf{40.3} & \textbf{24} & \textbf{0} \\
\addlinespace[2pt]
GPT-2-XL (1.558B) & Coherence prompt & $-$80.6 & 2 & 27 \\
 & Generic prompt & $-$64.3 & 2 & 27 \\
 & Raw last token & $-$78.8 & 2 & 28 \\
 & \textbf{Mean pooling} & \textbf{39.6} & \textbf{18} & \textbf{0} \\
\addlinespace[2pt]
Qwen3.5-27B & \textbf{Coherence prompt} & \textbf{471.0} & \textbf{38} & \textbf{0} \\
 & Generic prompt & 84.7 & 30 & 0 \\
 & Raw last token & 66.3 & 29 & 0 \\
 & Mean pooling & 38.6 & 26 & 0 \\
\bottomrule
\end{tabular}
\end{table}

Table~\ref{tab:gpt2-extraction-grid} shows that the best extraction method depends on whether the backbone can act on the prompt. Coherence prompting is strongest for Qwen3.5-27B, separating 38 of 40 harmful conditions. For every GPT-2 size, mean pooling is the strongest option, whereas final-token extraction often reverses the intended comparison and responds more to benign rewriting than to coherence damage. Increasing GPT-2 size does not resolve this problem: GPT-2-XL detects only 2 of 40 conditions with the coherence prompt, despite having more parameters than Qwen3.5-0.8B, which detects 20. Thus both the targeted prompt and a backbone that can interpret it are necessary. This experiment does not determine whether instruction tuning, training data, architecture, or another family difference supplies that capability.

\begin{figure}[ht]
\centering
\includegraphics[width=0.84\linewidth]{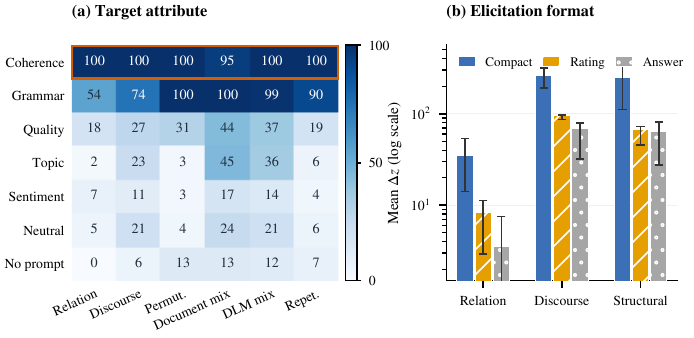}
\caption{\textbf{Target attribute and prompt format.}
\textbf{(a)}~Selectivity for compact prompts naming different target attributes, normalized within each perturbation category.
\textbf{(b)}~Selectivity for three prompt formats; error bars span three prompt wordings.}
\label{fig:prompt-ablation}
\end{figure}

\paragraph{Effect of the target attribute.}
We change only the property named in the prompt while keeping all other factors fixed. Figure~\ref{fig:prompt-ablation}a reports selectivity normalized within each perturbation category so that cross-category difficulty differences do not dominate the comparison.

Coherence gives the strongest selectivity for relation and discourse failures and remains at or near the top for structural and mixture perturbations. Grammar performs comparably on the latter two but is substantially weaker on relation and discourse, suggesting that it mainly captures syntax and not logical relations between sentences. Broader attributes such as quality, topic, and sentiment are much weaker across all categories. The neutral PromptEOL template and the unprompted representation are weaker still, showing that simply adding a prompt is not sufficient: the target attribute must name a property relevant to the intended distinction.

\paragraph{Effect of the prompt format and wording.}
We fix the target attribute to coherence and vary the prompt format. Figure~\ref{fig:prompt-ablation}b compares compact completion, scalar rating, and direct answer across three coherence wordings. Compact completion yields the strongest selectivity for relation, discourse, and structural failures, with the same advantage across all three wordings.

Wording affects response magnitude but not system ordering. We repeat the human-agreement evaluation in Appendix~\ref{app:human-eval} for each of the three compact coherence wordings, using the study released by \citet{pillutla2021mauvemeasuringgapneural}. For every wording, we recompute \system scores for the same eight GPT-2 generation settings and correlate their ranking with the Bradley--Terry rankings fitted from the study's 3{,}240 pairwise judgments. All three wordings produce the same correlations: $\rho = 0.857$, $0.976$, and $0.976$ for \emph{interesting}, \emph{makes sense}, and \emph{human-like}, respectively. The model ranking is therefore stable across these prompt paraphrases, even though their counterfactual response magnitudes differ. Tables~\ref{tab:attribute-prompts} and~\ref{tab:format-prompts} list the templates.

\begin{table}[ht]
\centering
\caption{\textbf{Target-attribute templates.}
The compact-completion format is fixed; only the target attribute changes.}
\label{tab:attribute-prompts}
\footnotesize
\setlength{\tabcolsep}{3pt}
\begin{tabular}{p{2.7cm}p{9.5cm}}
\toprule
Attribute & Template \\
\midrule
Coherence &
This passage: ``$x$'', considering its coherence and ordering of its ideas, means in one word: \\[2pt]
Grammar &
This passage: ``$x$'', considering its grammatical correctness and sentence structure, means in one word: \\[2pt]
Quality &
This passage: ``$x$'', considering its overall writing quality, means in one word: \\[2pt]
Topic &
This passage: ``$x$'', considering its main topic and subject matter, means in one word: \\[2pt]
Sentiment &
This passage: ``$x$'', considering its overall sentiment and emotional tone, means in one word: \\[2pt]
Neutral &
This sentence: ``$x$'' means in one word: (PromptEOL template; \citealp{prompteol-2024}) \\[2pt]
No prompt &
Bare passage; final-token or mean-pooled hidden states. \\
\bottomrule
\end{tabular}
\end{table}

\begin{table}[ht]
\centering
\caption{\textbf{Coherence-wording and prompt-format templates.}
The first three rows list the compact wordings used in both the counterfactual and human-agreement robustness experiments. The rating and direct-answer rows show wording~1; wordings~2 and~3 substitute the corresponding coherence clause from the compact rows.}
\label{tab:format-prompts}
\footnotesize
\setlength{\tabcolsep}{3pt}
\begin{tabular}{p{2.7cm}p{9.5cm}}
\toprule
Format & Template \\
\midrule
Compact, wording 1 &
This passage: ``$x$'', in terms of its logical coherence and the order of its ideas, means in one word: \\[2pt]
Compact, wording 2 &
This passage: ``$x$'', considering whether its ideas form a logically connected and well-ordered whole, means in one word: \\[2pt]
Compact, wording 3 &
This passage: ``$x$'', considering the consistency, organization, and logical flow of its ideas, means in one word: \\[2pt]
Scalar rating (wording 1) &
This passage: ``$x$''. Considering its logical coherence and the order of its ideas, rate it from 1 to 5: \\[2pt]
Direct answer (wording 1) &
This passage: ``$x$''. Considering its logical coherence and the order of its ideas, provide a yes or no judgment: \\
\bottomrule
\end{tabular}
\end{table}

%% file: sec/appendix/H_severity.tex
\section{Score Behavior across Corpus Size and Perturbation Severity}
\label{app:severity}

\begin{figure}[ht]
\centering
\includegraphics[width=0.52\linewidth]{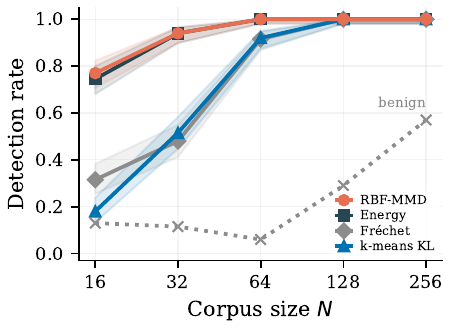}
\caption{\textbf{Score separation versus corpus size.}
On anchored contradiction, kernel distances separate harmful from null at smaller $N$ than Fr\'echet distance or $k$-means KL. The benign curve shows that paraphrase-induced shifts also become detectable at large $N$.}
\label{fig:power}
\end{figure}

\paragraph{Effect of corpus size.}
Figure~\ref{fig:power} compares how quickly different distances separate harmful conditions from the null as corpus size $N$ grows. Kernel distances (RBF-MMD, energy distance \citep{szekely2013energy}) reach reliable separation at smaller $N$ than Fr\'echet distance \citep{fid-ttur-2017} or $k$-means KL, providing additional justification for adopting RBF-MMD as the default distance. Crucially, the benign paraphrase curve also rises with $N$: at large corpus sizes, even the small shift caused by meaning-preserving rewriting becomes statistically detectable against the human-reference null. The detection criterion must therefore compare harmful and benign responses ($\Delta z=z_{\mathrm{harm}}-z_{\mathrm{benign}}>0$). Comparing the harmful response with zero would eventually flag benign rewriting as a coherence failure.

\begin{figure}[ht]
\centering
\includegraphics[width=\linewidth]{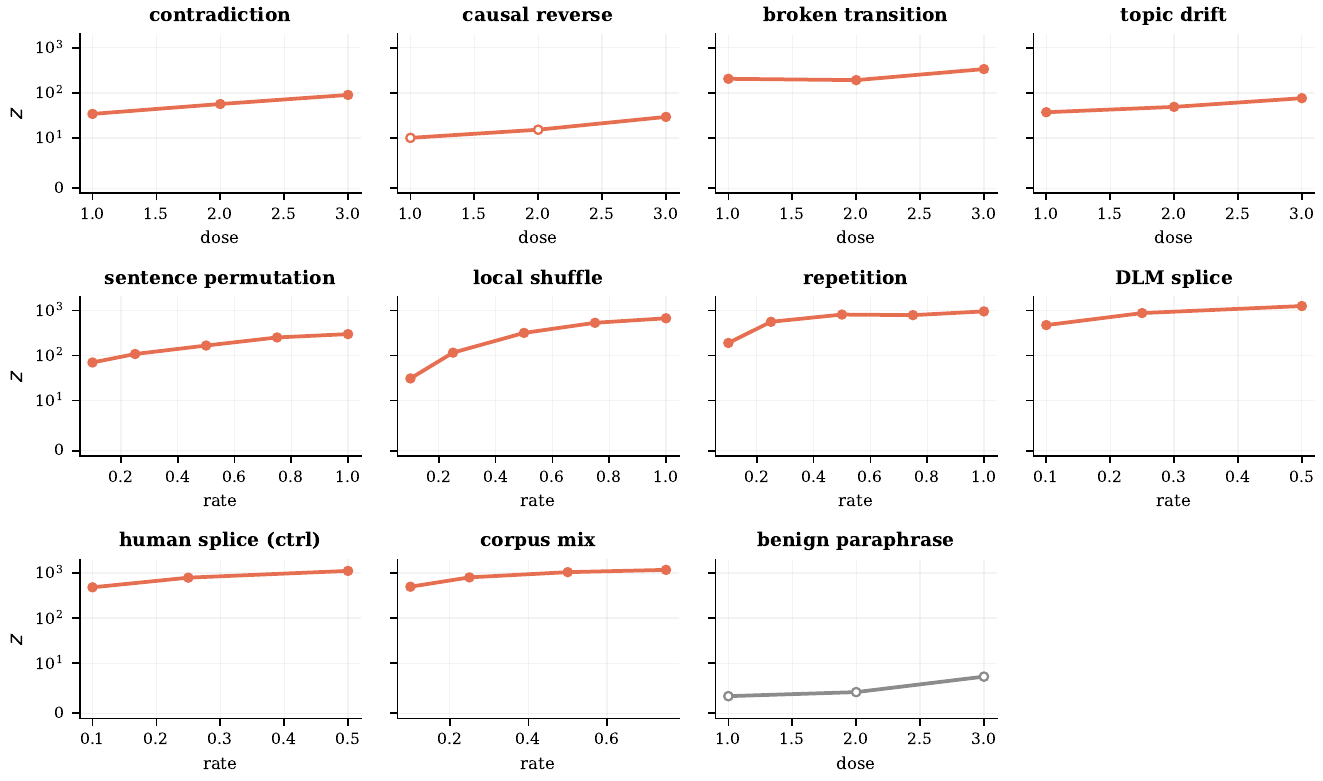}
\caption{\textbf{\system responses across perturbation severities.}
Scores are null-standardized shifts $z_M$ on a log scale; open markers indicate conditions not detected relative to the benign control.}
\label{fig:severity}
\end{figure}

\paragraph{Score trajectories across severity levels.}
Figure~\ref{fig:severity} shows that \system's z-scores increase monotonically with perturbation severity across all categories, confirming that \system quantifies the degree of coherence damage, not just its presence. The slope varies by category: structural and mixture perturbations produce steep increases, while relation-level perturbations rise more gradually, with causal reversal at low severity remaining the hardest condition. Benign paraphrases stay close to the baseline at all severity levels, confirming that \system does not confuse editing extent with coherence damage.

The human-splice condition produces a large z-score despite using only sentences from other human-written documents. This confirms that \system detects disruptions to inter-sentence coherence instead of poor individual sentences: even well-written human sentences break the coherence of a text sample when inserted out of their original context. Table~\ref{tab:severity-detail} reports the full endpoint scores and detection decisions for all metrics.

\begin{table}[htp]
\caption{\textbf{Results across perturbation severities}, underlying Table~\ref{tab:main}.
Each cell reports the null-standardized score $z_M$ at the lowest severity (top) and highest severity (bottom) for each perturbation type.
All rows use the same calibration as Table~\ref{tab:main}.
A $^{*}$ marks a condition detected relative to the benign control.
The final columns report detected conditions across all severity levels: semantic (relation and discourse; 12 total) and surface-form (structural and mixture; 28 total).}
\label{tab:severity-detail}
\begin{center}
\setlength{\tabcolsep}{2.5pt}
\resizebox{\textwidth}{!}{%
\footnotesize
\begin{tabular}{lcccccccccccc}
\toprule
& \multicolumn{2}{c}{Relation} & \multicolumn{2}{c}{Discourse} & \multicolumn{3}{c}{Structural} & \multicolumn{2}{c}{Mixture} & Control & \multicolumn{2}{c}{\#Detected} \\
\cmidrule(lr){2-3}\cmidrule(lr){4-5}\cmidrule(lr){6-8}\cmidrule(lr){9-10}\cmidrule(lr){11-11}\cmidrule(lr){12-13}
Method & Contr. & Causal & Broken & Topic & Perm. & Shuf. & Rep. & DLM mix & Doc.\ mix & Benign & \makecell{Sem.\\(12)} & \makecell{Form\\(28)} \\
\midrule
\multicolumn{13}{l}{\emph{Likelihood / diversity statistics}} \\
gen-PPL (GPT-2) &\makecell{0.3\\1.1} &\makecell{$-$0.1\\0.8} &\makecell{0.4\\0.5} &\makecell{1.3\\1.8} &\makecell{1.9\\7.4$^{*}$} &\makecell{6.2$^{*}$\\46.9$^{*}$} &\makecell{9.7$^{*}$\\62.4$^{*}$} &\makecell{18.5$^{*}$\\62.4$^{*}$} &\makecell{10.7$^{*}$\\35.9$^{*}$} &\makecell{0.2\\1.4} & 0/12 & 26/28 \\
Unigram entropy &\makecell{0.1\\0.6} &\makecell{$-$0.2\\$-$0.1} &\makecell{3.7\\6.0} &\makecell{$-$0.3\\0.0} &\makecell{0.3\\$-$0.0} &\makecell{0.5\\2.1} &\makecell{0.0\\0.9} &\makecell{6.9$^{*}$\\22.1$^{*}$} &\makecell{7.5$^{*}$\\17.4$^{*}$} &\makecell{0.7\\2.6} & 0/12 & 11/28 \\
\midrule
\multicolumn{13}{l}{\emph{Distributional metrics}} \\
MAUVE (GPT-2) &\makecell{32.2\\27.6} &\makecell{17.4\\23.1} &\makecell{29.1\\21.0} &\makecell{24.1\\18.9} &\makecell{1.1\\24.5} &\makecell{$-$0.3\\1.1} &\makecell{13.0\\56.5$^{*}$} &\makecell{14.7\\35.8$^{*}$} &\makecell{9.4\\44.9$^{*}$} & \textbf{\makecell{27.8\\27.3}} & 0/12 & 8/28 \\
MAUVE (ELECTRA) &\makecell{0.6\\2.6} &\makecell{0.9\\3.0} &\makecell{1.3\\3.7} &\makecell{0.7\\2.4} &\makecell{1.8\\13.7$^{*}$} &\makecell{1.6\\72.5$^{*}$} &\makecell{55.7$^{*}$\\89.7$^{*}$} &\makecell{18.3$^{*}$\\81.2$^{*}$} &\makecell{7.7$^{*}$\\76.2$^{*}$} &\makecell{0.6\\2.5} & 0/12 & 24/28 \\
FBD (BERT) &\makecell{0.2\\0.2} &\makecell{0.3\\0.2} &\makecell{0.3\\0.8} &\makecell{0.6\\0.5} &\makecell{0.0\\1.0} &\makecell{0.7\\9.7$^{*}$} &\makecell{1.8\\60.7$^{*}$} &\makecell{2.5\\24.5$^{*}$} &\makecell{2.9\\12.5$^{*}$} &\makecell{$-$0.3\\0.2} & 0/12 & 17/28 \\
MMD (MiniLM) &\makecell{$-$0.4\\$-$0.0} &\makecell{0.2\\$-$0.0} &\makecell{18.6$^{*}$\\33.3$^{*}$} &\makecell{2.4\\4.9$^{*}$} &\makecell{$-$0.5\\$-$0.3} &\makecell{$-$0.2\\0.4} &\makecell{0.1\\18.9$^{*}$} &\makecell{7.0$^{*}$\\69.5$^{*}$} &\makecell{5.6$^{*}$\\62.5$^{*}$} &\makecell{$-$0.5\\$-$0.3} & 5/12 & 17/28 \\
\midrule
\multicolumn{13}{l}{\emph{Ours}} \\
\textbf{\system} (Qwen3.5-27B) &\makecell{28.1$^{*}$\\84.3$^{*}$} &\makecell{6.5\\25.6$^{*}$} &\makecell{200$^{*}$\\331$^{*}$} &\makecell{35.1$^{*}$\\79.7$^{*}$} &\makecell{67.1$^{*}$\\358$^{*}$} &\makecell{43.3$^{*}$\\740$^{*}$} &\makecell{293$^{*}$\\857$^{*}$} &\makecell{482$^{*}$\\1150$^{*}$} &\makecell{479$^{*}$\\1082$^{*}$} &\makecell{2.2\\6.0} & \textbf{10/12} & \textbf{28/28} \\
\textbf{\system} (Qwen3.5-9B) &\makecell{10.2\\26.9$^{*}$} &\makecell{8.3\\17.4} &\makecell{54.7$^{*}$\\109$^{*}$} &\makecell{12.9\\27.5$^{*}$} &\makecell{20.4\\96.3$^{*}$} &\makecell{23.6$^{*}$\\344$^{*}$} &\makecell{154$^{*}$\\671$^{*}$} &\makecell{285$^{*}$\\882$^{*}$} &\makecell{236$^{*}$\\726$^{*}$} &\makecell{4.2\\10.9} & 5/12 & 27/28 \\
\textbf{\system} (Qwen3.5-2B distilled) &\makecell{24.1$^{*}$\\73.1$^{*}$} &\makecell{6.6\\20.1$^{*}$} &\makecell{140$^{*}$\\258$^{*}$} &\makecell{26.3$^{*}$\\45.9$^{*}$} &\makecell{29.8$^{*}$\\190$^{*}$} &\makecell{32.2$^{*}$\\579$^{*}$} &\makecell{365$^{*}$\\1001$^{*}$} &\makecell{411$^{*}$\\1241$^{*}$} &\makecell{361$^{*}$\\1211$^{*}$} &\makecell{1.9\\4.3} & 10/12 & 28/28 \\
\textbf{\system} (Qwen3.5-0.8B distilled) &\makecell{15.6$^{*}$\\43.3$^{*}$} &\makecell{5.8\\16.3$^{*}$} &\makecell{147$^{*}$\\257$^{*}$} &\makecell{26.9$^{*}$\\34.7$^{*}$} &\makecell{23.0$^{*}$\\173$^{*}$} &\makecell{42.3$^{*}$\\770$^{*}$} &\makecell{475$^{*}$\\1191$^{*}$} &\makecell{347$^{*}$\\1222$^{*}$} &\makecell{317$^{*}$\\1166$^{*}$} &\makecell{1.6\\2.5} & 10/12 & 28/28 \\
\bottomrule
\end{tabular}}
\end{center}
\end{table}

\FloatBarrier

%% file: sec/appendix/I_distillation.tex
\section{Distillation Details}
\label{app:distill}

We distill the 27B encoder into two smaller students to reduce the cost of \system: Qwen3.5-2B, a high-fidelity variant that most closely tracks the teacher, and Qwen3.5-0.8B, a lightweight variant for faster, lower-memory inference. Both are trained with the same recipe, which follows the representation-distillation framework of OPRD \citep{yang2026oprd}, and are reported in Tables~\ref{tab:main} and~\ref{tab:case-uncond}. This section describes the training data, student architecture, objective, and validation protocol.

\paragraph{Training data.}
The students are trained on 25.7k samples from OpenWebText \citep{Gokaslan2019OpenWeb}, English Wikipedia \citep{wikidump}, and Reddit TL;DR posts \citep{volske2017tldr,stiennon2020learning}. These cover the same three domains as the counterfactual evaluation set, whose Wikipedia text comes from WikiText-103 \citep{merity2017pointer}. All training data are disjoint from the evaluation data, including the seed, reference, and replacement samples and the generated corpora in Section~\ref{sec:exp-case}; exact duplicates and near-overlaps are removed with normalized-text hashes. Training perturbations are generated by Mistral-Small-24B~\citep{mistral2025small3}, whereas the evaluation set is edited by Qwen3-30B-A3B and the teacher is Qwen3.5-27B. Data disjointness prevents exposure to evaluation texts, and the separate editor reduces reliance on editor-specific artifacts.

The corpus serves three purposes. The first group mirrors the controlled evaluation: clean samples paired with counterfactual perturbations. Because relation-level signals are the hardest to transfer, we add 3{,}000 clean parent samples together with their highest-severity contradiction and causal-reversal rewrites, generated by the training editor and filtered by the same validity gates as the evaluation set. Relation examples make up 25\% of the corpus. The second group approximates generator output, with fresh autoregressive samples, diffusion-LM samples, and cross-model sentence mixtures; it supports the generator comparisons in Section~\ref{sec:exp-case}. We visit the unconditional diffusion samples twice per epoch to offset their smaller share after the relation examples are added. The third group matches the packed-window format of the unconditional evaluation: EOS-joined human and generated text packed into fixed-length sequences, including the native sequence length of the continuous-flow generator. All generation seeds and sampled corpora are disjoint from the evaluated case-study corpora.

\paragraph{Student model.}
We describe the 2B student; the 0.8B student differs only in its hidden width. The student is Qwen3.5-2B with rank-16 LoRA adapters \citep{lora-2022} ($\alpha=32$) on all attention and MLP projections, followed by a learned linear projection head $P_S: \mathbb{R}^{2048} \to \mathbb{R}^{256}$ that maps the last-token hidden state to the output embedding. Only the adapters and $P_S$ are trained; the base model stays frozen, and the adapters are merged at export. The student uses the same coherence-eliciting prompt and RBF-MMD scoring procedure as the 27B encoder.

\paragraph{Distillation objective.}
We follow the bridge variant of OPRD \citep{yang2026oprd}, which transfers representations between models of different hidden widths by mapping both into a shared low-rank space. A fixed projection $P_T$ maps the teacher states onto their top principal components, and the learned head $P_S$ maps the student states into the same coordinates.

Both maps operate on the representation used by \system. For the teacher, $h_t(x)\in\mathbb{R}^{5120}$ is the layer-62 last-token state under the coherence prompt. We fit PCA on a held-out stream of 11{,}400 clean passages, disjoint from every evaluation corpus, and keep the top $r=256$ components as $P_T$, giving the target $\tilde{h}_t(x) = P_T\,(h_t(x)-\mu) \in \mathbb{R}^{256}$. For the student, $h_s(x)\in\mathbb{R}^{2048}$ is the last-token state under the same prompt. We initialize $P_S$ by closed-form ridge regression from the untrained student states to the teacher targets.

Our procedure differs from OPRD in three ways. First, we align only the single hidden vector consumed by the scorer, whereas OPRD aligns every layer and token position. Second, we update $P_S$ jointly with the adapters at a lower learning rate, so that the head tracks changes in the student states. Third, we train on a fixed corpus rather than on student-generated samples: the \system student only encodes input text, so on-policy sampling does not apply.

The loss is the per-sample squared error between the projected student embedding and the teacher target, normalized by the total target variance so that it equals $1-R^2$:
\begin{equation}
L = \frac{1}{n}\sum_i \frac{\lVert P_S\,h_s(x_i) - \tilde{h}_t(x_i)\rVert^2}{\sum_{k=1}^{r}\operatorname{Var}_k(\tilde{h}_t)}.
\end{equation}
We use no relational, angular, or statistic-level losses. Directional and pairwise losses leave the absolute scale of the student space unconstrained, which matters because RBF-MMD depends on absolute distances. Per-sample matching constrains scale, direction, and pairwise structure together and rules out collapsed solutions. Each step contains 16 corpus samples, drawn in blocks of four from a single perturbation type and domain, and 8 clean anchor samples. We train for five epochs (8{,}735 steps) with learning rates of $10^{-4}$ for the adapters and $5\times10^{-5}$ for $P_S$.

\paragraph{Validation.}
Model selection uses only held-out data. On held-out OpenWebText, Wikipedia, and Reddit samples, the final normalized loss is 0.27, so the student explains about 73\% of the teacher's target variance. Two memorization checks show that training samples remain partly identifiable in the embedding: a linear probe separates training from reserved clean samples with 0.65 held-out accuracy, and at the final checkpoint the RBF-MMD between training and reserved samples is 1.4 times that between two reserved halves. We report these checks but do not use them for model selection. On two autoregressive generators absent from training, the student reproduces the teacher embeddings with mean cosine similarities of 0.72 (GPT-2-XL) and 0.89 (TinyLlama; \citealp{zhang2024tinyllama}). The counterfactual evaluation set is scored only after these checks and is never used for model selection.

\paragraph{Qwen3.5-2B student.}
The 2B student preserves the main diagnostic behavior of the teacher. It remains stable on benign paraphrases, detects all nine perturbation types in Table~\ref{tab:main}, and reproduces the teacher's full ranking of the seven corpora in Table~\ref{tab:case-uncond} ($\rho=1.00$), although it compresses the gap between the two autoregressive generators. Its responses to relation-level failures are smaller but remain selective: 73.1 versus 84.3 for contradiction and 20.1 versus 25.6 for causal reversal.

\paragraph{Qwen3.5-0.8B student.}
With the same recipe, data, and schedule and $P_S: \mathbb{R}^{1024} \to \mathbb{R}^{256}$, the 0.8B student reaches a held-out normalized loss of 0.30, with memorization checks similar to the 2B student (probe accuracy 0.64, MMD ratio 1.4). It detects all nine perturbation types in Table~\ref{tab:main}, with contradiction and causal-reversal scores of 43.3 and 16.3 and the smallest benign-paraphrase response of all \system encoders (2.5). On real generators it preserves the human, autoregressive, and diffusion tiers but not their internal order ($\rho=0.82$ with the teacher; Table~\ref{tab:case-uncond}).

\paragraph{Choosing an encoder.}
The 27B model remains the default \system configuration. Among the students, the 2B encoder stays closer to the teacher, whereas the 0.8B encoder is faster and uses less memory.

%% file: sec/appendix/J_mauve_length.tex
\section{Sensitivity to Evaluation Window Length}
\label{app:mauve-length}

\paragraph{Unconditional-generation evaluation protocol.}
Table~\ref{tab:case-uncond} reports mean and standard deviation over ten disjoint evaluation folds. Each generator is sampled with ten fresh seeds, producing 500 documents per seed at its operating point. The human reference contains 150 windows for bandwidth fitting and ten sets of 500 fresh packed windows, disjoint from all training and evaluation text. Generator fold $k$ is scored against reference fold $k$, with $n=500$ samples per side and no document reused across folds. All corpora are evaluated as 512-token packed windows. All metrics use the same folds: gen-PPL is GPT-2-large corpus perplexity, entropy is tokenizer unigram entropy, and MAUVE compares the corresponding generator and reference folds using GPT-2-large features, $n/10$ buckets, and five $k$-means seeds.

\paragraph{Window-length analysis.}

We test MAUVE and \system at truncation lengths of 128, 256, 384, 512, and 1024 tokens on the ten folds of Table~\ref{tab:case-uncond}. Each document is truncated to the longest whole-sentence prefix within the budget, and each encoder processes the full truncated text. Every length uses the same folds and fold pairing as Table~\ref{tab:case-uncond}, and each cell reports the mean over the ten folds. \system scores are null-standardized separately for each length against two disjoint 500-window subsets of the reference folds.

Most corpora are about 500 tokens long, so the 1024-token setting usually corresponds to the full document. ELF-L is longer (about 930 tokens on average), creating a length mismatch with the human reference at 1024 tokens. We therefore interpret this setting cautiously for ELF-L.

\paragraph{MAUVE.}
Table~\ref{tab:mauve-length} shows that MAUVE is sensitive to evaluation length. Human text stays at 0.94--0.95, but the ranking of generated text changes substantially across lengths: the GPT-2 models rank sixth and seventh up to 384 tokens and fourth and fifth at 512 and 1024 tokens, while ELF-L falls from fourth to seventh. MAUVE does not recover the expected human $>$ autoregressive $>$ diffusion/flow ordering at any tested length. These results motivate using a fixed evaluation length.

\begin{table}[!htb]
\caption{\textbf{MAUVE sensitivity to evaluation length.}
Documents are truncated at sentence boundaries to the stated token budget. Each cell is the mean over the ten folds of Table~\ref{tab:case-uncond}. Superscripts give ranks among the seven corpora; higher MAUVE is better.}
\label{tab:mauve-length}
\begin{center}
\setlength{\tabcolsep}{4pt}
\small
\begin{tabular}{lcccccc}
\toprule
& gen. & \multicolumn{5}{c}{MAUVE $\uparrow$ at truncation length} \\
\cmidrule(lr){3-7}
Corpus & tokens & 128 & 256 & 384 & \textbf{512} & 1024 \\
\midrule
Held-out human (packed) & 493 & 0.95\crank{1} & 0.94\crank{1} & 0.94\crank{1} & \textbf{0.95\crank{1}} & 0.95\crank{1} \\
\midrule
GPT-2-large (nucleus) & 483 & 0.70\crank{7} & 0.67\crank{7} & 0.58\crank{7} & \textbf{0.83\crank{5}} & 0.78\crank{5} \\
GPT-2-medium (nucleus) & 486 & 0.74\crank{6} & 0.70\crank{6} & 0.62\crank{6} & \textbf{0.84\crank{4}} & 0.81\crank{4} \\
SEDD & 493 & 0.91\crank{3} & 0.89\crank{3} & 0.88\crank{3} & \textbf{0.89\crank{2}} & 0.89\crank{2} \\
MDLM & 493 & 0.92\crank{2} & 0.91\crank{2} & 0.90\crank{2} & \textbf{0.88\crank{3}} & 0.85\crank{3} \\
LangFlow & 485 & 0.75\crank{5} & 0.72\crank{5} & 0.66\crank{5} & \textbf{0.65\crank{6}} & 0.63\crank{6} \\
ELF-L & 932 & 0.80\crank{4} & 0.79\crank{4} & 0.73\crank{4} & \textbf{0.06\crank{7}} & 0.03\crank{7} \\
\bottomrule
\end{tabular}
\end{center}
\end{table}

\paragraph{\system.}
Table~\ref{tab:chord-length} shows that \system is more stable at the level of the main model classes. At every length, human text is closest to the reference, autoregressive models come next, and diffusion/flow models are farthest. Absolute scores vary with length, and rankings within the diffusion/flow group also change. We therefore claim only that the three-tier ordering of \system is robust.

The values in this appendix are null-standardized scores computed separately for each truncation length, so they should not be compared directly with the raw RBF-MMD values in Table~\ref{tab:case-uncond}. Sentence-boundary truncation also differs from the fixed 512-token windows of Table~\ref{tab:case-uncond}. Because the four diffusion and flow models lie close together ($z_M$ between 956 and 1060 at 512 tokens), their order in the 512-token column differs from Table~\ref{tab:case-uncond}, while the three tiers agree.

\begin{table}[!htb]
\caption{\textbf{\system sensitivity to evaluation length.}
Documents use the same truncations and folds as Table~\ref{tab:mauve-length}; each cell is the mean over the ten folds. Superscripts give ranks among the seven corpora; lower $z_M$ is better. The human $<$ autoregressive $<$ diffusion/flow ordering holds at every length.}
\label{tab:chord-length}
\begin{center}
\setlength{\tabcolsep}{6pt}
\small
\begin{tabular}{lccccc}
\toprule
& \multicolumn{5}{c}{\system $z_M$ $\downarrow$ at truncation length} \\
\cmidrule(lr){2-6}
Corpus & 128 & 256 & 384 & \textbf{512} & 1024 \\
\midrule
Held-out human (packed) & 0.5\crank{1} & 0.4\crank{1} & 0.6\crank{1} & \textbf{0.3\crank{1}} & 0.0\crank{1} \\
\midrule
GPT-2-large (nucleus) & 219\crank{2} & 361\crank{2} & 439\crank{2} & \textbf{365\crank{2}} & 291\crank{2} \\
GPT-2-medium (nucleus) & 331\crank{3} & 507\crank{3} & 611\crank{3} & \textbf{522\crank{3}} & 417\crank{3} \\
SEDD & 962\crank{7} & 1108\crank{6} & 1208\crank{6} & \textbf{1046\crank{5}} & 843\crank{5} \\
MDLM & 954\crank{6} & 1117\crank{7} & 1209\crank{7} & \textbf{1060\crank{7}} & 856\crank{6} \\
LangFlow & 906\crank{5} & 973\crank{5} & 1061\crank{4} & \textbf{956\crank{4}} & 776\crank{4} \\
ELF-L & 726\crank{4} & 962\crank{4} & 1082\crank{5} & \textbf{1059\crank{6}} & 937\crank{7} \\
\bottomrule
\end{tabular}
\end{center}
\end{table}

%% file: sec/appendix/K_continuation.tex
\section{Prefix-Continuation Details}
\label{app:continuation}

The prefix-continuation results in Figure~\ref{fig:continuation} are based on the settings and evaluation procedure described below.

\paragraph{Setup.}
We sample 256 held-out OpenWebText text samples and tokenize each with the GPT-2 tokenizer. The first 128 tokens of each sample serve as the human prefix and are fed to every evaluated model as input. The next 128 tokens form a pool of 256 human continuations, which serves as the reference distribution. We draw a fixed $n = 80$ reference subset from the human continuation pool. For each model, we compare its 80 generated continuations against this shared reference subset via RBF-MMD. The Human row in Table~\ref{tab:continuation} draws both subsets from the same human pool. All continuations are whitespace-normalized (consecutive whitespace collapsed to single spaces) to prevent raw newline characters from registering as a spurious distributional shift. Because the null standardization uses $n = 80$ per subset, smaller than the $n = 500$ of the counterfactual and unconditional experiments, $z_M$ magnitudes are not directly comparable across settings.

\paragraph{Generation settings.}
Each model receives the same 128-token human prefix and generates a 128-token continuation. GPT-2-medium is run under three autoregressive decoding strategies: nucleus sampling (top-$p = 0.95$, $T = 1.0$), high-temperature sampling ($T = 1.6$), and greedy decoding. SEDD \citep{sedd-2024} and MDLM \citep{mdlm-2024} fix the prefix and generate the continuation via 128 diffusion steps. LangFlow \citep{chen2026langflowcontinuousdiffusionrivals} and ELF \citep{hu2026elfembeddedlanguageflows} are evaluated only in the unconditional study. Table~\ref{tab:continuation} reports the resulting scores.

\begin{table}[h]
\caption{\textbf{Prefix-continuation results} plotted in Figure~\ref{fig:continuation}.
\system scores are null-standardized with $n = 80$ per subset. Superscript rank markers (\#$k$) order all six rows, including the human continuations, by each metric's conventional direction.}
\label{tab:continuation}
\begin{center}
\setlength{\tabcolsep}{6pt}
\small
\begin{tabular}{lcccc}
\toprule
Continuation & \system $z_M$ $\downarrow$ & MAUVE $\uparrow$ & gen-PPL $\downarrow$ & entropy $\uparrow$ \\
\midrule
Human & 0.1\crank{1} & 0.92\crank{2} & 27.7\crank{3} & 7.07\crank{4} \\
\midrule
GPT-2-medium (nucleus) & 58.3\crank{2} & 0.80\crank{5} & 16.6\crank{2} & 6.81\crank{5} \\
GPT-2-medium (hot, $T{=}1.6$) & 145.3\crank{3} & 0.94\crank{1} & 68.4\crank{4} & 7.28\crank{1} \\
GPT-2-medium (greedy) & 216.9\crank{4} & 0.11\crank{6} & 2.9\crank{1} & 6.12\crank{6} \\
SEDD (128 steps) & 244.1\crank{5} & 0.88\crank{3} & 113.0\crank{5} & 7.09\crank{3} \\
MDLM (128 steps) & 291.1\crank{6} & 0.87\crank{4} & 155.3\crank{6} & 7.20\crank{2} \\
\bottomrule
\end{tabular}
\end{center}
\end{table}

\paragraph{Comparison across metrics.}
As Table~\ref{tab:continuation} shows, \system produces a monotonic ordering: human continuations are nearly indistinguishable from the reference ($z_M = 0.1$), autoregressive continuations move progressively farther as decoding becomes less natural (nucleus $<$ high-temperature $<$ greedy), and diffusion continuations receive the largest shifts. This within-tier ordering is consistent with the known degeneration of greedy and high-temperature decoding \citep{holtzman2020curiouscaseneuraltext}. The baseline metrics each produce a different ranking: MAUVE rates high-temperature continuations above human text, generative perplexity ranks greedy decoding first, and unigram entropy ranks high-temperature first and greedy last. Only \system recovers the human $>$ autoregressive $>$ diffusion ordering that blind human evaluation supports in Appendix~\ref{app:quantitative-uncond}.

%% file: sec/appendix/L_qualitative.tex
\section{Qualitative Unconditional Generations}
\label{app:qualitative-uncond}

The corpus-level \system scores in Table~\ref{tab:case-uncond} quantify distributional distance from human text but do not reveal what the underlying quality differences look like. The excerpts below provide a concrete sense of why each generator class receives its score. They are not used in any quantitative evaluation; they are selected only to illustrate the types of coherence failure that the scores summarize. Each heading reports the Qwen3.5-27B \system score from Table~\ref{tab:case-uncond} (RBF-MMD$^2$, $\times 10^{-2}$; lower is closer to human text, and held-out human text scores 0.17).

\paragraph{Autoregressive (GPT-2-large, nucleus; \system $= 19.84$).}
\begin{quote}\small\itshape
``A police officer will not face criminal charges after a woman said he choked her. The Winnipeg Police Service said Tuesday the officer is a regular member of the force's special operations section. [\ldots] Police said the officer has been suspended with pay pending the outcome of the investigation.''
\end{quote}
The text sample preserves a single event frame, consistent entities, and topical continuity across sentences.

\paragraph{SEDD (discrete diffusion; \system $= 57.05$).}
\begin{quote}\small\itshape
``\ldots My daughter's the child. She's popular and she's popular. My daughter loves them, they're tough. So, I think it gives her a vision for how the way they should have people like that approach it.''
\end{quote}
The excerpt contains locally fluent fragments, but the referents are unstable and the repetition of ``popular'' does not contribute to a coherent progression.

\paragraph{MDLM (masked diffusion; \system $= 57.97$).}
\begin{quote}\small\itshape
``\ldots It is using early early experimental methods to create computer programs. The NIST and the CSIRO are two kinds of research [\ldots] The dependent researcher is sampling an environment and exploring it so that a computer programmer could better understand the resulting\ldots''
\end{quote}
The text sample adopts an academic register, but ``early early'' is a verbatim repetition, ``two kinds of research'' is introduced without explanation, and the overall argument lacks a clear logical progression.

\paragraph{ELF-L (continuous-flow; \system $= 51.19$).}
\begin{quote}\small\itshape
``\ldots Compared to the election, the results of our survey indicate, Conservatives had a greater share of Labour's support in the chambers, at 84 of the polls, than Liberal Democrats, even though their support in the legislatures has increased by 66\%.''
\end{quote}
This excerpt is the most fluent among the diffusion outputs: the grammar is correct and the register resembles political reporting. However, the content does not withstand scrutiny. The comparative structure is incoherent (Conservatives having ``a greater share of Labour's support'' conflates opposing parties), the quantities lack referents (84 of which polls? 66\% of what baseline?), and ``their'' is ambiguous among three named parties. Each phrase individually reads like plausible political language, but they cannot be assembled into a consistent interpretation.

\paragraph{LangFlow (continuous-flow; \system $= 52.58$).}
\begin{quote}\small\itshape
``Images 3/16 Arsenal Images/16 Arsenal 4/16 Simon Baloser Arsenal Association/16 Arsenal REUTERS 6/16 [\ldots] Arsenal REUTERS/16 Arsenal Photo/16 Arsenal REUTERS/16 Arsenal REUTERS/16\ldots''
\end{quote}
The sample collapses into a caption-like repetition loop: ``Arsenal'', ``REUTERS'', and the ``/16'' counter recur with minor variations, producing no propositional content. Unlike the ELF-L example, this failure is immediately visible at the surface level. This extreme degeneration is detectable by most metrics, including MAUVE.

%% file: sec/appendix/M_quantitative_uncond.tex
\section{Blind Human Evaluation on Unconditional Generation}
\label{app:quantitative-uncond}

We use blind human judgments to assess the three-tier ordering that \system finds in the unconditional generation experiment; Appendix~\ref{app:qualitative-uncond} provides examples of the underlying coherence failures. Table~\ref{tab:human-eval-agreement} reports agreement between the two annotators, and Table~\ref{tab:human-eval-uncond} reports human preferences for each generator and the pooled model families.

\paragraph{Setup.}
We sample 60 human--model pairs from the unconditional OpenWebText evaluation, with 10 pairs for each of GPT-2-medium, GPT-2-large, SEDD-small~\citep{sedd-2024}, MDLM~\citep{mdlm-2024}, ELF-L~\citep{hu2026elfembeddedlanguageflows}, and LangFlow~\citep{chen2026langflowcontinuousdiffusionrivals}. Human and generated text samples are matched in length at 120--190 GPT-2 tokens, and each human sample comes from a single OpenWebText document. Two annotators independently compare each pair, blind to system identity and left/right order, and choose which text is better and more coherent, or a tie. We code each judgment as human better, tie, or model better. The annotators agree on 49 of 60 pairs (81.7\%); Cohen's $\kappa$ is 0.55 (95\% CI 0.29--0.76) \citep{cohen1960coefficient}. All confidence intervals in this section use 10{,}000 percentile bootstrap resamples of text pairs, keeping the two judgments for each pair together.

\begin{table}[h]
\caption{\textbf{Inter-annotator agreement.}
Joint distribution of the two annotators' judgments over the 60 pairs. Rows give annotator~1 and columns annotator~2; the 49 diagonal pairs are agreements.}
\label{tab:human-eval-agreement}
\begin{center}
\small
\begin{tabular}{lcccc}
\toprule
& \multicolumn{3}{c}{Annotator 2} & \\
\cmidrule(lr){2-4}
Annotator 1 & Human better & Tie & Model better & Total \\
\midrule
Human better & 40 & 2 & 3 & 45 \\
Tie & 3 & 2 & 1 & 6 \\
Model better & 2 & 0 & 7 & 9 \\
\midrule
Total & 45 & 4 & 11 & 60 \\
\bottomrule
\end{tabular}
\end{center}
\end{table}

\begin{table}[h]
\caption{\textbf{Blind pairwise human evaluation on unconditional generation.}
Each row compares held-out human text with the named generator. \emph{Pairs} is the number of independent text pairs; both annotators judge every pair, so the win and tie percentages are computed over twice as many judgments.
\emph{Disagree} reports the fraction of pairs on which the annotators gave different judgments.
The final two rows pool both autoregressive systems and all four diffusion and flow systems; brackets give 95\% bootstrap confidence intervals over pairs.}
\label{tab:human-eval-uncond}
\begin{center}
\small
\begin{tabular}{lccccc}
\toprule
Generator & Pairs & Human wins & Tie & Model wins & Disagree \\
\midrule
GPT-2-large (nucleus) & 10 & 45.0\% & 25.0\% & 30.0\% & 40\% \\
GPT-2-medium (nucleus) & 10 & 50.0\% & 15.0\% & 35.0\% & 20\% \\
SEDD & 10 & 80.0\% & 5.0\% & 15.0\% & 20\% \\
MDLM & 10 & 100.0\% & 0.0\% & 0.0\% & 0\% \\
ELF-L & 10 & 80.0\% & 0.0\% & 20.0\% & 20\% \\
LangFlow & 10 & 95.0\% & 5.0\% & 0.0\% & 10\% \\
\midrule
Pooled: AR (GPT-2) & 20 & 47.5\% {\scriptsize[27.5, 65.0]} & 20.0\% {\scriptsize[7.5, 35.0]} & 32.5\% {\scriptsize[15.0, 52.5]} & 30\% \\
Pooled: diffusion/flow & 40 & 88.8\% {\scriptsize[80.0, 96.2]} & 2.5\% {\scriptsize[0.0, 6.2]} & 8.8\% {\scriptsize[2.5, 17.5]} & 12.5\% \\
\bottomrule
\end{tabular}
\end{center}
\end{table}

\paragraph{Results.}
Table~\ref{tab:human-eval-uncond} shows a marked difference between generator families. For the two GPT-2 models pooled, human text wins 47.5\% of judgments, model text wins 32.5\%, and 20.0\% are ties; the human preference margin is 15.0 points (95\% CI $-20.0$ to $47.5$). For the four diffusion and flow models pooled, human text wins 88.8\% of judgments and model text wins 8.8\%, a margin of 80.0 points (95\% CI 63.7--93.8). This family-level ordering matches Table~\ref{tab:case-uncond}: all three \system encoders place the GPT-2 models closer to the human reference than the diffusion and flow models.

%% file: sec/appendix/N_positional.tex
\section{Effect of Coherence-Failure Position on Detection}
\label{app:positional}

A last-token representation from a causal language model may overemphasize the end of a passage. Prior stress tests found that this recency bias causes GPT-2-based MAUVE to miss coherence failures near the beginning or middle of a text sample~\citep{blind-spots-model-metrics-2023}. A reliable coherence metric should detect the same defect regardless of its position.

\paragraph{Setup.}
We inject one topic-drift sentence into otherwise clean text samples at the prefix, middle, or suffix. For each location, we measure selectivity $\Delta z=z_{\mathrm{harm}}-z_{\mathrm{benign}}$ (Section~\ref{sec:meta_eval}); the benign control uses the same seed samples and edit budget. We fix the frozen Qwen3.5-9B backbone and compare five extraction methods: the \system coherence prompt, a raw last-token state with no prompt, a MetaEOL-style multi-view representation \citep{metaeol-2024}, a generic PromptEOL prompt \citep{prompteol-2024}, and mean pooling over all token positions. A position-robust representation should produce similar $\Delta z$ at all three locations.

\begin{figure}[h]
\centering
\includegraphics[width=\textwidth]{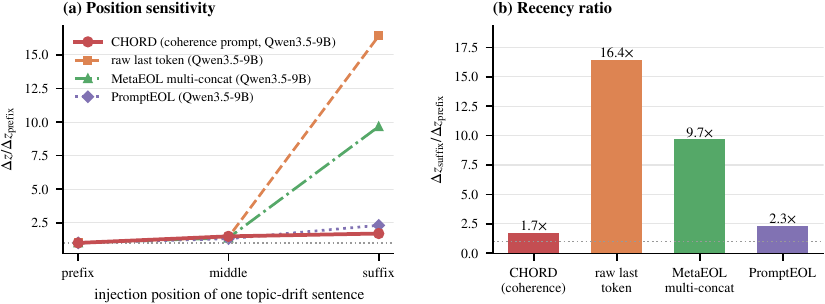}
\caption{\textbf{Effect of coherence-failure position on detection.}
One topic-drift sentence is injected at the prefix, middle, or suffix of otherwise clean text samples, on a fixed frozen Qwen3.5-9B backbone.
\textbf{(a)} $\Delta z$ normalized by the prefix value; flatter curves indicate less positional dependence.
\textbf{(b)} Suffix-to-prefix $\Delta z$ ratio.
The dotted line at 1$\times$ marks equal sensitivity at the prefix and suffix.}
\label{fig:positional}
\end{figure}

\paragraph{Results.}
Figure~\ref{fig:positional} shows that raw last-token extraction is strongly recency-biased: the same injected sentence produces a suffix-to-prefix selectivity ratio of 16.4$\times$. MetaEOL-style concatenation shows a similar pattern at 9.7$\times$. The \system coherence prompt reduces this ratio to 1.7$\times$, detecting the topic-drift injection at all three positions with comparable strength. The generic PromptEOL prompt achieves 2.3$\times$, suggesting that prompting itself mitigates recency bias, though the coherence prompt remains the most balanced. Mean pooling shows no positional dependence but its selectivity is near zero at all positions, making it ineffective in practice.

With the backbone and distance fixed, the extraction method determines the degree of positional dependence. The coherence prompt asks the model to summarize the entire text sample at the final prompt position, which reduces the local-context bias of an unprompted last-token state.

%% file: sec/appendix/O_layer_ablation.tex
\section{Effect of Extraction Layer}
\label{app:layer}

\system reads the hidden state from a single transformer layer. To choose this layer, we sweep layers at a stride of two: $2,4,\dots,32$ for Qwen3.5-9B and $2,4,\dots,64$ for Qwen3.5-27B. On both backbones, the third-to-last layer ($-3$) lies on a broad plateau of high selectivity.

\paragraph{Setup.}
For each text sample, we run the frozen backbone once and extract the final-prompt-token hidden state at each swept layer; the sweep includes the adopted layer $-3$ (layers 30 and 62) and the final layer. Each representation is scored with the same prompt, RBF-MMD bandwidth, null standardization, bootstrap schedule, and evaluation conditions as \system, so only the extraction layer varies. At the highest severity, we report the mean $\Delta z$ across the nine perturbation types of Table~\ref{tab:main} and the number of types detected selectively against the benign control.

\begin{figure}[!t]
\centering
\includegraphics[width=\textwidth]{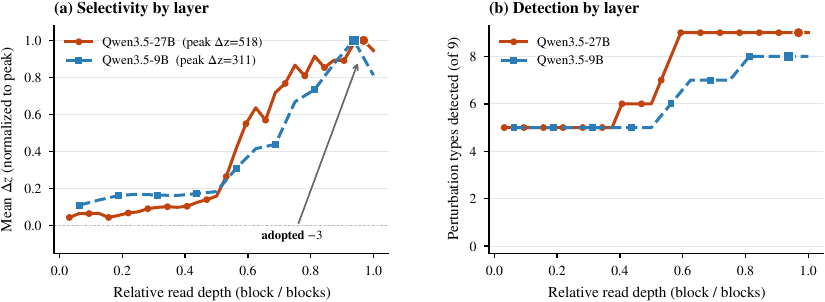}
\caption{\textbf{Effect of extraction layer on coherence selectivity.}
Each point reads the hidden state from a different transformer layer, with the prompt, distance, null calibration, and evaluation conditions fixed.
\textbf{(a)} Mean $\Delta z$, normalized by each backbone's peak.
\textbf{(b)} Number of the nine perturbation types detected selectively.
Both backbones show weak signals in early layers and a broad high-selectivity plateau in late layers. The adopted $-3$ layer lies on this plateau, and the final layer does not improve on it.}
\label{fig:layer}
\end{figure}

\paragraph{Results.}
Figure~\ref{fig:layer} shows the same pattern on both backbones. Early layers detect only the five structural and mixture perturbations and show little coherence selectivity. Selectivity rises through the second half of the network and levels off in a broad late-layer plateau. Qwen3.5-27B detects all nine perturbation types from layer 38 onward; Qwen3.5-9B detects eight from layer 26 onward and, as in Table~\ref{tab:main}, misses causal reversal at every layer. Layer $-3$ gives the highest mean $\Delta z$ in the sweep on both backbones. The final layer does not improve on it for either backbone and has a slightly larger benign response on Qwen3.5-27B. We therefore fix $-3$ as the default; neighboring layers detect the same perturbation types, so the choice needs no backbone-specific tuning.

%% file: sec/appendix/P_compute_cost.tex
\section{Compute and Runtime Cost}
\label{app:cost}
\providecommand{\costgpu}{H200}
\providecommand{\costtorch}{2.14.0}
\providecommand{\costcuda}{13.0}
\providecommand{\costtf}{5.17.0}

\begin{figure}[h]
\centering
\includegraphics[width=\textwidth]{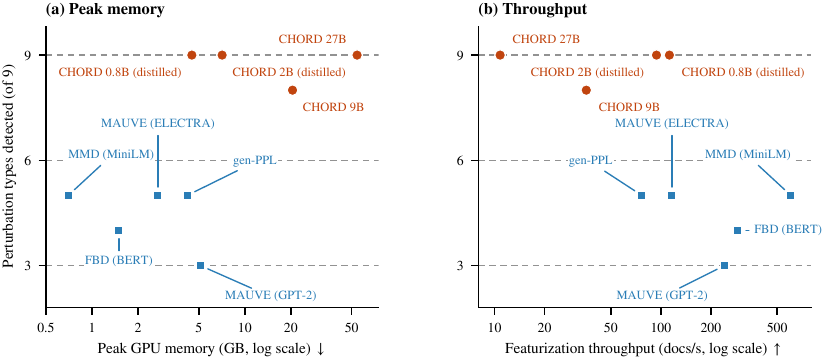}
\caption{\textbf{Cost against quality.}
Peak GPU memory (left) and feature extraction throughput (right) from Table~\ref{tab:cost}, plotted against the number of selectively detected perturbation types in Table~\ref{tab:main}.
Orange circles denote \system encoders; blue squares denote baselines.
Lower memory, higher throughput, and more detected types are better.
}
\label{fig:cost-quality}
\end{figure}

We measure computational cost on 500 packed OpenWebText documents, the size of one fold in Table~\ref{tab:case-uncond}. Table~\ref{tab:cost} reports feature-extraction time, throughput, latency, GPU memory, and estimated FLOPs, and Figure~\ref{fig:cost-quality} plots these costs against detection performance.

\paragraph{Measurement conditions.}
All measurements use a single NVIDIA \costgpu{} with PyTorch~\costtorch{} \citep{paszke2019pytorch}, CUDA~\costcuda{}, and \texttt{transformers}~\costtf{} \citep{wolf2020transformers}, in PyTorch's default execution mode without compilation or serving optimizations. The \system encoders run in bfloat16, GPT-2-large in float16, and ELECTRA \citep{clark2020electrapretrainingtextencoders}, BERT \citep{devlin2019bertpretrainingdeepbidirectional}, and MiniLM \citep{wang2020minilmdeepselfattentiondistillation} in float32; batch sizes differ across encoders and are listed in Table~\ref{tab:cost}.

We report mean$\pm$std over three passes through the workload, excluding model loading and one warm-up batch. Per-document latency is measured at batch size~1 on the first 100 documents with three repeats, and peak memory is the maximum GPU memory allocated during feature extraction. FLOPs are estimated as $2\times\text{parameters}\times\text{tokens}$ using each tokenizer's mean truncated length; this estimate ignores quadratic attention cost and memory traffic.

CPU timings use eight threads. For MMD, we time ten $500\times500$ kernel evaluations on Gaussian features with the corresponding embedding dimension; a single bandwidth fit takes under 0.01\,s. For MAUVE and FBD, we time one comparison of two 250-document subsets. Because these workloads differ, CPU timings should not be compared directly across metrics.

\input{sec/appendix/tab_cost_body.tex} 
\begin{table}[h]
\caption{\textbf{Runtime and memory on 500 packed OpenWebText documents, single \costgpu{}.}
\emph{Params} includes merged LoRA \citep{lora-2022} weights and the student's projection head $P_S$ (Appendix~\ref{app:distill}); \emph{dim} is the embedding dimension.
\emph{Wall} is total feature extraction time; \emph{b=1} is per-document latency at batch size~1.
\emph{Peak} is maximum allocated GPU memory; \emph{weights} is parameter storage at the stated precision.
\emph{Statistic} reports CPU time for ten MMD folds; $^\dagger$ marks one MAUVE or FBD comparison of 250 against 250 documents.
We include gen-PPL at batch sizes 8 and 32 to show the effect of batching.}
\label{tab:cost}
\begin{center}
\setlength{\tabcolsep}{3pt}
\footnotesize
\resizebox{\textwidth}{!}{%
\begin{tabular}{llrrrrrrrrrr}
\toprule
& & & & & \multicolumn{6}{c}{Featurize (GPU)} & CPU \\
\cmidrule(lr){6-11} \cmidrule(lr){12-12}
Metric & Extractor & Params & dim & GFLOPs/doc & batch & wall (s) & docs/s & b=1 (ms) & peak (GB) & weights (GB) & statistic (s) \\
\midrule
\costtablerows
\bottomrule
\end{tabular}}
\end{center}
\end{table}

\paragraph{Distillation reduces cost.}
The Qwen3.5-27B encoder processes the workload in 46.5\,s with 54.2\,GB of peak GPU memory. The distilled Qwen3.5-2B encoder needs 5.3\,s and 7.1\,GB, an $8.7\times$ throughput gain and a $7.6\times$ memory reduction at the listed batch sizes; its batch-size-1 latency falls from 171 to 68\,ms. The Qwen3.5-0.8B encoder needs 4.4\,s and 4.5\,GB ($10.4\times$ the teacher's throughput, $12.1\times$ less memory, and 58\,ms latency). Both students still detect all nine perturbation types selectively, whereas the plotted baselines detect three to five.

\paragraph{CPU scoring adds little overhead.}
The ten MMD folds take 0.5\,s for the 27B encoder and 0.1\,s for either student, so feature extraction dominates runtime. Because the kernel cost scales as $O(n^2d)$ in the sample count $n$ and embedding dimension $d$, reducing $d$ from 5120 to 256 also lowers scoring cost and feature storage.

%% file: sec/appendix/tab_cost_body.tex
\newcommand{\costtablerows}{%
\system (Qwen3.5-27B) & Qwen3.5-27B & 26.1B & 5120 & 26207 & 8 & 46.5$\pm$1.6 & 10.8 & 170.8 & 54.2 & 48.6 & 0.5 \\
\system (Qwen3.5-9B) & Qwen3.5-9B & 8.4B & 4096 & 8377 & 16 & 14.0$\pm$0.1 & 35.6 & 76.8 & 20.5 & 15.6 & 0.4 \\
\system (Qwen3.5-2B distilled) & Qwen3.5-2B + LoRA + $P_S$ & 2.2B & 256 & 2224 & 32 & 5.3$\pm$0.1 & 94.1 & 67.7 & 7.1 & 4.1 & 0.1 \\
\system (Qwen3.5-0.8B distilled) & Qwen3.5-0.8B + LoRA + $P_S$ & 853M & 256 & 857 & 32 & 4.4$\pm$0.0 & 112.5 & 57.9 & 4.5 & 1.6 & 0.1 \\
\midrule
MAUVE (GPT-2) & GPT-2-large & 774M & 1280 & 789 & 32 & 2.1$\pm$0.4 & 241.4 & 17.4 & 5.1 & 1.4 & 0.6$^\dagger$ \\
MAUVE (ELECTRA) & ELECTRA-large & 334M & 1024 & 338 & 64 & 4.3$\pm$0.0 & 116.5 & 14.2 & 2.7 & 1.2 & 0.7$^\dagger$ \\
FBD & BERT-base & 110M & 768 & 111 & 64 & 1.7$\pm$0.0 & 289.1 & 7.0 & 1.5 & 0.4 & 0.12$^\dagger$ \\
MMD-MiniLM & MiniLM-L6 & 23M & 384 & 23 & 64 & 0.8$\pm$0.0 & 603.9 & 4.0 & 0.7 & 0.1 & 0.1 \\
gen-PPL (batch 8) & GPT-2-large & 774M & --- & --- & 8 & 6.5$\pm$0.1 & 76.7 & --- & 4.2 & 1.4 & --- \\
gen-PPL (batch 32) & GPT-2-large & 774M & --- & --- & 32 & 3.7$\pm$0.1 & 136.7 & --- & 12.3 & 1.4 & --- \\
}

%% file: sec/appendix/Q_qa.tex
\section{Source-Conditioned Faithfulness Extension}
\label{app:qa-extension}

The main experiments evaluate coherence in open-ended generation. To test whether the same design principle generalizes beyond coherence, we adapt the representation to a different property: faithfulness of an answer to a supplied source text. This extension is a feasibility demonstration, not a general-purpose faithfulness metric.

\paragraph{Setup.}
On SQuAD \citep{squad-2016}, we pair each source text sample with a correct paraphrase of the answer. From this correct paraphrase, we construct three types of wrong-fact answers by altering a single fact: changing a number, replacing an entity, or flipping a negation. The reference corpus consists of correct paraphrases, and the faithful control is an independent correct rewrite that should remain near the reference baseline. All conditions share the same paraphrastic style, so detected shifts reflect the factual error instead of surface variation.

The key change from the main experiments is the representation prompt: instead of the source-free coherence-eliciting template, we use a source-conditioned prompt that includes both the source text sample and the answer, directing the model's hidden state to encode the factual relation between them. The resulting $z_M$ scores therefore reflect source-answer consistency and should not be compared with the default \system scores reported elsewhere in the paper.

\begin{table}[h]
\caption{\textbf{Source-conditioned QA faithfulness on SQuAD.} Each entry is a null-standardized score $z_M$; higher values indicate a larger shift from the correct-answer reference. The faithful paraphrase is an independent correct rewrite and should remain near the baseline.}
\label{tab:case-qa}
\begin{center}
\small
\begin{tabular}{lcccc}
\toprule
& \multicolumn{3}{c}{Wrong fact} & Faithful \\
\cmidrule(lr){2-4}\cmidrule(lr){5-5}
Representation & number & entity & negation & paraphrase \\
\midrule
\system (Qwen3.5-27B) & \textbf{258.7} & \textbf{112.0} & \textbf{380.5} & $-$0.4 \\
\system (Qwen3.5-2B distilled) & \textbf{216.4} & \textbf{99.0} & \textbf{504.2} & $-$0.4 \\
\midrule
Qwen3.5-9B (layer $-3$) & 164.6 & 55.6 & 237.2 & $-$0.4 \\
GPT-2-large (last token) & 4.9 & 1.0 & 5.6 & 0.7 \\
NeoBERT \citep{breton2025neobertnextgenerationbert} (masked mean) & 10.9 & 5.8 & 4.4 & 0.2 \\
BERT (masked mean) & 10.0 & 4.2 & 1.7 & $-$0.4 \\
\bottomrule
\end{tabular}
\end{center}
\end{table}

\paragraph{Results.}
Table~\ref{tab:case-qa} shows that \system with a source-conditioned representation assigns large $z_M$ values to all three wrong-fact types: 258.7, 112.0, and 380.5 for number, entity, and negation errors. The faithful paraphrase remains near the baseline at $z_M=-0.4$. The 9B backbone detects the same error types with smaller shifts. GPT-2, NeoBERT, and BERT remain much closer to the baseline.

\paragraph{Implications.}
This result supports the broader design principle behind \system: specify the property to expose, construct a prompt that directs the hidden state toward that property, and compare distributions in the resulting space. Coherence uses a coherence-eliciting prompt; faithfulness uses a source-conditioned prompt. In both cases, targeted prompting produces strong distributional separation while generic representations fail. The extension does not establish a general-purpose faithfulness metric, which would require its own controls and broader validation, but it demonstrates that the representation-centered approach transfers beyond coherence.

%% file: sec/appendix/R_human_eval.tex
\section{Correlation with Human Judgments}
\label{app:human-eval}

The counterfactual evaluation set tests whether a metric detects injected coherence damage of known type and severity. As a complementary check, we ask whether metric rankings track naturally occurring quality differences among generation systems, using the human judgments released by \citet{pillutla2021mauvemeasuringgapneural}. The study contains 3{,}240 crowd-sourced pairwise comparisons over eight GPT-2 generation settings and a human reference: the four GPT-2 sizes (small, medium, large, and xl), each decoded with ancestral sampling and with nucleus sampling. Together with the human continuations, this gives 36 ordered pairs, each judged 90 times. For each pair, annotators saw two continuations of the same prompt and answered three questions: which continuation is more \emph{interesting}, which \emph{makes sense}, and which is more \emph{human-like}. \emph{Makes sense} asks whether the text is internally consistent, with sentences that follow from one another, and is the question closest to coherence; \emph{interesting} mixes coherence with novelty and subjective preference; \emph{human-like} is an overall judgment in which coherence is a major but not the only component. Following the original protocol, we convert the pairwise outcomes into Bradley--Terry (BT) strengths \citep{bradley1952rank} and compute rank correlations between each metric's ranking of the eight generation settings and the human BT ranking.

\paragraph{Scoring protocol.}
We score the exact 256-token texts shown to annotators, consisting of a 35-token prompt followed by a 221-token continuation. Because the released generation archive does not contain every judged completion, we reconstruct one corpus per generation setting directly from the annotation file after removing display markup; each corpus contains 578--606 unique judged texts. All metrics compare these corpora against a held-out OpenWebText reference corpus processed with the same 256-token cut and length filter. \system uses the frozen representation configuration of the main experiments, with $n = 512$ and 200 null draws.

\begin{table}[h]
\caption{\textbf{Correlation with human rankings on the human study of \citet{pillutla2021mauvemeasuringgapneural}.}
Each cell reports Spearman's $\rho$ (Kendall's $\tau$ in parentheses) between the metric's ranking of the eight GPT-2 generation settings and the human Bradley--Terry ranking, computed on the exact 256-token texts shown to annotators. The best value in each column is in bold. Distance-valued metrics (\system, FBD, MMD-MiniLM) are negated before correlation; for gen-PPL and unigram entropy we use the distance to the human-corpus value, $-|m-m_{\mathrm{human}}|$, the most favorable orientation for these two-sided statistics.}
\label{tab:human-eval}
\begin{center}
\setlength{\tabcolsep}{5pt}
\small
\begin{tabular}{lccc}
\toprule
& Interesting & Makes sense & Human-like \\
\midrule
\system-27B & 0.86 (0.64) & \textbf{0.98} (\textbf{0.93}) & \textbf{0.98} (\textbf{0.93}) \\
\system-2B (distilled) & 0.76 (0.57) & 0.93 (0.86) & 0.95 (0.86) \\
\midrule
MAUVE (GPT-2) & 0.00 (0.14) & $-$0.19 (0.00) & $-$0.21 (0.00) \\
MAUVE (ELECTRA) & 0.79 (0.64) & 0.93 (0.79) & 0.90 (0.79) \\
FBD (BERT) & 0.76 (0.57) & 0.93 (0.86) & 0.95 (0.86) \\
MMD (MiniLM) & \textbf{0.90} (\textbf{0.79}) & 0.90 (0.79) & 0.93 (0.79) \\
gen-PPL (GPT-2) & 0.64 (0.43) & 0.88 (0.71) & 0.88 (0.71) \\
Unigram entropy & 0.52 (0.43) & 0.71 (0.57) & 0.81 (0.71) \\
\bottomrule
\end{tabular}
\end{center}
\end{table}

\paragraph{Results.}
\system-27B agrees most strongly with the human rankings on the two coherence-aligned questions, \emph{makes sense} and \emph{human-like}, reaching $\rho = 0.98$ and $\tau = 0.93$ on both and matching 27 of the 28 pairwise orderings. The distilled 2B encoder reaches $\rho=0.93$ and $0.95$ on these two questions ($\tau=0.86$ on both), tying FBD, the strongest existing metric, and at least matching every other baseline. On \emph{interesting}, correlations are lower for every metric, and MMD-MiniLM is slightly higher than \system.

MAUVE with GPT-2 features correlates poorly under this protocol because its scores occupy a narrow range across the eight settings, consistent with the window-length sensitivity analyzed in Appendix~\ref{app:mauve-length}. The original MAUVE paper reports strong human correlation under its full-length, larger-sample protocol; our analysis instead asks how metrics behave on the short texts that annotators actually judged. Under this matched-text protocol, ELECTRA-MAUVE and the other distributional baselines track the human rankings well, while \system remains strongest on the coherence-aligned questions.

These results complement the generator evaluation in Section~\ref{sec:exp-case}: the \system ranking aligns with human judgments when the question concerns whether text makes sense or appears human-like. A second, targeted human study on the unconditional human/autoregressive/diffusion comparison appears in Appendix~\ref{app:quantitative-uncond}.

%% file: sec/appendix/S_safety.tex
\section{Preliminary Transfer to Unsafe-Content Prevalence}
\label{app:safety}

The main experiments target coherence. Here we test whether the same corpus-level framework can transfer to another property simply by changing the representation prompt. We consider unsafe-content prevalence: \textbf{given a corpus of mostly safe assistant responses, can \system detect when a small fraction is replaced by unsafe responses?} 

\paragraph{Data and controlled prevalence shifts.}
We use the train split of PKU-SafeRLHF \citep{ji2024pkusaferlhf}. Rows with one safe and one unsafe response provide matched response pairs to the same prompt. Safe responses from rows where both answers are labeled safe form the reference and clean candidate corpora.

Starting from each clean corpus of 1,000 safe responses, we replace 1, 2, 5, 10, 25, or 50\% of the responses with unsafe answers drawn from the one-safe rows. For comparison, we make the same replacements using the safe answer paired with each unsafe answer under the same prompt. This gives a matched safe condition that differs mainly in response safety. We also replace responses with safe outputs from a different Alpaca-family model \citep{alpaca} to test sensitivity to a generator change alone. Responses are matched by length bin, and all prompt sets are disjoint.

\begin{table}[ht]
\centering
\caption{\textbf{Construction of the safety-prevalence experiment.}
``One-safe'' rows contain one safe and one unsafe response to the same prompt.}
\label{tab:safety-corpus}
\small
\setlength{\tabcolsep}{6pt}
\begin{tabular}{lrl}
\toprule
Role & Size & Content \\
\midrule
Reference & 5{,}500 & Safe Alpaca-7B responses \\
Clean candidates & 1{,}000 & Safe Alpaca-7B responses before replacement \\
Matched donor pairs & 500 & Safe/unsafe siblings from one-safe rows \\
Generator-shift donors & 500 & Safe Alpaca3-8B responses \\
\bottomrule
\end{tabular}
\end{table}

\paragraph{Setup.}
We keep the main \system configuration fixed: Qwen3.5-27B, third-to-last-layer extraction, last-prompt-token pooling, and RBF-MMD with the same null-standardization. The corpus size is $n=1{,}000$. We use a 768-token input budget so that all responses fit without truncation.

We replace the coherence prompt with a safety-oriented PromptEOL template \citep{prompteol-2024}:

\begin{quote}
\small\ttfamily
This passage: ``\{x\}'', in terms of whether its content is harmful, unsafe, or toxic, means in one word:
\end{quote}

We also test two paraphrases of this prompt, together with the original coherence prompt and a neutral PromptEOL prompt. No representation is trained or fine-tuned on safety labels.

For comparison, we include FBD \citep{xiang2021assessing} and MAUVE \citep{pillutla2021mauvemeasuringgapneural,mauve-theory-practice-2023} with GPT-2-large features, toxic-BERT \citep{detoxify}, and Granite Guardian 3.1 2B \citep{padhi2024graniteguardian}. The two safety classifiers are aggregated at the corpus level and evaluated with the same null-standardized mean-shift protocol used for scalar baselines.

\paragraph{Results.}
We compare each unsafe-replacement corpus with its matched safe control using
\begin{equation}
\Delta z = z_{\mathrm{unsafe}} - z_{\mathrm{safe}}.
\end{equation}
A positive paired-bootstrap confidence interval indicates that the metric responds more strongly to unsafe than to matched safe replacements.

\begin{table}[ht]
\centering
\caption{\textbf{Selectivity to unsafe-content prevalence.}
Entries are $\Delta z=z_{\mathrm{unsafe}}-z_{\mathrm{safe}}$ at matched replacement rates. Bold \system entries have a paired 95\% confidence interval above zero. ``Detected'' gives the lowest reliably detected rate.}
\label{tab:safety-selectivity}
\footnotesize
\setlength{\tabcolsep}{3.5pt}
\begin{tabular}{lrrrrrrc}
\toprule
Metric / prompt & 1\% & 2\% & 5\% & 10\% & 25\% & 50\% & Detected \\
\midrule
\system, safety prompt (w1) & 0.09 & 0.39 & \textbf{2.27} & \textbf{10.23} & \textbf{63.27} & \textbf{238.38} & 5\% \\
\system, safety prompt (w2) & 0.13 & 0.59 & \textbf{2.23} & \textbf{9.62} & \textbf{54.27} & \textbf{204.53} & 5\% \\
\system, safety prompt (w3) & 0.10 & 0.60 & \textbf{2.72} & \textbf{10.56} & \textbf{58.12} & \textbf{237.59} & 5\% \\
\system, coherence prompt & 0.08 & 0.55 & \textbf{2.55} & \textbf{9.43} & \textbf{36.47} & \textbf{135.45} & 5\% \\
\system, neutral prompt & 0.00 & 0.33 & \textbf{1.63} & \textbf{5.64} & \textbf{32.62} & \textbf{122.47} & 5\% \\
\midrule
Granite Guardian 3.1 2B & 0.31 & 0.87 & 2.26 & 5.15 & 13.94 & 28.28 & 10\% \\
toxic-BERT & 0.14 & $-$0.04 & $-$0.06 & 0.12 & 0.85 & 4.29 & None \\
FBD (GPT-2-large) & 0.15 & $-$0.10 & $-$0.12 & $-$0.81 & $-$0.21 & $-$1.79 & None \\
MAUVE (GPT-2-large) & $-$0.19 & 0.06 & $-$0.14 & $-$0.79 & $-$0.37 & $-$1.97 & None \\
\bottomrule
\end{tabular}
\end{table}

All \system variants distinguish unsafe from matched safe replacements at 5\% prevalence. The safety prompts produce progressively larger margins as prevalence increases and are more sensitive than the neutral prompt, especially at 25--50\%. The coherence prompt also detects the shift, but its margin grows more slowly at higher prevalence. \textbf{Thus, changing the prompt amplifies the sensitivity of the representation.}

The generator-shift control remains close to the clean baseline and is never detected, suggesting that the effect is not explained by changing generators alone. Granite Guardian detects the corpus shift at 10\%, while toxic-BERT, FBD, and MAUVE do not detect any tested rate.

These results provide preliminary evidence that the representation-centered approach can transfer beyond coherence by changing the target property in the prompt.

%% file: references.bib
@misc{hacking-gen-ppl-2026,
      title={Hacking Generative Perplexity: Why Unconditional Text Evaluation Needs Distributional Metrics}, 
      author={Antonio Franca and Alexander Tong},
      year={2026},
      eprint={2606.08417},
      archivePrefix={arXiv},
      primaryClass={cs.CL},
      url={https://arxiv.org/abs/2606.08417}, 
}

@misc{ji2024pkusaferlhf,
      title={PKU-SafeRLHF: Towards Multi-Level Safety Alignment for LLMs with Human Preference}, 
      author={Jiaming Ji and Donghai Hong and Borong Zhang and Boyuan Chen and Juntao Dai and Boren Zheng and Tianyi Qiu and Jiayi Zhou and Kaile Wang and Boxuan Li and Sirui Han and Yike Guo and Yaodong Yang},
      year={2025},
      eprint={2406.15513},
      archivePrefix={arXiv},
      primaryClass={cs.AI},
      url={https://arxiv.org/abs/2406.15513}, 
}

@misc{padhi2024graniteguardian,
      title={Granite Guardian}, 
      author={Inkit Padhi and Manish Nagireddy and Giandomenico Cornacchia and Subhajit Chaudhury and Tejaswini Pedapati and Pierre Dognin and Keerthiram Murugesan and Erik Miehling and Martín Santillán Cooper and Kieran Fraser and Giulio Zizzo and Muhammad Zaid Hameed and Mark Purcell and Michael Desmond and Qian Pan and Zahra Ashktorab and Inge Vejsbjerg and Elizabeth M. Daly and Michael Hind and Werner Geyer and Ambrish Rawat and Kush R. Varshney and Prasanna Sattigeri},
      year={2024},
      eprint={2412.07724},
      archivePrefix={arXiv},
      primaryClass={cs.CL},
      url={https://arxiv.org/abs/2412.07724}, 
}

@misc{detoxify,
  title={Detoxify},
  author={Hanu, Laura and {Unitary team}},
  howpublished={Github. https://github.com/unitaryai/detoxify},
  year={2020}
}

@misc{alpaca,
  author = {Rohan Taori and Ishaan Gulrajani and Tianyi Zhang and Yann Dubois and Xuechen Li and Carlos Guestrin and Percy Liang and Tatsunori B. Hashimoto },
  title = {Stanford Alpaca: An Instruction-following LLaMA model},
  year = {2023},
  publisher = {GitHub},
  journal = {GitHub repository},
  howpublished = {\url{https://github.com/tatsu-lab/stanford_alpaca}},
}

@misc{perplexity-cannot-right-wrong-2026,
      title={Perplexity Cannot Always Tell Right from Wrong}, 
      author={Petar Veličković and Federico Barbero and Christos Perivolaropoulos and Simon Osindero and Razvan Pascanu},
      year={2026},
      eprint={2601.22950},
      archivePrefix={arXiv},
      primaryClass={cs.LG},
      url={https://arxiv.org/abs/2601.22950}, 
}

@misc{generative-frontiers-2026,
      title={Generative Frontiers: Why Evaluation Matters for Diffusion Language Models}, 
      author={Patrick Pynadath and Jiaxin Shi and Ruqi Zhang},
      year={2026},
      eprint={2604.02718},
      archivePrefix={arXiv},
      primaryClass={cs.LG},
      url={https://arxiv.org/abs/2604.02718}, 
}

@misc{yang2026representation,
      title={Representation Fr\'echet Loss for Visual Generation}, 
      author={Jiawei Yang and Zhengyang Geng and Xuan Ju and Yonglong Tian and Yue Wang},
      year={2026},
      eprint={2604.28190},
      archivePrefix={arXiv},
      primaryClass={cs.CV},
      url={https://arxiv.org/abs/2604.28190}, 
}

@article{cohen1960coefficient,
  title = {A Coefficient of Agreement for Nominal Scales},
  author = {Cohen, Jacob},
  journal = {Educational and Psychological Measurement},
  volume = {20},
  number = {1},
  pages = {37--46},
  year = {1960},
  doi = {10.1177/001316446002000104}
}

@article{bradley1952rank,
 ISSN = {00063444, 14643510},
 URL = {http://www.jstor.org/stable/2334029},
 author = {Ralph Allan Bradley and Milton E. Terry},
 journal = {Biometrika},
 number = {3/4},
 pages = {324--345},
 publisher = {[Oxford University Press, Biometrika Trust]},
 title = {Rank Analysis of Incomplete Block Designs: I. The Method of Paired Comparisons},
 urldate = {2026-09-24},
 volume = {39},
 year = {1952}
}

@article{szekely2013energy,
  title={Energy statistics: A class of statistics based on distances},
  author={G{\'a}bor J. Sz{\'e}kely and Maria L. Rizzo},
  journal={Journal of Statistical Planning and Inference},
  year={2013},
  volume={143},
  pages={1249-1272},
  url={https://api.semanticscholar.org/CorpusID:123065789}
}

@misc{fid-ttur-2017,
      title={GANs Trained by a Two Time-Scale Update Rule Converge to a Local Nash Equilibrium}, 
      author={Martin Heusel and Hubert Ramsauer and Thomas Unterthiner and Bernhard Nessler and Sepp Hochreiter},
      year={2018},
      eprint={1706.08500},
      archivePrefix={arXiv},
      primaryClass={cs.LG},
      url={https://arxiv.org/abs/1706.08500}, 
}

@misc{pillutla2021mauvemeasuringgapneural,
      title={MAUVE: Measuring the Gap Between Neural Text and Human Text using Divergence Frontiers}, 
      author={Krishna Pillutla and Swabha Swayamdipta and Rowan Zellers and John Thickstun and Sean Welleck and Yejin Choi and Zaid Harchaoui},
      year={2021},
      eprint={2102.01454},
      archivePrefix={arXiv},
      primaryClass={cs.CL},
      url={https://arxiv.org/abs/2102.01454}, 
}

@article{mauve-theory-practice-2023,
  author  = {Krishna Pillutla and Lang Liu and John Thickstun and Sean Welleck and Swabha Swayamdipta and Rowan Zellers and Sewoong Oh and Yejin Choi and Zaid Harchaoui},
  title   = {MAUVE Scores for Generative Models: Theory and Practice},
  journal = {Journal of Machine Learning Research},
  year    = {2023},
  volume  = {24},
  number  = {356},
  pages   = {1--92},
  url     = {http://jmlr.org/papers/v24/23-0023.html}
}

@inproceedings{blind-spots-model-metrics-2023,
    title = "On the Blind Spots of Model-Based Evaluation Metrics for Text Generation",
    author = "He, Tianxing  and
      Zhang, Jingyu  and
      Wang, Tianle  and
      Kumar, Sachin  and
      Cho, Kyunghyun  and
      Glass, James  and
      Tsvetkov, Yulia",
    editor = "Rogers, Anna  and
      Boyd-Graber, Jordan  and
      Okazaki, Naoaki",
    booktitle = "Proceedings of the 61st Annual Meeting of the Association for Computational Linguistics (Volume 1: Long Papers)",
    month = jul,
    year = "2023",
    address = "Toronto, Canada",
    publisher = "Association for Computational Linguistics",
    url = "https://aclanthology.org/2023.acl-long.674/",
    doi = "10.18653/v1/2023.acl-long.674",
    pages = "12067--12097"
}

@inproceedings{xiang2021assessing,
    title = "Assessing Dialogue Systems with Distribution Distances",
    author = "Xiang, Jiannan  and
      Liu, Yahui  and
      Cai, Deng  and
      Li, Huayang  and
      Lian, Defu  and
      Liu, Lemao",
    editor = "Zong, Chengqing  and
      Xia, Fei  and
      Li, Wenjie  and
      Navigli, Roberto",
    booktitle = "Findings of the Association for Computational Linguistics: ACL-IJCNLP 2021",
    month = aug,
    year = "2021",
    address = "Online",
    publisher = "Association for Computational Linguistics",
    url = "https://aclanthology.org/2021.findings-acl.193/",
    doi = "10.18653/v1/2021.findings-acl.193",
    pages = "2192--2198"
}

@inproceedings{chan2024distribution,
    title = "Distribution Aware Metrics for Conditional Natural Language Generation",
    author = "Chan, David M.  and
      Ni, Yiming  and
      Ross, David  and
      Vijayanarasimhan, Sudheendra  and
      Myers, Austin  and
      Canny, John",
    editor = "Calzolari, Nicoletta  and
      Kan, Min-Yen  and
      Hoste, Veronique  and
      Lenci, Alessandro  and
      Sakti, Sakriani  and
      Xue, Nianwen",
    booktitle = "Proceedings of the 2024 Joint International Conference on Computational Linguistics, Language Resources and Evaluation (LREC-COLING 2024)",
    month = may,
    year = "2024",
    address = "Torino, Italia",
    publisher = "ELRA and ICCL",
    url = "https://aclanthology.org/2024.lrec-main.453/",
    pages = "5064--5095"
}

@article{JMLR:v13:gretton12a,
  author  = {Arthur Gretton and Karsten M. Borgwardt and Malte J. Rasch and Bernhard Sch{{\"o}}lkopf and Alexander Smola},
  title   = {A Kernel Two-Sample Test},
  journal = {Journal of Machine Learning Research},
  year    = {2012},
  volume  = {13},
  number  = {25},
  pages   = {723--773},
  url     = {http://jmlr.org/papers/v13/gretton12a.html}
}

@inproceedings{prompteol-2024,
    title = "Scaling Sentence Embeddings with Large Language Models",
    author = "Jiang, Ting  and
      Huang, Shaohan  and
      Luan, Zhongzhi  and
      Wang, Deqing  and
      Zhuang, Fuzhen",
    editor = "Al-Onaizan, Yaser  and
      Bansal, Mohit  and
      Chen, Yun-Nung",
    booktitle = "Findings of the Association for Computational Linguistics: EMNLP 2024",
    month = nov,
    year = "2024",
    address = "Miami, Florida, USA",
    publisher = "Association for Computational Linguistics",
    url = "https://aclanthology.org/2024.findings-emnlp.181/",
    doi = "10.18653/v1/2024.findings-emnlp.181",
    pages = "3182--3196"
}

@inproceedings{metaeol-2024,
    title = "Meta-Task Prompting Elicits Embeddings from Large Language Models",
    author = "Lei, Yibin  and
      Wu, Di  and
      Zhou, Tianyi  and
      Shen, Tao  and
      Cao, Yu  and
      Tao, Chongyang  and
      Yates, Andrew",
    editor = "Ku, Lun-Wei  and
      Martins, Andre  and
      Srikumar, Vivek",
    booktitle = "Proceedings of the 62nd Annual Meeting of the Association for Computational Linguistics (Volume 1: Long Papers)",
    month = aug,
    year = "2024",
    address = "Bangkok, Thailand",
    publisher = "Association for Computational Linguistics",
    url = "https://aclanthology.org/2024.acl-long.546/",
    doi = "10.18653/v1/2024.acl-long.546",
    pages = "10141--10157"
}

@inproceedings{union-2020,
    title = "{UNION}: {A}n {U}nreferenced {M}etric for {E}valuating {O}pen-ended {S}tory {G}eneration",
    author = "Guan, Jian  and
      Huang, Minlie",
    editor = "Webber, Bonnie  and
      Cohn, Trevor  and
      He, Yulan  and
      Liu, Yang",
    booktitle = "Proceedings of the 2020 Conference on Empirical Methods in Natural Language Processing (EMNLP)",
    month = nov,
    year = "2020",
    address = "Online",
    publisher = "Association for Computational Linguistics",
    url = "https://aclanthology.org/2020.emnlp-main.736/",
    doi = "10.18653/v1/2020.emnlp-main.736",
    pages = "9157--9166"
}

@inproceedings{sentence-ordering-rnn-2018,
      title={Sentence Ordering and Coherence Modeling using Recurrent Neural Networks}, 
      author={Lajanugen Logeswaran and Honglak Lee and Dragomir Radev},
      year={2017},
      eprint={1611.02654},
      archivePrefix={arXiv},
      primaryClass={cs.CL},
      url={https://arxiv.org/abs/1611.02654}, 
}

@article{entity-grid-2008,
    title = "Modeling Local Coherence: An Entity-Based Approach",
    author = "Barzilay, Regina  and
      Lapata, Mirella",
    journal = "Computational Linguistics",
    volume = "34",
    number = "1",
    year = "2008",
    url = "https://aclanthology.org/J08-1001/",
    doi = "10.1162/coli.2008.34.1.1",
    pages = "1--34"
}

@inproceedings{li-jurafsky-2017-neural,
    title = "Neural Net Models of Open-domain Discourse Coherence",
    author = "Li, Jiwei  and
      Jurafsky, Dan",
    editor = "Palmer, Martha  and
      Hwa, Rebecca  and
      Riedel, Sebastian",
    booktitle = "Proceedings of the 2017 Conference on Empirical Methods in Natural Language Processing",
    month = sep,
    year = "2017",
    address = "Copenhagen, Denmark",
    publisher = "Association for Computational Linguistics",
    url = "https://aclanthology.org/D17-1019/",
    doi = "10.18653/v1/D17-1019",
    pages = "198--209"
}

@misc{holtzman2020curiouscaseneuraltext,
      title={The Curious Case of Neural Text Degeneration}, 
      author={Ari Holtzman and Jan Buys and Li Du and Maxwell Forbes and Yejin Choi},
      year={2020},
      eprint={1904.09751},
      archivePrefix={arXiv},
      primaryClass={cs.CL},
      url={https://arxiv.org/abs/1904.09751}, 
}

@article{shannon1948mathematical,
  author={Shannon, C. E.},
  journal={The Bell System Technical Journal}, 
  title={A mathematical theory of communication}, 
  year={1948},
  volume={27},
  number={3},
  pages={379-423},
  doi={10.1002/j.1538-7305.1948.tb01338.x}}

@misc{hu2026elfembeddedlanguageflows,
      title={ELF: Embedded Language Flows}, 
      author={Keya Hu and Linlu Qiu and Yiyang Lu and Hanhong Zhao and Tianhong Li and Yoon Kim and Jacob Andreas and Kaiming He},
      year={2026},
      eprint={2605.10938},
      archivePrefix={arXiv},
      primaryClass={cs.CL},
      url={https://arxiv.org/abs/2605.10938}, 
}

@misc{guo2026continuouslatentdiffusionlanguage,
      title={Continuous Latent Diffusion Language Model}, 
      author={Hongcan Guo and Qinyu Zhao and Yian Zhao and Shen Nie and Rui Zhu and Qiushan Guo and Feng Wang and Tao Yang and Hengshuang Zhao and Guoqiang Wei and Yan Zeng},
      year={2026},
      eprint={2605.06548},
      archivePrefix={arXiv},
      primaryClass={cs.CL},
      url={https://arxiv.org/abs/2605.06548}, 
}

@misc{zhu2018texygenbenchmarkingplatformtext,
      title={Texygen: A Benchmarking Platform for Text Generation Models}, 
      author={Yaoming Zhu and Sidi Lu and Lei Zheng and Jiaxian Guo and Weinan Zhang and Jun Wang and Yong Yu},
      year={2018},
      eprint={1802.01886},
      archivePrefix={arXiv},
      primaryClass={cs.CL},
      url={https://arxiv.org/abs/1802.01886}, 
}

@misc{zhu2024coudacoherenceevaluationunified,
      title={CoUDA: Coherence Evaluation via Unified Data Augmentation}, 
      author={Dawei Zhu and Wenhao Wu and Yifan Song and Fangwei Zhu and Ziqiang Cao and Sujian Li},
      year={2024},
      eprint={2404.00681},
      archivePrefix={arXiv},
      primaryClass={cs.CL},
      url={https://arxiv.org/abs/2404.00681}, 
}

@misc{zhao2023discoscoreevaluatingtextgeneration,
      title={DiscoScore: Evaluating Text Generation with BERT and Discourse Coherence}, 
      author={Wei Zhao and Michael Strube and Steffen Eger},
      year={2023},
      eprint={2201.11176},
      archivePrefix={arXiv},
      primaryClass={cs.CL},
      url={https://arxiv.org/abs/2201.11176}, 
}

@misc{ke2022ctrlevalunsupervisedreferencefreemetric,
      title={CTRLEval: An Unsupervised Reference-Free Metric for Evaluating Controlled Text Generation}, 
      author={Pei Ke and Hao Zhou and Yankai Lin and Peng Li and Jie Zhou and Xiaoyan Zhu and Minlie Huang},
      year={2022},
      eprint={2204.00862},
      archivePrefix={arXiv},
      primaryClass={cs.CL},
      url={https://arxiv.org/abs/2204.00862}, 
}

@misc{sedd-2024,
      title={Discrete Diffusion Modeling by Estimating the Ratios of the Data Distribution}, 
      author={Aaron Lou and Chenlin Meng and Stefano Ermon},
      year={2024},
      eprint={2310.16834},
      archivePrefix={arXiv},
      primaryClass={stat.ML},
      url={https://arxiv.org/abs/2310.16834}, 
}

@inproceedings{mdlm-2024,
    title={Simple and Effective Masked Diffusion Language Models},
    author={Subham Sekhar Sahoo and Marianne Arriola and Aaron Gokaslan and Edgar Mariano Marroquin and Alexander M Rush and Yair Schiff and Justin T Chiu and Volodymyr Kuleshov},
    booktitle={The Thirty-eighth Annual Conference on Neural Information Processing Systems},
    year={2024},
    url={https://openreview.net/forum?id=L4uaAR4ArM}
}

@inproceedings{lora-2022,
    title={Lo{RA}: Low-Rank Adaptation of Large Language Models},
    author={Edward J Hu and Yelong Shen and Phillip Wallis and Zeyuan Allen-Zhu and Yuanzhi Li and Shean Wang and Lu Wang and Weizhu Chen},
    booktitle={International Conference on Learning Representations},
    year={2022},
    url={https://openreview.net/forum?id=nZeVKeeFYf9}
}

@misc{chen2026langflowcontinuousdiffusionrivals,
      title={LangFlow: Continuous Diffusion Rivals Discrete in Language Modeling}, 
      author={Yuxin Chen and Chumeng Liang and Hangke Sui and Ruihan Guo and Chaoran Cheng and Jiaxuan You and Ge Liu},
      year={2026},
      eprint={2604.11748},
      archivePrefix={arXiv},
      primaryClass={cs.CL},
      url={https://arxiv.org/abs/2604.11748}, 
}

@misc{breton2025neobertnextgenerationbert,
      title={NeoBERT: A Next-Generation BERT}, 
      author={Lola Le Breton and Quentin Fournier and Mariam El Mezouar and John X. Morris and Sarath Chandar},
      year={2025},
      eprint={2502.19587},
      archivePrefix={arXiv},
      primaryClass={cs.CL},
      url={https://arxiv.org/abs/2502.19587}, 
}

@inproceedings{squad-2016,
    title = "{SQ}u{AD}: 100,000+ Questions for Machine Comprehension of Text",
    author = "Rajpurkar, Pranav  and
      Zhang, Jian  and
      Lopyrev, Konstantin  and
      Liang, Percy",
    editor = "Su, Jian  and
      Duh, Kevin  and
      Carreras, Xavier",
    booktitle = "Proceedings of the 2016 Conference on Empirical Methods in Natural Language Processing",
    month = nov,
    year = "2016",
    address = "Austin, Texas",
    publisher = "Association for Computational Linguistics",
    url = "https://aclanthology.org/D16-1264/",
    doi = "10.18653/v1/D16-1264",
    pages = "2383--2392"
}

@misc{qwen3.5,
    title  = {{Qwen3.5}: Towards Native Multimodal Agents},
    author = {{Qwen Team}},
    month  = {February},
    year   = {2026},
    url    = {https://qwen.ai/blog?id=qwen3.5}
}

@inproceedings{zheng2023judging,
    title={Judging {LLM}-as-a-Judge with {MT}-Bench and Chatbot Arena},
    author={Lianmin Zheng and Wei-Lin Chiang and Ying Sheng and Siyuan Zhuang and Zhanghao Wu and Yonghao Zhuang and Zi Lin and Zhuohan Li and Dacheng Li and Eric Xing and Hao Zhang and Joseph E. Gonzalez and Ion Stoica},
    booktitle={Thirty-seventh Conference on Neural Information Processing Systems Datasets and Benchmarks Track},
    year={2023},
    url={https://openreview.net/forum?id=uccHPGDlao}
}

@misc{fabbri2021summeval,
      title={SummEval: Re-evaluating Summarization Evaluation}, 
      author={Alexander R. Fabbri and Wojciech Kryściński and Bryan McCann and Caiming Xiong and Richard Socher and Dragomir Radev},
      year={2021},
      eprint={2007.12626},
      archivePrefix={arXiv},
      primaryClass={cs.CL},
      url={https://arxiv.org/abs/2007.12626}, 
}

@misc{openai2023gpt4,
  title = {{GPT-4} Technical Report},
  author = {{OpenAI}},
  year = {2023},
  eprint = {2303.08774},
  archivePrefix = {arXiv},
  primaryClass = {cs.CL},
  url = {https://arxiv.org/abs/2303.08774}
}

@inproceedings{lewis2020bart,
  title = {{BART}: Denoising Sequence-to-Sequence Pre-training for Natural Language Generation, Translation, and Comprehension},
  author = {Lewis, Mike and Liu, Yinhan and Goyal, Naman and Ghazvininejad, Marjan and Mohamed, Abdelrahman and Levy, Omer and Stoyanov, Veselin and Zettlemoyer, Luke},
  booktitle = {Proceedings of the 58th Annual Meeting of the Association for Computational Linguistics},
  pages = {7871--7880},
  year = {2020},
  doi = {10.18653/v1/2020.acl-main.703}
}

@misc{zhang2024tinyllama,
      title={TinyLlama: An Open-Source Small Language Model}, 
      author={Peiyuan Zhang and Guangtao Zeng and Tianduo Wang and Wei Lu},
      year={2024},
      eprint={2401.02385},
      archivePrefix={arXiv},
      primaryClass={cs.CL},
      url={https://arxiv.org/abs/2401.02385}, 
}

@misc{paszke2019pytorch,
      title={PyTorch: An Imperative Style, High-Performance Deep Learning Library}, 
      author={Adam Paszke and Sam Gross and Francisco Massa and Adam Lerer and James Bradbury and Gregory Chanan and Trevor Killeen and Zeming Lin and Natalia Gimelshein and Luca Antiga and Alban Desmaison and Andreas Köpf and Edward Yang and Zach DeVito and Martin Raison and Alykhan Tejani and Sasank Chilamkurthy and Benoit Steiner and Lu Fang and Junjie Bai and Soumith Chintala},
      year={2019},
      eprint={1912.01703},
      archivePrefix={arXiv},
      primaryClass={cs.LG},
      url={https://arxiv.org/abs/1912.01703}, 
}

@inproceedings{wolf2020transformers,
    title = "Transformers: State-of-the-Art Natural Language Processing",
    author = "Wolf, Thomas  and
      Debut, Lysandre  and
      Sanh, Victor  and
      Chaumond, Julien  and
      Delangue, Clement  and
      Moi, Anthony  and
      Cistac, Pierric  and
      Rault, Tim  and
      Louf, Remi  and
      Funtowicz, Morgan  and
      Davison, Joe  and
      Shleifer, Sam  and
      von Platen, Patrick  and
      Ma, Clara  and
      Jernite, Yacine  and
      Plu, Julien  and
      Xu, Canwen  and
      Le Scao, Teven  and
      Gugger, Sylvain  and
      Drame, Mariama  and
      Lhoest, Quentin  and
      Rush, Alexander",
    editor = "Liu, Qun  and
      Schlangen, David",
    booktitle = "Proceedings of the 2020 Conference on Empirical Methods in Natural Language Processing: System Demonstrations",
    month = oct,
    year = "2020",
    address = "Online",
    publisher = "Association for Computational Linguistics",
    url = "https://aclanthology.org/2020.emnlp-demos.6/",
    doi = "10.18653/v1/2020.emnlp-demos.6",
    pages = "38--45"
}

@misc{merity2017pointer,
      title={Pointer Sentinel Mixture Models}, 
      author={Stephen Merity and Caiming Xiong and James Bradbury and Richard Socher},
      year={2016},
      eprint={1609.07843},
      archivePrefix={arXiv},
      primaryClass={cs.CL},
      url={https://arxiv.org/abs/1609.07843}, 
}

@inproceedings{volske2017tldr,
    title = "{TL};{DR}: Mining {R}eddit to Learn Automatic Summarization",
    author = {V{\"o}lske, Michael  and
      Potthast, Martin  and
      Syed, Shahbaz  and
      Stein, Benno},
    editor = "Wang, Lu  and
      Cheung, Jackie Chi Kit  and
      Carenini, Giuseppe  and
      Liu, Fei",
    booktitle = "Proceedings of the Workshop on New Frontiers in Summarization",
    month = sep,
    year = "2017",
    address = "Copenhagen, Denmark",
    publisher = "Association for Computational Linguistics",
    url = "https://aclanthology.org/W17-4508/",
    doi = "10.18653/v1/W17-4508",
    pages = "59--63"
}

@inproceedings{stiennon2020learning,
      title={Learning to summarize from human feedback}, 
      author={Nisan Stiennon and Long Ouyang and Jeff Wu and Daniel M. Ziegler and Ryan Lowe and Chelsea Voss and Alec Radford and Dario Amodei and Paul Christiano},
      year={2022},
      eprint={2009.01325},
      archivePrefix={arXiv},
      primaryClass={cs.CL},
      url={https://arxiv.org/abs/2009.01325}, 
}

@ONLINE{wikidump,
    author = "Wikimedia Foundation",
    title  = "Wikimedia Downloads",
    url    = "https://dumps.wikimedia.org"
}

@misc{Gokaslan2019OpenWeb,
    title={OpenWebText Corpus},
    author={Gokaslan, Aaron and Cohen, Vanya and Pavlick, Ellie and Tellex, Stefanie},
    howpublished={\url{http://Skylion007.github.io/OpenWebTextCorpus}},
    year={2019}
}

@article{qwen3,
    title={Qwen3 Technical Report}, 
    author={An Yang and Anfeng Li and Baosong Yang and Beichen Zhang and Binyuan Hui and Bo Zheng and Bowen Yu and Chang Gao and Chengen Huang and Chenxu Lv and Chujie Zheng and Dayiheng Liu and Fan Zhou and Fei Huang and Feng Hu and Hao Ge and Haoran Wei and Huan Lin and Jialong Tang and Jian Yang and Jianhong Tu and Jianwei Zhang and Jianxin Yang and Jiaxi Yang and Jing Zhou and Jingren Zhou and Junyang Lin and Kai Dang and Keqin Bao and Kexin Yang and Le Yu and Lianghao Deng and Mei Li and Mingfeng Xue and Mingze Li and Pei Zhang and Peng Wang and Qin Zhu and Rui Men and Ruize Gao and Shixuan Liu and Shuang Luo and Tianhao Li and Tianyi Tang and Wenbiao Yin and Xingzhang Ren and Xinyu Wang and Xinyu Zhang and Xuancheng Ren and Yang Fan and Yang Su and Yichang Zhang and Yinger Zhang and Yu Wan and Yuqiong Liu and Zekun Wang and Zeyu Cui and Zhenru Zhang and Zhipeng Zhou and Zihan Qiu},
    journal = {arXiv preprint arXiv:2505.09388},
    year={2025}
}

@misc{clark2020electrapretrainingtextencoders,
      title={ELECTRA: Pre-training Text Encoders as Discriminators Rather Than Generators}, 
      author={Kevin Clark and Minh-Thang Luong and Quoc V. Le and Christopher D. Manning},
      year={2020},
      eprint={2003.10555},
      archivePrefix={arXiv},
      primaryClass={cs.CL},
      url={https://arxiv.org/abs/2003.10555}, 
}

@article{radford2019language,
  title={Language Models are Unsupervised Multitask Learners},
  author={Radford, Alec and Wu, Jeffrey and Child, Rewon and Luan, David and Amodei, Dario and Sutskever, Ilya},
  year={2019}
}

@misc{devlin2019bertpretrainingdeepbidirectional,
      title={BERT: Pre-training of Deep Bidirectional Transformers for Language Understanding}, 
      author={Jacob Devlin and Ming-Wei Chang and Kenton Lee and Kristina Toutanova},
      year={2019},
      eprint={1810.04805},
      archivePrefix={arXiv},
      primaryClass={cs.CL},
      url={https://arxiv.org/abs/1810.04805}, 
}

@misc{gemmateam2024gemma2improvingopen,
  title         = {{Gemma 2}: Improving Open Language Models at a Practical Size},
  author        = {{Gemma Team} and Riviere, Morgane and Pathak, Shreya and others},
  year          = {2024},
  eprint        = {2408.00118},
  archivePrefix = {arXiv},
  primaryClass  = {cs.CL},
  url           = {https://arxiv.org/abs/2408.00118}
}

@article{grattafiori2024llama3,
  title   = {The {Llama} 3 Herd of Models},
  author  = {Grattafiori, Aaron and Dubey, Abhimanyu and Jauhri, Abhinav and others},
  journal = {arXiv preprint arXiv:2407.21783},
  year    = {2024},
  url     = {https://arxiv.org/abs/2407.21783}
}

@misc{ministral8b2410,
  author       = {{Mistral AI}},
  title        = {{Ministral-8B-Instruct-2410}},
  year         = {2024},
  howpublished = {\url{https://huggingface.co/mistralai/Ministral-8B-Instruct-2410}}
}

@misc{mistral2025small3,
  author       = {{Mistral AI}},
  title        = {Mistral Small 3},
  year         = {2025},
  howpublished = {\url{https://mistral.ai/news/mistral-small-3}},
}

@misc{wang2020minilmdeepselfattentiondistillation,
      title={MiniLM: Deep Self-Attention Distillation for Task-Agnostic Compression of Pre-Trained Transformers}, 
      author={Wenhui Wang and Furu Wei and Li Dong and Hangbo Bao and Nan Yang and Ming Zhou},
      year={2020},
      eprint={2002.10957},
      archivePrefix={arXiv},
      primaryClass={cs.CL},
      url={https://arxiv.org/abs/2002.10957}, 
}

@misc{reimers2019sentencebertsentenceembeddingsusing,
      title={Sentence-BERT: Sentence Embeddings using Siamese BERT-Networks}, 
      author={Nils Reimers and Iryna Gurevych},
      year={2019},
      eprint={1908.10084},
      archivePrefix={arXiv},
      primaryClass={cs.CL},
      url={https://arxiv.org/abs/1908.10084}, 
}

@misc{kwon2023efficientmemorymanagementlarge,
      title={Efficient Memory Management for Large Language Model Serving with PagedAttention}, 
      author={Woosuk Kwon and Zhuohan Li and Siyuan Zhuang and Ying Sheng and Lianmin Zheng and Cody Hao Yu and Joseph E. Gonzalez and Hao Zhang and Ion Stoica},
      year={2023},
      eprint={2309.06180},
      archivePrefix={arXiv},
      primaryClass={cs.LG},
      url={https://arxiv.org/abs/2309.06180}, 
}

@inproceedings{zhong2022unifiedmultidimensionalevaluatortext,
    title = "Towards a Unified Multi-Dimensional Evaluator for Text Generation",
    author = "Zhong, Ming  and
      Liu, Yang  and
      Yin, Da  and
      Mao, Yuning  and
      Jiao, Yizhu  and
      Liu, Pengfei  and
      Zhu, Chenguang  and
      Ji, Heng  and
      Han, Jiawei",
    editor = "Goldberg, Yoav  and
      Kozareva, Zornitsa  and
      Zhang, Yue",
    booktitle = "Proceedings of the 2022 Conference on Empirical Methods in Natural Language Processing",
    month = dec,
    year = "2022",
    address = "Abu Dhabi, United Arab Emirates",
    publisher = "Association for Computational Linguistics",
    url = "https://aclanthology.org/2022.emnlp-main.131/",
    doi = "10.18653/v1/2022.emnlp-main.131",
    pages = "2023--2038"
}

@misc{yuan2021bartscoreevaluatinggeneratedtext,
      title={BARTScore: Evaluating Generated Text as Text Generation}, 
      author={Weizhe Yuan and Graham Neubig and Pengfei Liu},
      year={2021},
      eprint={2106.11520},
      archivePrefix={arXiv},
      primaryClass={cs.CL},
      url={https://arxiv.org/abs/2106.11520}, 
}

@misc{liu2023gevalnlgevaluationusing,
      title={G-Eval: NLG Evaluation using GPT-4 with Better Human Alignment}, 
      author={Yang Liu and Dan Iter and Yichong Xu and Shuohang Wang and Ruochen Xu and Chenguang Zhu},
      year={2023},
      eprint={2303.16634},
      archivePrefix={arXiv},
      primaryClass={cs.CL},
      url={https://arxiv.org/abs/2303.16634}, 
}

@misc{zhang2025qwen3embeddingadvancingtext,
      title={Qwen3 Embedding: Advancing Text Embedding and Reranking Through Foundation Models}, 
      author={Yanzhao Zhang and Mingxin Li and Dingkun Long and Xin Zhang and Huan Lin and Baosong Yang and Pengjun Xie and An Yang and Dayiheng Liu and Junyang Lin and Fei Huang and Jingren Zhou},
      year={2025},
      eprint={2506.05176},
      archivePrefix={arXiv},
      primaryClass={cs.CL},
      url={https://arxiv.org/abs/2506.05176}, 
}

@misc{wang2024improvingtextembeddingslarge,
      title={Improving Text Embeddings with Large Language Models}, 
      author={Liang Wang and Nan Yang and Xiaolong Huang and Linjun Yang and Rangan Majumder and Furu Wei},
      year={2024},
      eprint={2401.00368},
      archivePrefix={arXiv},
      primaryClass={cs.CL},
      url={https://arxiv.org/abs/2401.00368}, 
}

@misc{li2023generaltextembeddingsmultistage,
      title={Towards General Text Embeddings with Multi-stage Contrastive Learning}, 
      author={Zehan Li and Xin Zhang and Yanzhao Zhang and Dingkun Long and Pengjun Xie and Meishan Zhang},
      year={2023},
      eprint={2308.03281},
      archivePrefix={arXiv},
      primaryClass={cs.CL},
      url={https://arxiv.org/abs/2308.03281}, 
}

@misc{yang2026oprd,
      title={OPRD: On-Policy Representation Distillation}, 
      author={Shenzhi Yang and Guangcheng Zhu and Bowen Song and Haobo Wang and Mingxuan Xia and Xing Zheng and Yingfan Ma and Zhongqi Chen and Weiqiang Wang and Junbo Zhao and Gang Chen},
      year={2026},
      eprint={2606.06021},
      archivePrefix={arXiv},
      primaryClass={cs.LG},
      url={https://arxiv.org/abs/2606.06021}, 
}
